\documentclass[11pt]{article}

\usepackage[margin=1in]{geometry}
\usepackage{amsmath,amssymb,amsfonts}
\newcommand{\cmark}{\ding{51}}
\usepackage{pifont}
\usepackage{graphicx}
\usepackage{booktabs}
\usepackage{hyperref}
\usepackage{cleveref}
\usepackage{natbib}
\usepackage{xcolor}
\usepackage{caption}
\usepackage{subcaption}
\usepackage{microtype}
\usepackage{multirow}

\newcommand{\R}{\mathbb{R}}
\newcommand{\norm}[1]{\left\|#1\right\|}

\title{The Geometry of Multi-Task Grokking: \\
Transverse Instability, Superposition, and Weight Decay Phase Structure}

\author{%
  Yongzhong Xu
  \thanks{abbyxu@gmail.com; code at \url{https://github.com/skydancerosel/grokking-integrability}}
}

\date{February 19, 2026}

\begin{document}
\maketitle

\begin{abstract}

Grokking---the abrupt transition from memorization to generalization long after near-zero training loss---has been extensively studied in single-task settings.
Here we extend geometric analysis of grokking to multi-task modular arithmetic, training shared-trunk Transformers on dual-task (mod-add $+$ mod-mul) and tri-task (mod-add $+$ mod-mul $+$ mod-sq) objectives across a systematic weight decay sweep.
Across up to 90 runs, we combine trajectory PCA, commutator defect analysis, Hessian eigenspectra, and causal gradient perturbations to characterize the geometry of multi-task generalization.

We find five consistent phenomena.
\textbf{(1)~Staggered grokking order:} multiplication generalizes first, followed by squaring and then addition, with consistent delays across seeds.
\textbf{(2)~Universal integrability:} optimization trajectories remain confined to an empirically invariant low-dimensional execution manifold, with commutator defects orthogonal to this manifold and defect onset reliably preceding generalization ($42/42$ conditions).
\textbf{(3)~Weight decay phase structure:} grokking timescale, curvature depth, reconstruction threshold, and defect lead covary systematically with weight decay, revealing distinct dynamical regimes and a sharp no-decay failure mode.
\textbf{(4)~Holographic incompressibility:} although final solutions occupy only 4--8 principal trajectory directions, they are distributed across full-rank weights and are destroyed by minimal perturbations; neither per-layer SVD, magnitude pruning, nor uniform scaling preserves performance.
\textbf{(5)~Transverse fragility and redundancy:} removing less than 10\% of orthogonal gradient components eliminates grokking, yet dual-task models exhibit partial recovery under extreme deletion, suggesting redundant center manifolds enabled by overparameterization.
\textbf{(6)~Spectral geometry of the attention operator transition:} singular value decomposition of the attention weight matrices reveals that grokking corresponds to a spectral symmetry-breaking event---a degenerate query/key spectrum gives way to rank-1 dominance---and that every grokking run traces a universal loop in the spectral-gap--commutator phase plane, with the bottom transformer layer bearing $1.5$--$2.4\times$ stronger non-commutativity than the top layer across all conditions tested.

Together, these results support a dynamical picture in which multi-task grokking constructs a compact superposition subspace in parameter space, with weight decay acting as compression pressure and excess parameters supplying geometric redundancy in optimization pathways.

\end{abstract}

\section{Introduction}
\label{sec:intro}

Grokking refers to a striking training phenomenon in which a neural network memorizes its training data for a prolonged period before abruptly transitioning to near-perfect generalization.
Originally observed in small Transformers trained on modular arithmetic tasks \citep{power2022grokking}, grokking challenges conventional intuitions about overfitting, implicit regularization, and optimization geometry.
Recent work has analyzed grokking through the lenses of mechanistic interpretability \citep{nanda2023grokking, zhong2024clock}, circuit efficiency and competition \citep{merrill2023tale, varma2023explaining}, implicit bias dynamics \citep{lyu2024dichotomy, kumar2024grokking}, and loss landscape curvature \citep{liu2022omnigrok}.
However, most existing studies focus on single-task training, leaving open whether the geometric structure underlying grokking persists when a shared model must simultaneously learn multiple algorithmic circuits.

Multi-task grokking raises qualitatively new questions.
When a Transformer jointly learns modular addition and multiplication---or three operations at once---how are multiple algorithms accommodated within a shared parameter space?
Does the low-dimensional trajectory structure observed in single-task settings \citep{xu2026lowdim} survive?
Are the circuits geometrically separated, superposed, or competing for representational capacity?
And does overparameterization provide redundancy in the presence of multiple objectives?

Unlike prior work that focuses on either representational superposition \citep{elhage2022superposition} or learning dynamics in isolation \citep{davies2023unifying, kumar2024grokking}, our study combines trajectory geometry, curvature analysis, and causal perturbations to provide a unified dynamical account of multi-task grokking.

In companion studies, we showed that single-task grokking trajectories lie within a remarkably low-dimensional execution manifold and exhibit near-integrable dynamics, with curvature concentrated in transverse directions \citep{xu2026integrability}, and that qualitatively similar early-warning signals arise in Dyck languages and the SCAN compositional-generalization benchmark \citep{xu2026earlywarning}.
Here we extend that geometric framework to multi-task training.
We train shared-trunk Transformers on dual-task and tri-task modular arithmetic across a systematic weight decay sweep and analyze their dynamics using trajectory PCA, commutator defect projection, Hessian eigenvalue estimation, reconstruction experiments, and causal gradient perturbations.

Our findings reveal a structured and coherent picture:

\begin{itemize}
    \item \textbf{Staggered task generalization.}
    Across seeds and weight decay values, multiplication consistently generalizes before addition, with tri-task training introducing a further hierarchy (mul $\to$ sq $\to$ add).
    Task-specific heads converge toward near-orthogonality, indicating geometric separation of readout directions.

    \item \textbf{Empirical invariance of the execution manifold.}
    Despite multiple objectives, optimization trajectories remain confined to a low-dimensional manifold with near-perfect integrability.
    Commutator defect onset reliably precedes generalization, serving as an early warning signal across all tested regimes.

    \item \textbf{Weight decay as a phase parameter.}
    Grokking timescale scales systematically with weight decay, with distinct dynamical regimes emerging.
    Strong decay yields fast transitions and deeper saddle curvature; weak decay yields prolonged incubation and intermittent defect dynamics; zero decay produces curvature without generalization.

    \item \textbf{Low-dimensional yet globally distributed solutions.}
    Final generalizing models occupy only a small number of principal trajectory directions ($k^* \approx 4$--$8$) despite containing over $3 \times 10^5$ parameters.
    Yet these solutions are full-rank and highly fragile: post-hoc compression methods fail catastrophically, and even $\pm 5\%$ uniform scaling destroys performance.

    \item \textbf{Transverse fragility and manifold redundancy.}
    Fine-grained orthogonal gradient deletion reveals a sharp fragility threshold near 10\%.
    Dual-task models can recover under extreme deletion, whereas tri-task models cannot, suggesting that overparameterization supplies redundant center manifolds---alternative geometric pathways to generalization.

    \item \textbf{Spectral geometry of the attention operator transition.}
    SVD of the attention weight matrices reveals a universal phase trajectory: every grokking run traces a competition--instability--alignment loop in the $(\sigma_1 - \sigma_2,\;\norm{[W_Q, W_K]}_F)$ plane, with multi-task grokking occurring at successive positions along a single shared trajectory.
    The bottom transformer layer bears $1.5$--$2.4\times$ stronger non-commutativity than the top layer, and this amplitude asymmetry is robust across all seeds and datasets.
\end{itemize}

We interpret these results as evidence that multi-task grokking dynamically constructs a compact superposition subspace in parameter space.
Training proceeds along a dominant memorization scaffold, while generalization-relevant structure accumulates in orthogonal directions.
Weight decay regulates the dimensionality of this subspace, and excess parameters provide redundancy in escape pathways through the loss landscape.

By combining trajectory geometry, curvature analysis, and causal perturbations in a unified experimental framework, this work extends grokking from a single-task curiosity to a multi-task geometric phenomenon, revealing structured phase behavior, transverse fragility, and parameter-space superposition in shared-trunk Transformers.

\paragraph{Paper outline.}
\Cref{sec:setup} describes the experimental setup.
\Cref{sec:training,sec:manifold,sec:integ,sec:hessian,sec:wd_phase,sec:incompress} present empirical results across six domains: training dynamics, manifold structure, integrability, Hessian analysis, the weight decay phase diagram, and incompressibility.
\Cref{sec:spectral_geometry} analyzes the spectral geometry of the attention operator transition, revealing universal phase trajectories and layer asymmetry in instability magnitude.
\Cref{sec:superposition} synthesizes these findings into a unified account of superposition dynamics in parameter space.
\Cref{sec:overparam} develops a theory of overparameterization as manifold redundancy, grounded in the transverse ablation results.
\Cref{sec:discussion} connects our findings to broader themes, discusses limitations, and proposes interpretive frameworks.

\section{Experimental Setup}
\label{sec:setup}

\subsection{Models and Tasks}

All models use a Transformer encoder with pre-norm (LayerNorm before attention and FFN), $d_\text{model} = 128$, 4 attention heads, $d_\text{ff} = 256$, GELU activation, and no dropout.
Inputs are two integer tokens $a, b \in \{0, \ldots, 96\}$; predictions are made from the first token position through task-specific linear heads $\R^{128} \to \R^{97}$.

\paragraph{Dual-task.}
A 2-layer Transformer with a shared trunk and two classification heads, jointly trained on $(x + y) \bmod 97$ and $(x \cdot y) \bmod 97$.
Total parameters: ${\sim}303$k.
Loss is the sum of both cross-entropy terms.

\paragraph{Tri-task.}
The same architecture with an additional head for $(x^2 + y^2) \bmod 97$, giving ${\sim}315$k parameters.
Loss is the sum of three cross-entropy terms.

\subsection{Training Protocol}

Training uses AdamW ($\beta_1 = 0.9$, $\beta_2 = 0.98$) at learning rate $10^{-3}$, batch size 512, gradient clipping at 1.0, and a 50/50 train/test split.
We define \textbf{grokking} as test accuracy reaching 98\% and remaining there for 3 consecutive evaluations.
The \textbf{grok step} is the first step at which test accuracy reaches 90\%.

\subsection{Weight Decay Sweep}

We train models at weight decay $\lambda \in \{0.0, 0.1, 0.2, 0.3, 0.5, 1.0\}$ with 3 seeds each (42, 137, 2024).
For dual-task, all non-zero WD values yield grokking; $\lambda = 0$ never groks.
Training budgets range from 30k steps ($\lambda = 1.0$) to 250k steps ($\lambda = 0.1$) to accommodate the timescale variation.
Tri-task runs cover the full sweep $\lambda \in \{0.0, 0.1, 0.2, 0.3, 0.5, 1.0\}$ with training budgets up to 250k steps.

\subsection{Checkpointing}

We save full-parameter checkpoints every 200 steps (100 steps for faster runs), yielding 50--830 checkpoints per run.
For PCA, we subsample to ${\leq}100$ checkpoints, compute $\Delta\theta(t) = \theta(t) - \theta(0)$, and perform SVD on the resulting $T \times P$ matrix without centering (uncentered PCA), following the intrinsic dimensionality framework of \citet{li2018measuring}, which preserves the dominant drift direction essential for reconstruction.
We verified that centered PCA yields qualitatively similar $k^*$ and reconstruction thresholds, but suppresses the dominant drift direction associated with memorization.
Uncentered PCA preserves this structure and is therefore used throughout.

\label{sec:results}

\section{Training Dynamics: Staggered Multi-Task Grokking}
\label{sec:training}

All non-zero weight decay conditions produce successful multi-task grokking, with a consistent ordering across seeds and settings.

\paragraph{Dual-task ordering.}
Multiplication consistently groks before addition (\Cref{tab:dual_grok}), with delays of 100--400 steps.
This ordering holds across all five WD values and three seeds (with one exception at $\lambda = 0.1$, seed~2024, where addition groks first by 100 steps).

\begin{table}[t]
\centering
\caption{Dual-task grokking steps (mean $\pm$ std across 3 seeds). Multiplication consistently leads.}
\label{tab:dual_grok}
\begin{tabular}{@{}lrrrr@{}}
\toprule
$\lambda$ & Grok (Add) & Grok (Mul) & Gap & Timescale \\
\midrule
1.0 & $13{,}333 \pm 2{,}620$ & $13{,}133 \pm 2{,}748$ & $200$ & fast \\
0.5 & $16{,}267 \pm 685$ & $16{,}200 \pm 712$ & $67$ & fast \\
0.3 & $28{,}467 \pm 1{,}586$ & $28{,}267 \pm 1{,}190$ & $200$ & intermediate \\
0.2 & $45{,}167 \pm 1{,}307$ & $45{,}067 \pm 1{,}184$ & $100$ & intermediate \\
0.1 & $97{,}967 \pm 3{,}172$ & $98{,}833 \pm 2{,}001$ & $-866$ & slow \\
0.0 & --- & --- & --- & (never) \\
\bottomrule
\end{tabular}
\end{table}

\paragraph{Tri-task ordering.}
With three tasks, a clear hierarchy emerges: multiplication groks first, then the quadratic sum ($x^2 + y^2$), then addition (\Cref{tab:tri_grok}).
The ordering mul $\to$ sq $\to$ add is consistent across all three seeds at $\lambda = 1.0$ and makes intuitive sense: $x^2 + y^2$ shares multiplicative structure with $x \cdot y$, while addition is algebraically distinct.

\begin{table}[t]
\centering
\caption{Tri-task grokking steps (per-seed). The ordering mul $\to$ sq $\to$ add is consistent.}
\label{tab:tri_grok}
\begin{tabular}{@{}llrrr@{}}
\toprule
$\lambda$ & Seed & Grok (Add) & Grok (Mul) & Grok (Sq) \\
\midrule
\multirow{3}{*}{1.0} & 42 & 30,500 & 17,900 & 24,800 \\
 & 137 & 19,000 & 15,700 & 17,500 \\
 & 2024 & 13,900 & 12,300 & 13,000 \\
\midrule
\multirow{3}{*}{0.5} & 42 & 27,000 & 24,800 & 26,200 \\
 & 137 & 26,800 & 25,900 & 26,800 \\
 & 2024 & 25,000 & 24,300 & 24,300 \\
\midrule
\multirow{3}{*}{0.3} & 42 & 62,300 & 62,100 & 62,400 \\
 & 137 & 53,300 & 51,900 & 53,600 \\
 & 2024 & 47,800 & 47,800 & 47,800 \\
\midrule
\multirow{3}{*}{0.2} & 42 & 87,400 & 83,500 & 86,200 \\
 & 137 & 90,000 & 83,000 & 89,900 \\
 & 2024 & 69,100 & 69,100 & 69,400 \\
\midrule
\multirow{3}{*}{0.1} & 42 & 162,900 & 159,000 & 165,800 \\
 & 137 & 227,500 & 225,300 & 232,400 \\
 & 2024 & 149,200 & 149,300 & 145,900 \\
\bottomrule
\end{tabular}
\end{table}

\begin{figure}[t]
    \centering
    \begin{subfigure}[t]{0.48\textwidth}
        \centering
        \includegraphics[width=\textwidth]{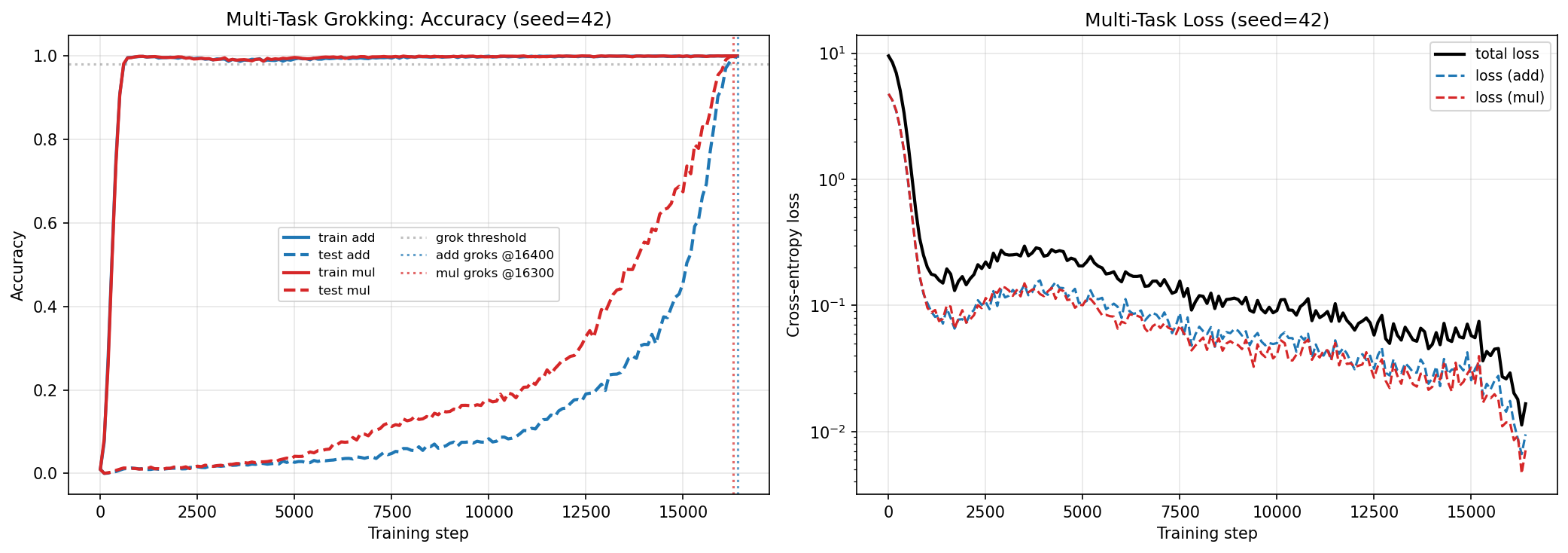}
        \caption{Dual-task (WD=1.0, seed~42): both tasks grok, multiplication leads by ${\sim}100$ steps.}
        \label{fig:dual_acc}
    \end{subfigure}
    \hfill
    \begin{subfigure}[t]{0.48\textwidth}
        \centering
        \includegraphics[width=\textwidth]{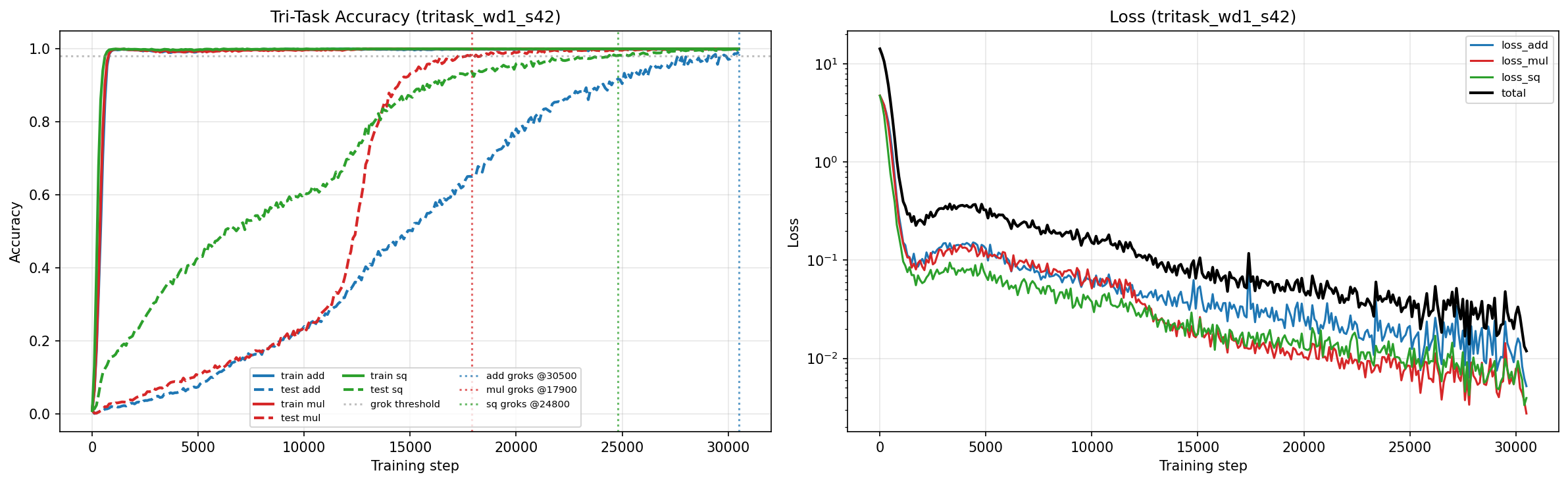}
        \caption{Tri-task (WD=1.0, seed~42): all three tasks grok with ordering mul $\to$ sq $\to$ add.}
        \label{fig:tri_acc}
    \end{subfigure}
    \caption{Multi-task grokking dynamics. \textbf{(a)}~Dual-task: multiplication leads addition. \textbf{(b)}~Tri-task: a three-way staggered ordering emerges.}
    \label{fig:training_curves}
\end{figure}

\paragraph{No-WD controls.}
At $\lambda = 0$, neither dual-task nor tri-task models generalize within extended training budgets (50k--250k steps), confirming that weight decay is essential for multi-task grokking as in the single-task case.

\section{Manifold Structure: Lower Rank and Orthogonal Heads}
\label{sec:manifold}

\begin{figure}[t]
    \centering
    \begin{subfigure}[t]{0.48\textwidth}
        \centering
        \includegraphics[width=\textwidth]{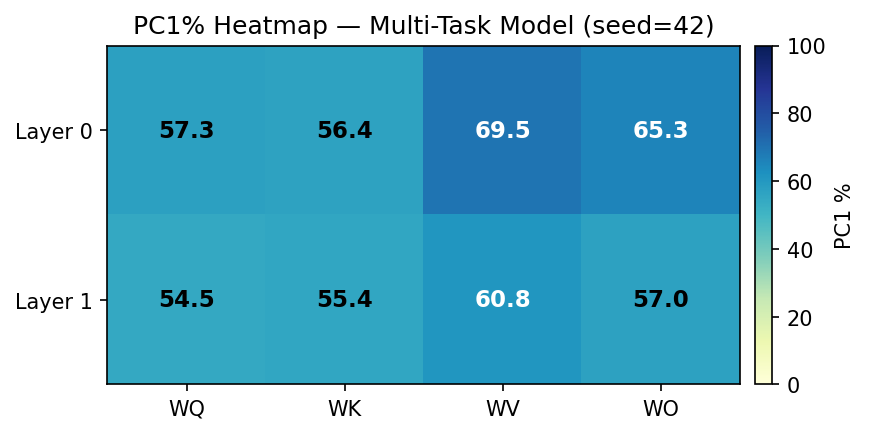}
        \caption{Dual-task PC1\% heatmap (seed~42, WD=1.0): values range from 55\% to 77\%.}
        \label{fig:dual_heatmap}
    \end{subfigure}
    \hfill
    \begin{subfigure}[t]{0.48\textwidth}
        \centering
        \includegraphics[width=\textwidth]{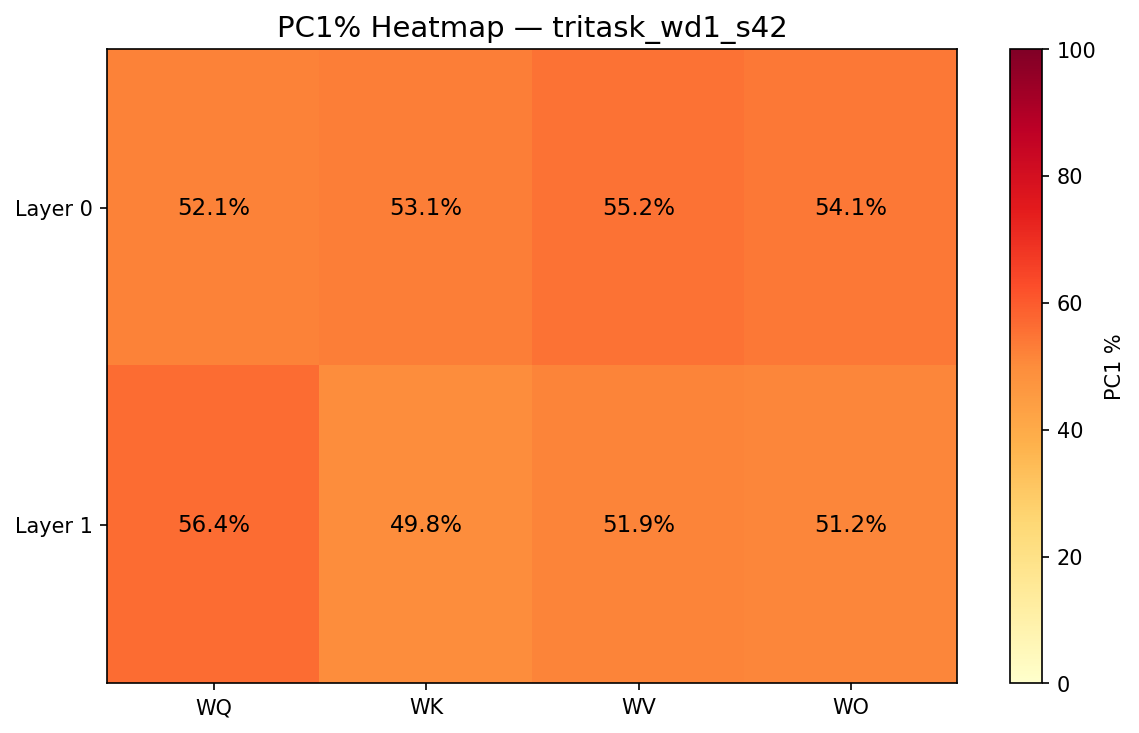}
        \caption{Tri-task PC1\% heatmap (seed~42, WD=1.0): values drop to 49--56\%.}
        \label{fig:tri_heatmap}
    \end{subfigure}
    \caption{PC1\% decreases with task count. \textbf{(a)}~Dual-task: 55--77\%. \textbf{(b)}~Tri-task: 49--56\%. The manifold is no longer rank-1 but remains strongly low-dimensional.}
    \label{fig:pca_multi}
\end{figure}

\paragraph{PC1\% is lower than single-task.}
PCA on attention weight trajectories reveals that multi-task grokking produces lower PC1\% than single-task training (\Cref{fig:pca_multi}).
Dual-task models show PC1\% of 55--77\% (vs.\ 70--94\% single-task), with $W_V$ matrices retaining the highest concentration (69--77\%) and $W_Q/W_K$ dropping to 55--59\%.
Tri-task models show even lower values (49--65\% at $\lambda = 1.0$), suggesting the effective manifold rank scales with task count.

\begin{figure}[t]
    \centering
    \begin{subfigure}[t]{0.48\textwidth}
        \centering
        \includegraphics[width=\textwidth]{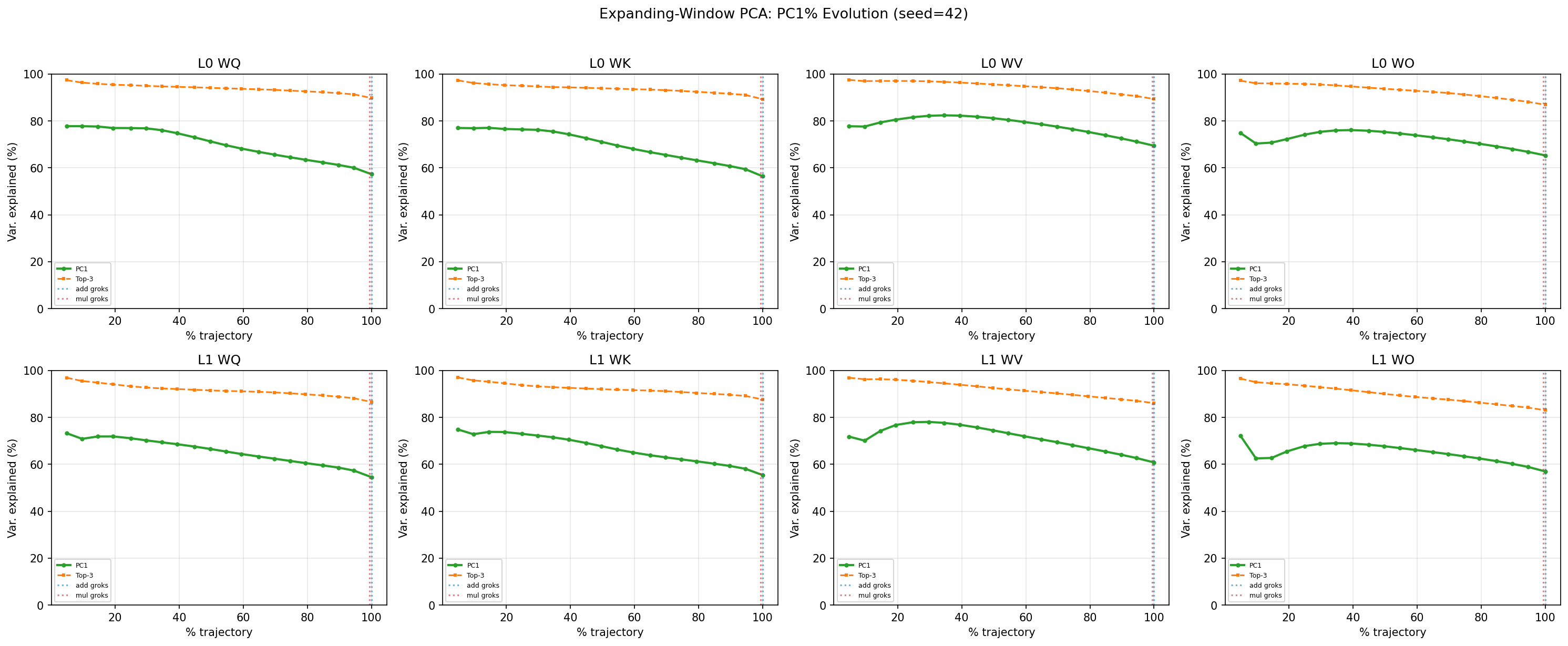}
        \caption{Dual-task expanding-window PC1\% (seed~42): declines over training.}
        \label{fig:dual_expanding}
    \end{subfigure}
    \hfill
    \begin{subfigure}[t]{0.48\textwidth}
        \centering
        \includegraphics[width=\textwidth]{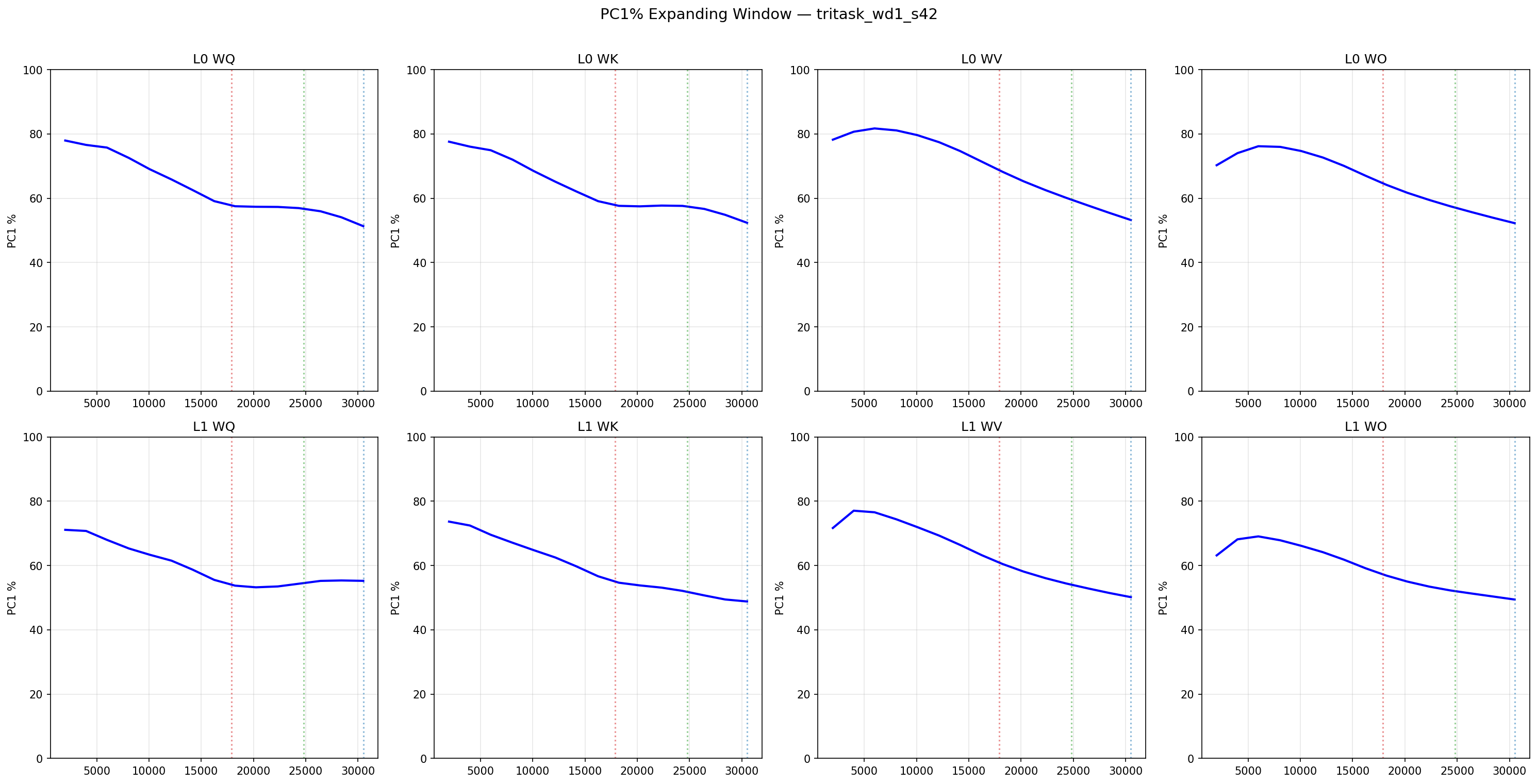}
        \caption{Tri-task expanding-window PC1\%: same declining pattern.}
        \label{fig:tri_expanding}
    \end{subfigure}
    \caption{PC1\% declines over training in multi-task settings, unlike single-task grokking where concentration increases.}
    \label{fig:expanding_pca}
\end{figure}

\paragraph{PC1\% declines during training.}
Expanding-window PCA shows PC1\% starting at ${\sim}80\%$ early in training and declining to ${\sim}50$--60\% by the end (\Cref{fig:expanding_pca}).
This is the opposite of single-task behavior (where concentration increases), reflecting trajectory diversification as the model learns to serve multiple circuits.

\begin{figure}[t]
    \centering
    \includegraphics[width=0.6\textwidth]{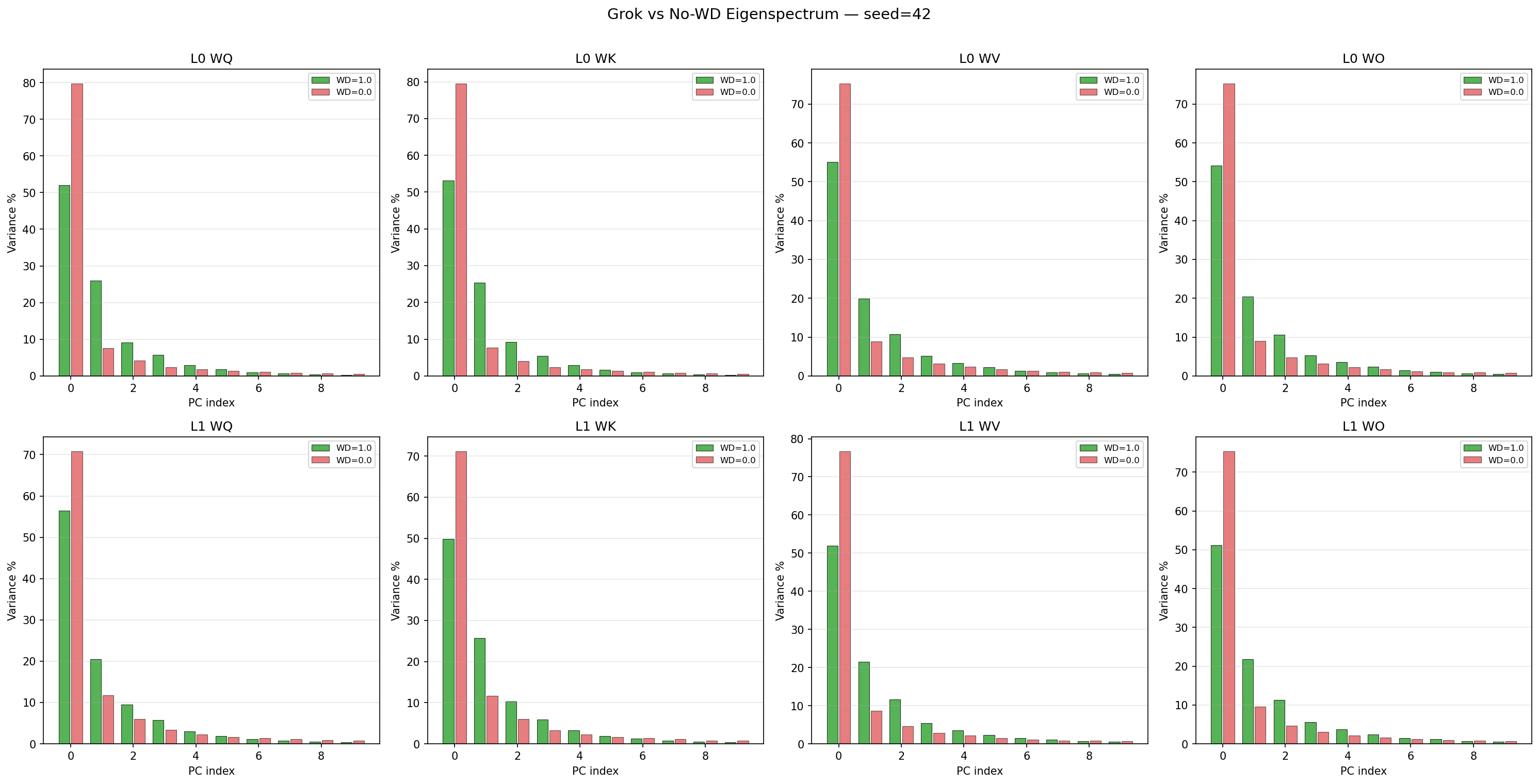}
    \caption{Grok (WD=1.0) vs.\ no-WD (WD=0.0) eigenspectra for tri-task (seed~42). No-WD has higher PC1\%, consistent with a simpler memorization trajectory.}
    \label{fig:tri_comparison}
\end{figure}

\paragraph{No-WD models have higher PC1\%.}
Counterintuitively, the no-WD control ($\lambda = 0$) produces \emph{higher} PC1\% (${\sim}70$--80\%) than grokking models (${\sim}50$--55\%) in the tri-task setting (\Cref{fig:tri_comparison}).
Without weight decay, the model memorizes along a simpler, lower-rank trajectory.
Grokking forces the model into a higher-rank manifold to accommodate generalizing solutions for all tasks.

\begin{figure}[t]
    \centering
    \begin{subfigure}[t]{0.48\textwidth}
        \centering
        \includegraphics[width=\textwidth]{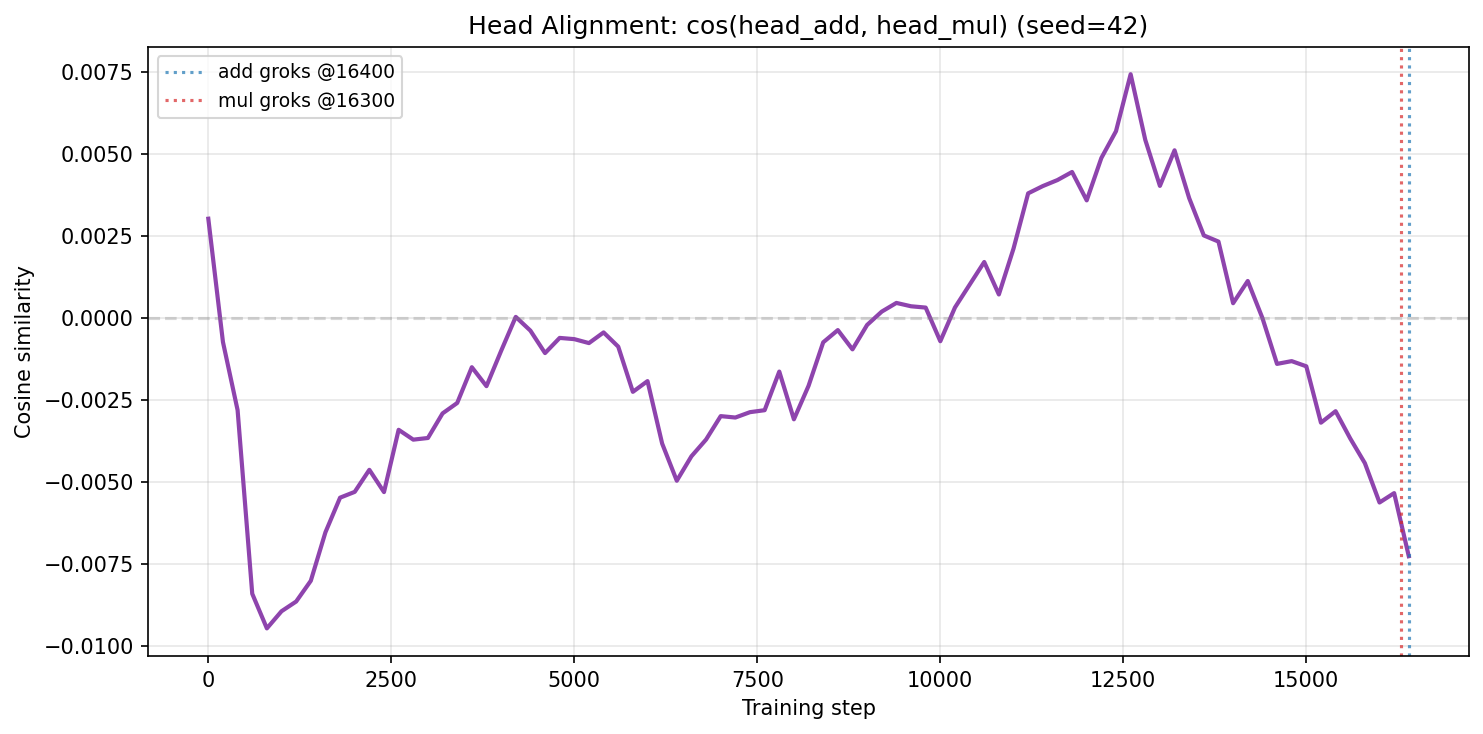}
        \caption{Dual-task: cos(head$_\text{add}$, head$_\text{mul}$) stays near zero.}
        \label{fig:dual_heads}
    \end{subfigure}
    \hfill
    \begin{subfigure}[t]{0.48\textwidth}
        \centering
        \includegraphics[width=\textwidth]{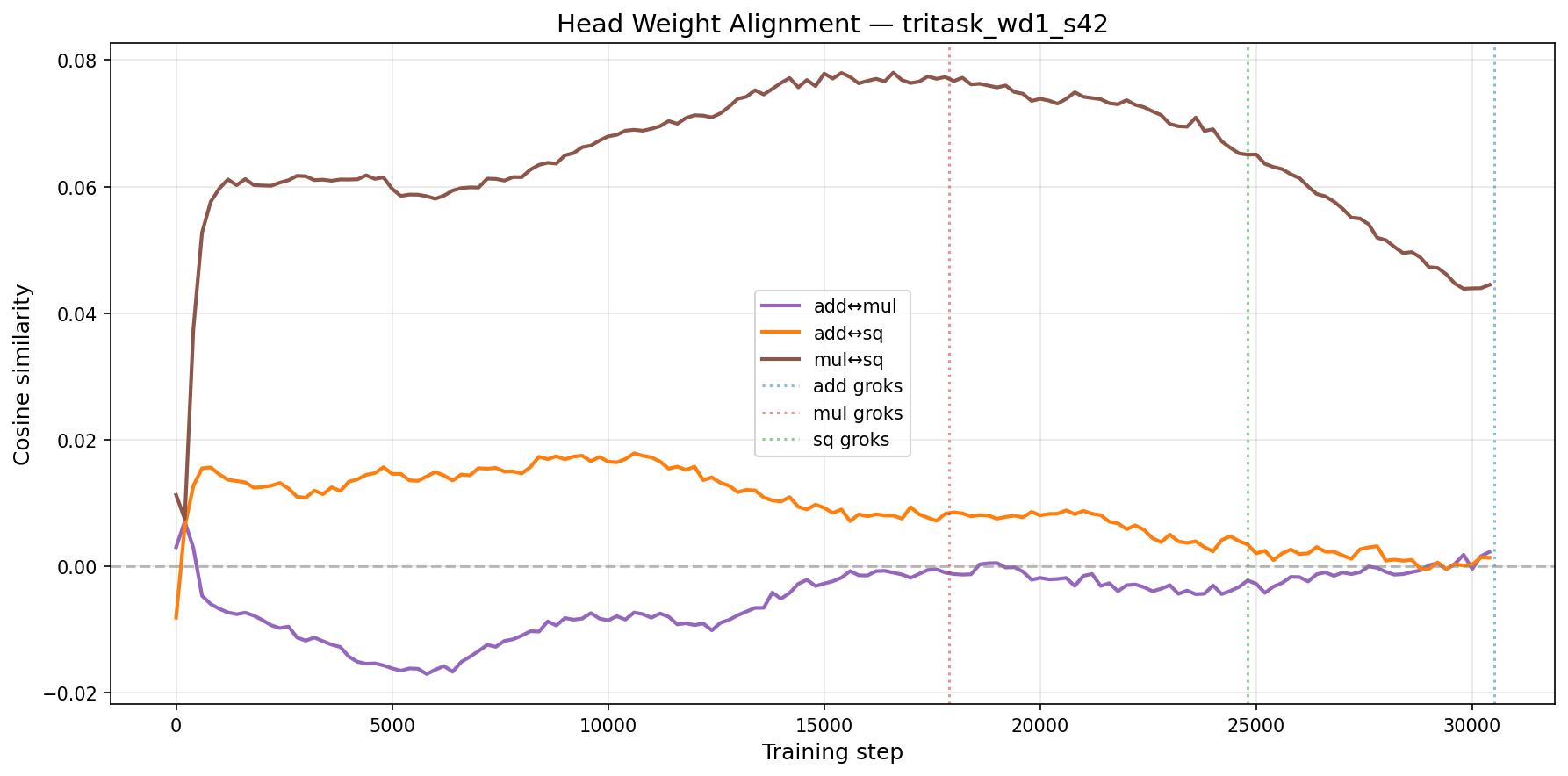}
        \caption{Tri-task: all three pairwise cosines stay near zero.}
        \label{fig:tri_heads}
    \end{subfigure}
    \caption{Task-specific head weights are nearly orthogonal. The shared trunk learns a representation where task readouts are geometrically separated.}
    \label{fig:head_alignment}
\end{figure}

\paragraph{Head weights are nearly orthogonal.}
The pairwise cosine similarity between task-specific classification head weights remains near zero throughout training ($|\cos| < 0.08$; \Cref{fig:head_alignment}).
In the dual-task setting, the two heads converge toward exact orthogonality by the end of training.
In the tri-task setting, all three pairs (add--mul, add--sq, mul--sq) stay within $[-0.01, +0.008]$, confirming that the tasks learn geometrically separated readout directions.

\section{Empirical Integrability Across Tasks and Weight Decay}
\label{sec:integ}

\begin{figure}[t]
    \centering
    \begin{subfigure}[t]{0.48\textwidth}
        \centering
        \includegraphics[width=\textwidth]{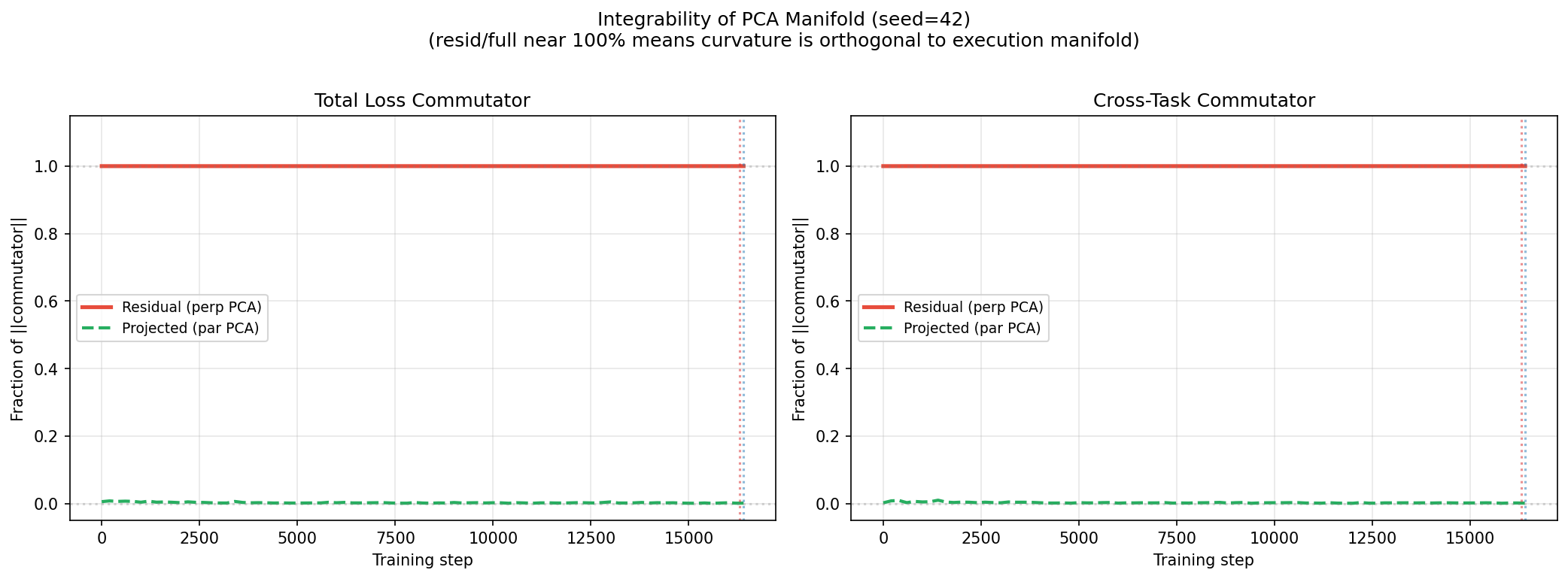}
        \caption{Dual-task (WD=1.0, seed~42): $\rho \approx 1.000$ at every checkpoint.}
        \label{fig:dual_integ}
    \end{subfigure}
    \hfill
    \begin{subfigure}[t]{0.48\textwidth}
        \centering
        \includegraphics[width=\textwidth]{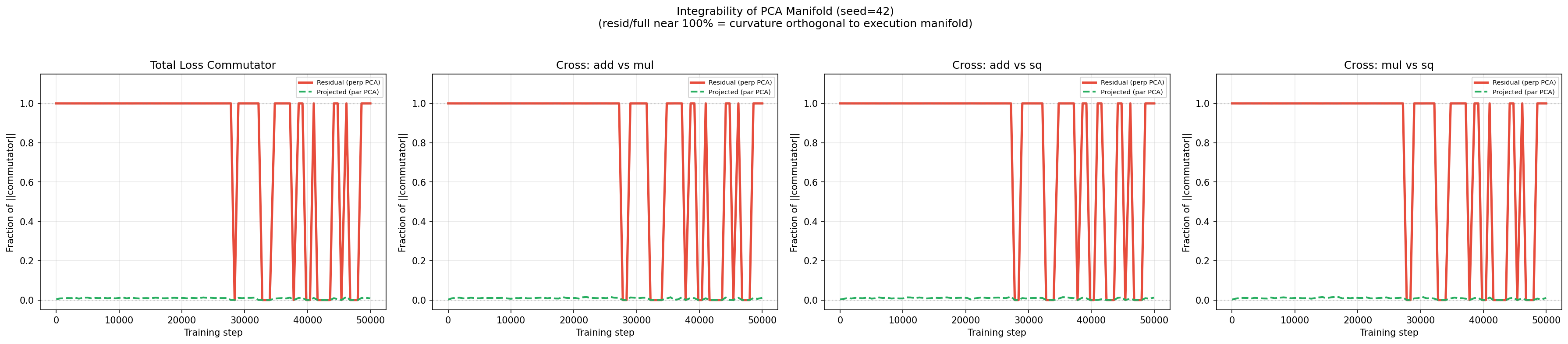}
        \caption{Tri-task (WD=1.0, seed~42): $\rho \approx 1.000$, identical to dual-task.}
        \label{fig:tri_integ}
    \end{subfigure}
    \caption{The execution manifold is empirically integrable in multi-task settings. The invariance measure $\rho \approx 1.000$ across all tested conditions, extending the single-task result.}
    \label{fig:integ_multi}
\end{figure}

We compute commutator defects and project them onto the PCA basis at regular checkpoints, measuring the invariance ratio $\rho = \norm{\delta_\perp} / \norm{\delta}$.
The central result: $\rho \approx 1.000$ within numerical precision across \emph{all} multi-task conditions---both dual-task and tri-task, all weight decay values where defect is non-zero, and all three seeds (\Cref{fig:integ_multi}).

This empirical integrability holds consistently regardless of:
\begin{itemize}
    \item \textbf{Task count}: 1-task (companion paper), 2-task, and 3-task settings all give $\rho \approx 1.000$.
    \item \textbf{Weight decay}: $\lambda \in \{0.1, 0.2, 0.3, 0.5, 1.0\}$ all give $\rho \approx 1.000$ for both total-loss and cross-task commutators.
    \item \textbf{Cross-task commutators}: Using different task losses for the two batches (e.g., add-loss for batch~$A$, mul-loss for batch~$B$) also yields $\rho \approx 1.000$.
\end{itemize}

\begin{figure}[t]
    \centering
    \begin{subfigure}[t]{0.48\textwidth}
        \centering
        \includegraphics[width=\textwidth]{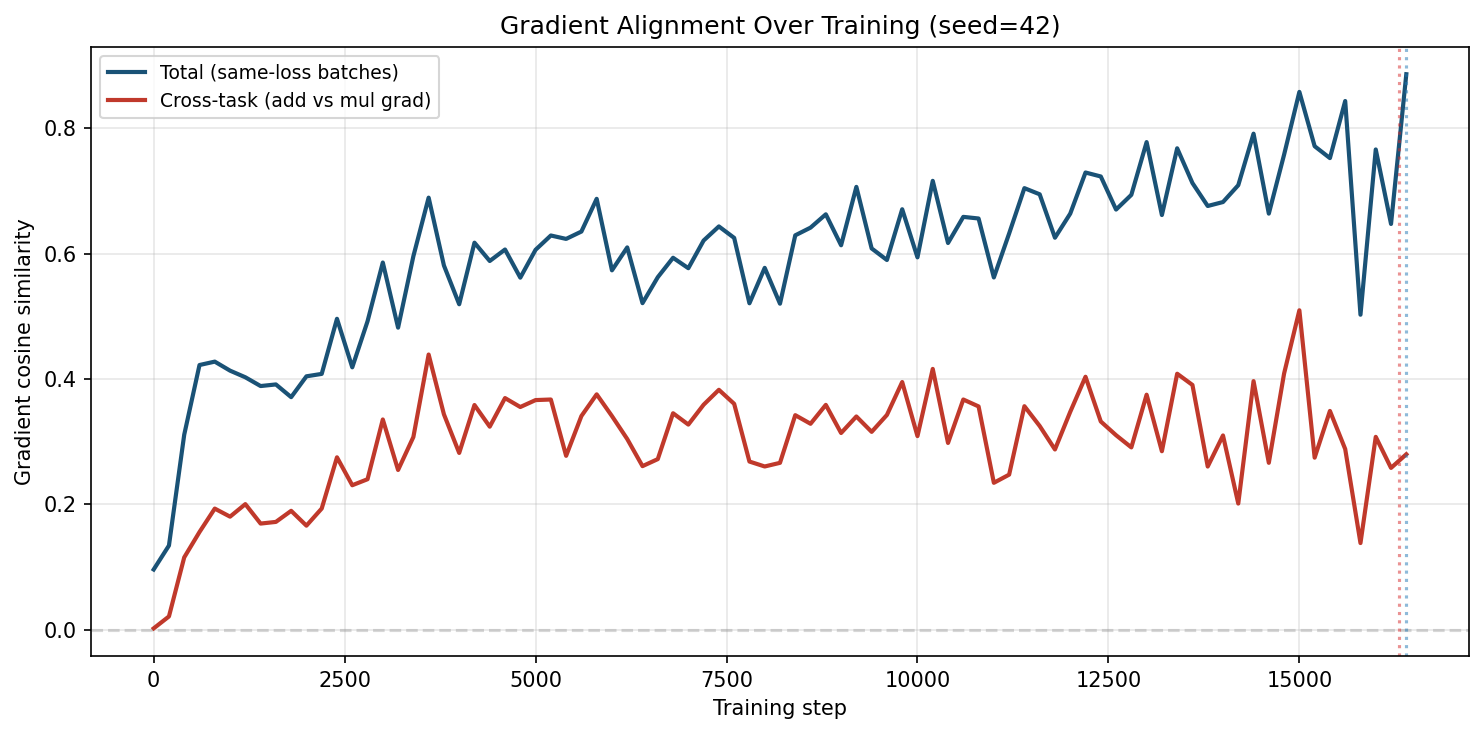}
        \caption{Dual-task gradient cosine similarity (seed~42, WD=1.0): same-task ${\sim}0.8$, cross-task ${\sim}0.3$.}
        \label{fig:dual_gradcos}
    \end{subfigure}
    \hfill
    \begin{subfigure}[t]{0.48\textwidth}
        \centering
        \includegraphics[width=\textwidth]{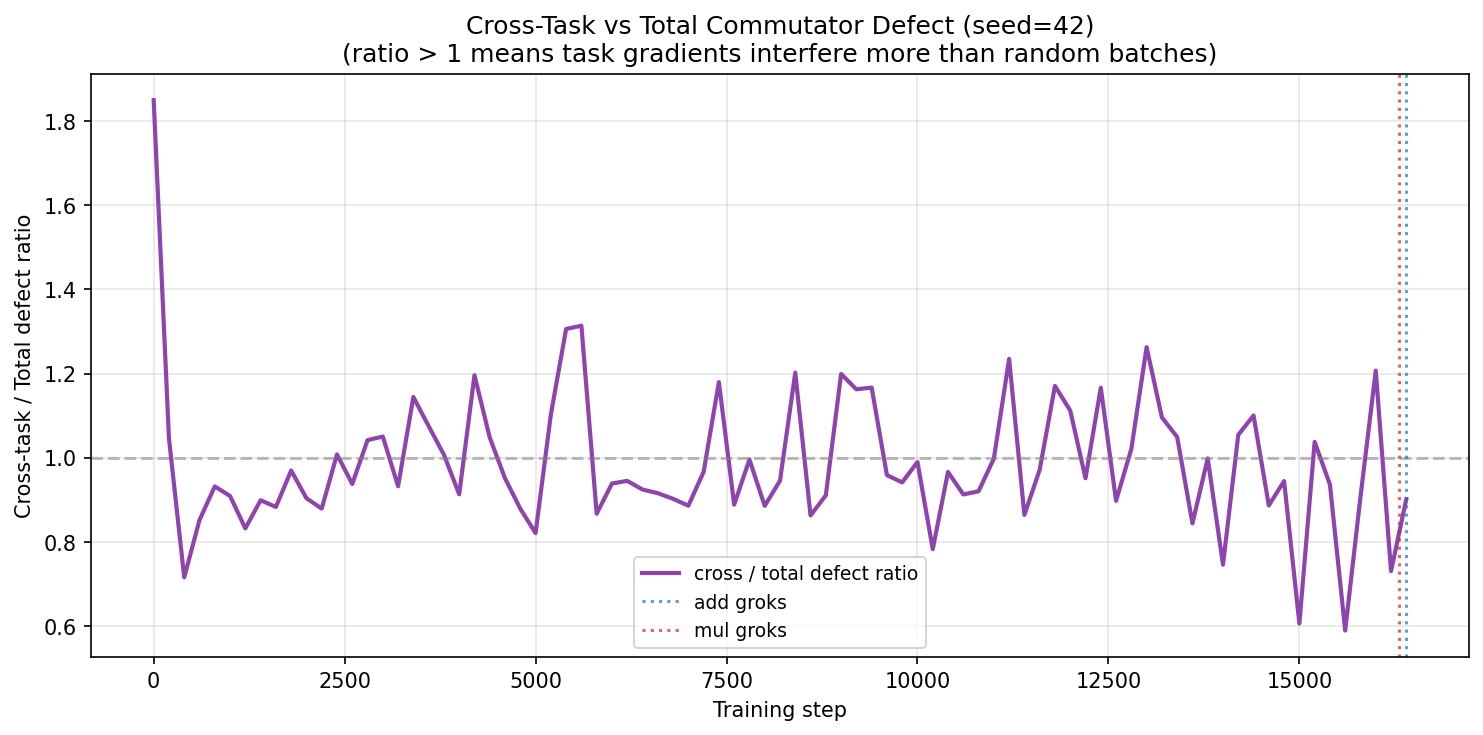}
        \caption{Cross-task vs.\ total defect (seed~42): both have comparable magnitude.}
        \label{fig:cross_total}
    \end{subfigure}
    \caption{Cross-task gradient structure. \textbf{(a)}~Same-task cosines are high (${\sim}0.8$), cross-task cosines are moderate (${\sim}0.3$). \textbf{(b)}~Cross-task defect has roughly the same magnitude as total-loss defect.}
    \label{fig:grad_cos}
\end{figure}

\paragraph{Cross-task gradient structure.}
While integrability is consistent across all tested regimes, the cross-task gradient structure varies with weight decay (\Cref{fig:grad_cos}).
At $\lambda = 1.0$, same-task gradient cosine similarity is 0.77--0.84 and cross-task cosine is 0.24--0.38, indicating moderate alignment.
At $\lambda = 0.5$, cross-task cosines collapse to near zero or even negative values ($-0.17$ to $+0.18$), suggesting the tasks compete for representational capacity at the grokking transition.
At $\lambda = 0.1$, both same-task (0.14--0.49) and cross-task cosines are low.

\section{Hessian Eigenvalue Analysis: Saddle-Mediated Transitions}
\label{sec:hessian}

To probe the loss landscape curvature directly \citep{li2018visualizing, fort2019large}, we compute the bottom eigenvalue of the Hessian (via power iteration on $-\nabla^2 \mathcal{L}$) at regular checkpoints for each task individually and for the total loss.

\begin{figure}[t]
    \centering
    \begin{subfigure}[t]{0.48\textwidth}
        \centering
        \includegraphics[width=\textwidth]{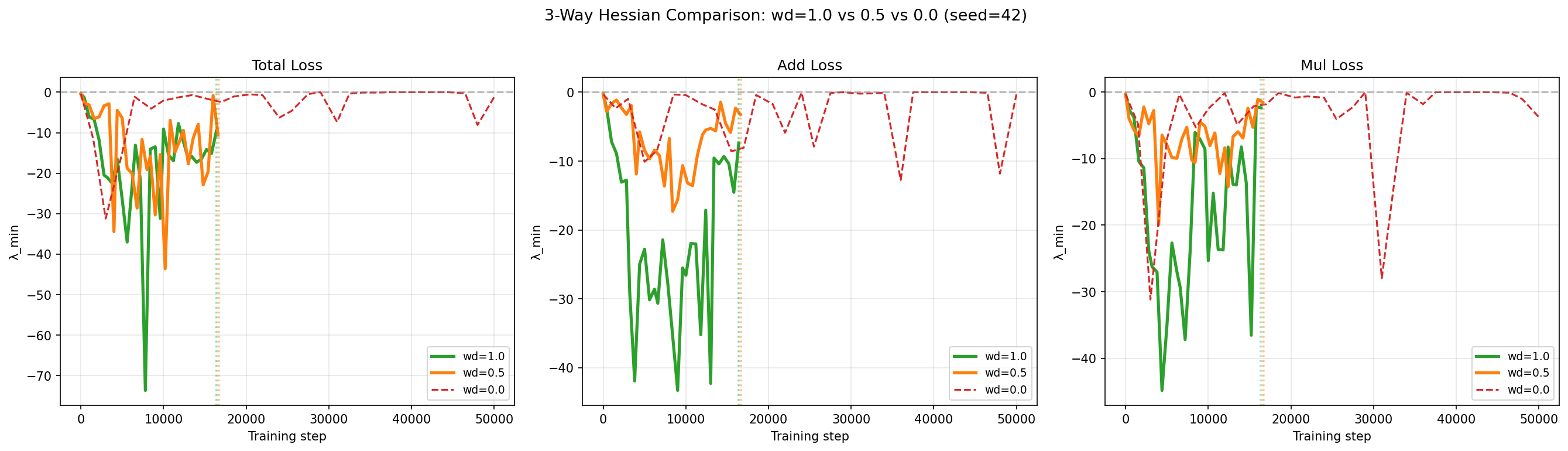}
        \caption{Three-way WD comparison (seed~42): $\lambda = 1.0$ (green) has the deepest negative curvature, $\lambda = 0.0$ (red) stays near zero.}
        \label{fig:hessian_3way}
    \end{subfigure}
    \hfill
    \begin{subfigure}[t]{0.48\textwidth}
        \centering
        \includegraphics[width=\textwidth]{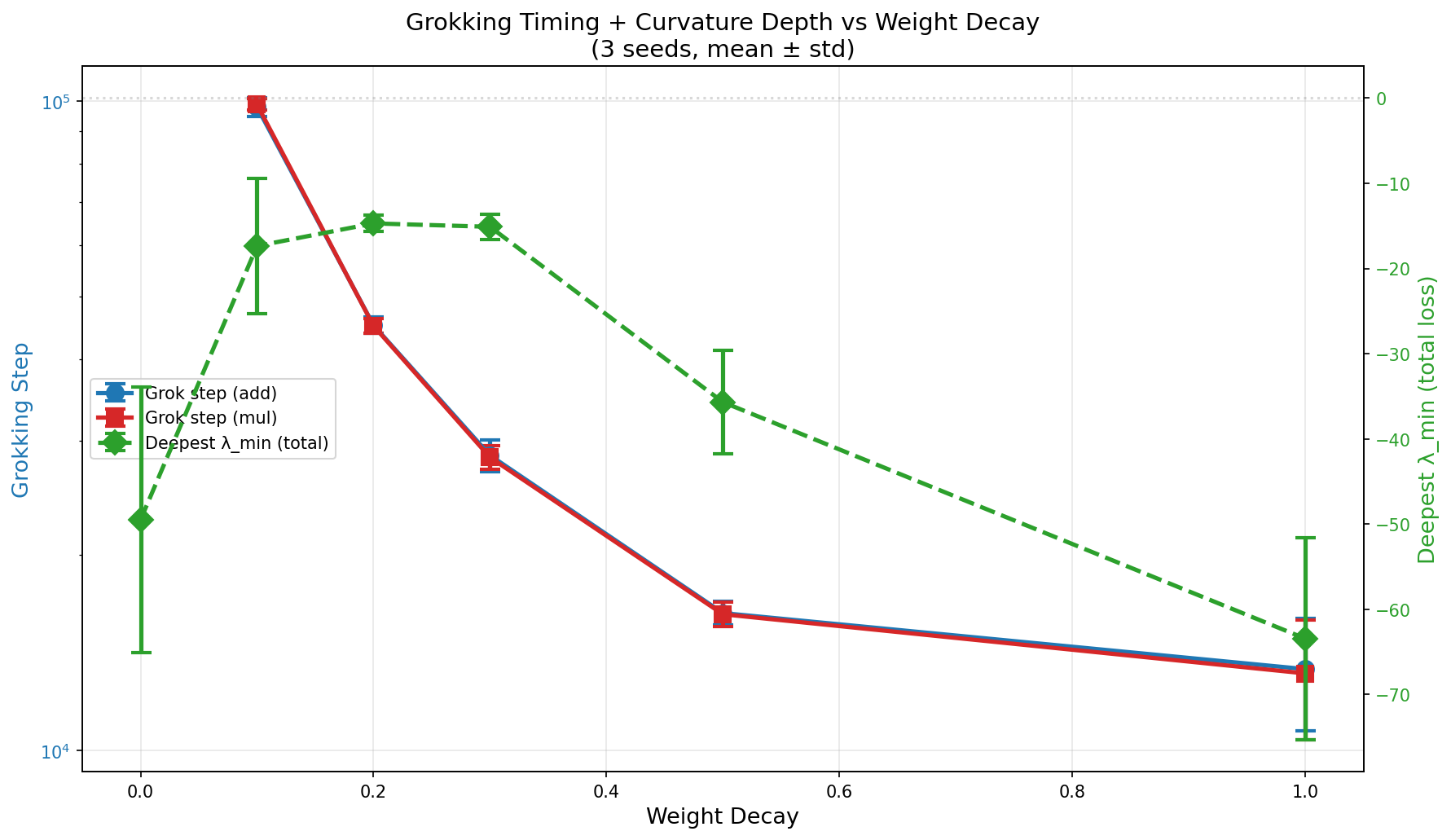}
        \caption{Dual-axis summary: grokking time (blue) and curvature depth (red) anti-correlate across the grokking regime, but $\lambda = 0$ breaks the trend.}
        \label{fig:hessian_dual}
    \end{subfigure}
    \caption{Hessian curvature depth scales with weight decay. \textbf{(a)}~Time series of $\lambda_\text{min}$ for three WD values. \textbf{(b)}~Grok time vs.\ curvature depth across WD, with $\lambda = 0$ as the decoupled outlier.}
    \label{fig:hessian_wd}
\end{figure}

\paragraph{WD controls curvature depth.}
The bottom eigenvalue $\lambda_\text{min}$ scales monotonically with weight decay (\Cref{fig:hessian_wd}).
At $\lambda = 1.0$, the total-loss Hessian reaches $\lambda_\text{min} \approx -63$ (mean across seeds), while at $\lambda = 0.1$, it reaches only $\lambda_\text{min} \approx -17$.
Stronger weight decay drives the loss landscape into deeper saddle regions.

\begin{figure}[t]
    \centering
    \includegraphics[width=0.6\textwidth]{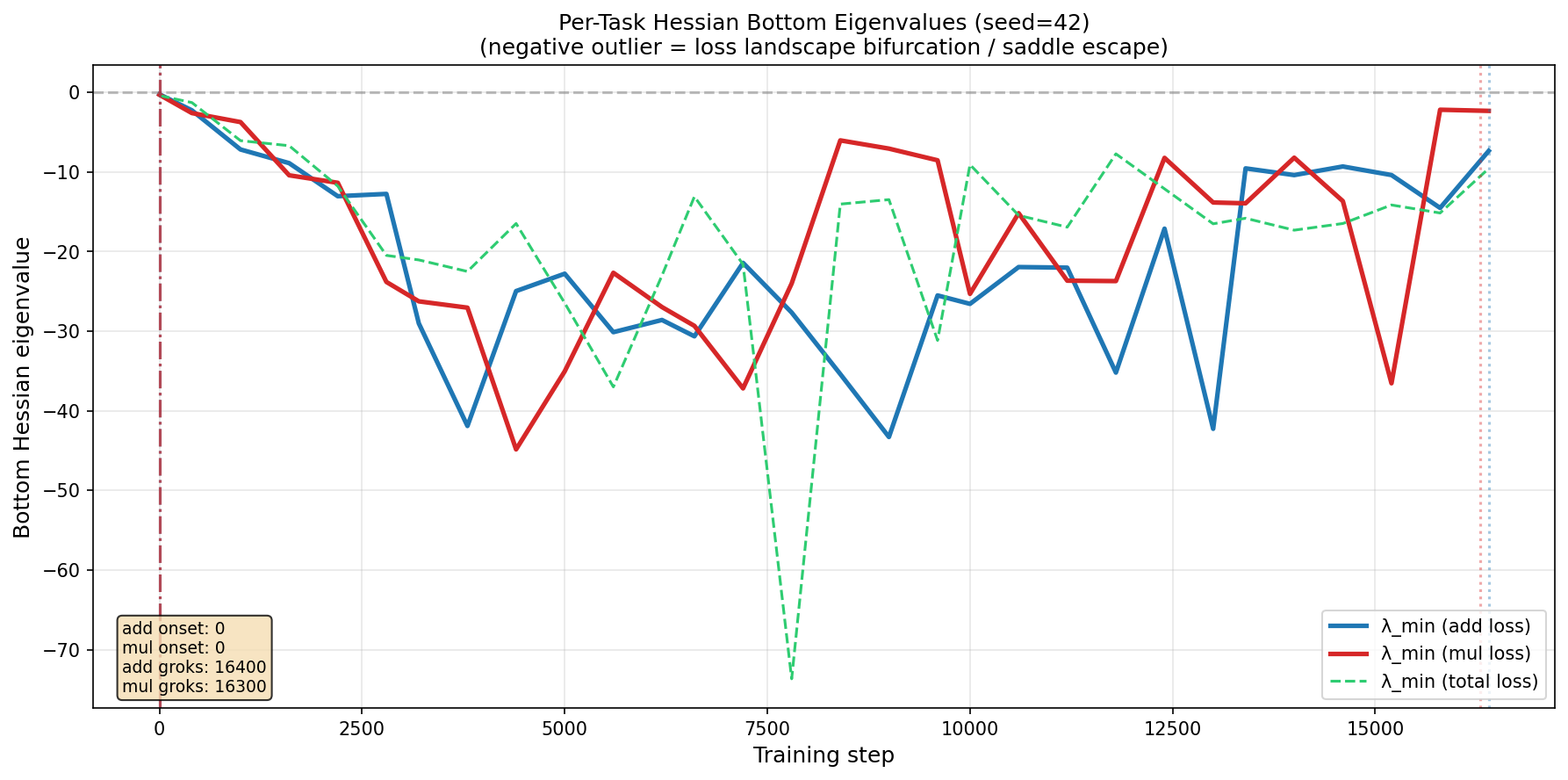}
    \caption{Per-task bottom Hessian eigenvalues (seed~42, WD=1.0). Mul (blue) has deeper negative curvature early and recovers first, consistent with its earlier grokking.}
    \label{fig:pertask_hessian}
\end{figure}

\paragraph{Per-task eigenvalues predict grokking order.}
In the dual-task setting, the task with deeper negative curvature early in training tends to grok first (\Cref{fig:pertask_hessian}).
At seed~42, $\lambda_\text{min}(\text{mul}) = -35.0$ at step~5000 vs.\ $\lambda_\text{min}(\text{add}) = -22.8$ (ratio $1.54\times$), and multiplication groks first.
This pattern holds for 2 of 3 seeds (seed~137 is the exception).
By grokking time, the first-to-grok task's eigenvalue has recovered closer to zero, while the other task is still mid-escape.

\begin{table}[t]
\centering
\caption{Per-task Hessian eigenvalues at mid-training (dual-task, WD=1.0). The task with deeper $\lambda_\text{min}$ tends to grok first (2/3 seeds).}
\label{tab:pertask_hessian}
\begin{tabular}{@{}lrrrl@{}}
\toprule
Seed & $\lambda_\text{min}(\text{mul})$ & $\lambda_\text{min}(\text{add})$ & Ratio & First to grok \\
\midrule
42 & $\mathbf{-35.0}$ & $-22.8$ & 1.54$\times$ & mul \\
137 & $-17.4$ & $\mathbf{-22.9}$ & 0.76$\times$ & mul \\
2024 & $\mathbf{-35.1}$ & $-21.4$ & 1.64$\times$ & mul \\
\bottomrule
\end{tabular}
\end{table}

\paragraph{$\lambda = 0$ has curvature but no grokking.}
The no-WD control shows non-trivial negative curvature ($\lambda_\text{min} \approx -50$ at WD=0 vs.\ $\lambda_\text{min} \approx -63$ at WD=1.0) yet never generalizes (\Cref{tab:hessian_summary}).
This confirms that negative Hessian curvature is \emph{necessary but not sufficient} for grokking \citep[cf.][]{ge2015escaping}: weight decay provides the regularization pressure to escape the memorization saddle and reach a generalizing basin.

\begin{table}[t]
\centering
\caption{Multi-seed Hessian summary (dual-task, 3 seeds $\times$ 4 WD values = 12 runs).}
\label{tab:hessian_summary}
\begin{tabular}{@{}lrrrrr@{}}
\toprule
$\lambda$ & Grok (Add) & Grok (Mul) & $\lambda_\text{min}$(total) & $\lambda_\text{min}$(add) & $\lambda_\text{min}$(mul) \\
\midrule
1.0 & $13{,}333 \pm 2{,}620$ & $13{,}133 \pm 2{,}748$ & $-63.4 \pm 11.9$ & $-50.8 \pm 5.6$ & $-43.8 \pm 2.2$ \\
0.5 & $16{,}267 \pm 685$ & $16{,}200 \pm 712$ & $-35.7 \pm 6.1$ & $-21.2 \pm 3.0$ & $-33.8 \pm 18.4$ \\
0.1 & $97{,}967 \pm 3{,}172$ & $98{,}833 \pm 2{,}001$ & $-17.4 \pm 7.9$ & $-11.8 \pm 0.7$ & $-11.8 \pm 4.2$ \\
0.0 & --- & --- & $-49.5 \pm 15.6$ & $-53.8 \pm 36.3$ & $-40.7 \pm 12.9$ \\
\bottomrule
\end{tabular}
\end{table}

\begin{figure}[t]
    \centering
    \begin{subfigure}[t]{0.48\textwidth}
        \centering
        \includegraphics[width=\textwidth]{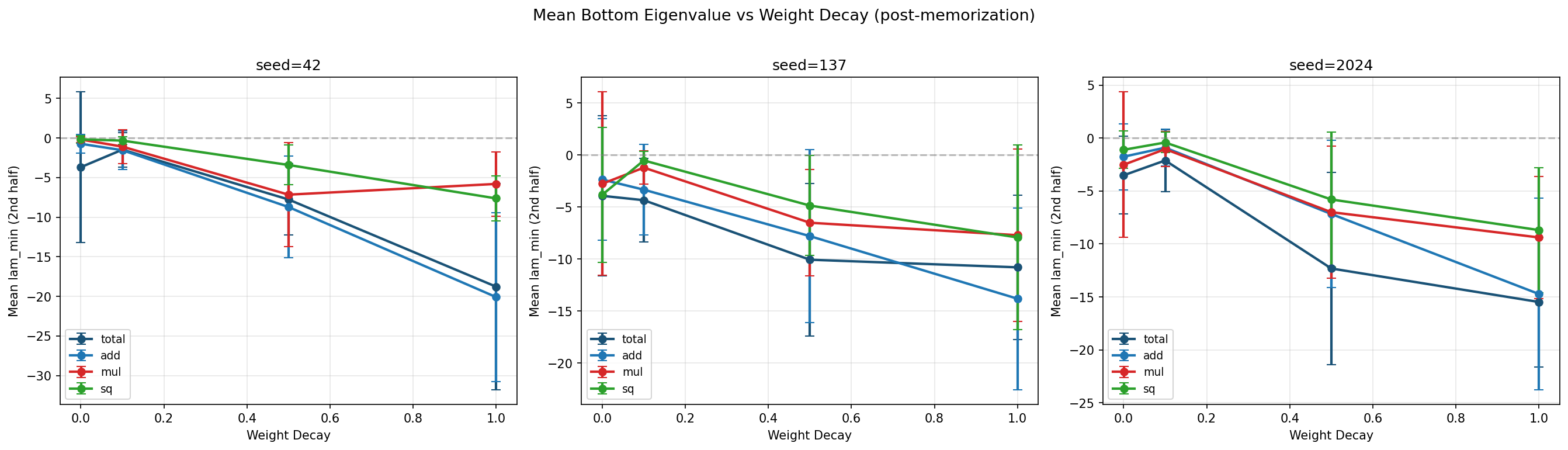}
        \caption{Tri-task: bottom eigenvalue vs.\ WD at different training fractions. Curvature depth increases monotonically with WD.}
        \label{fig:tri_wd_eig}
    \end{subfigure}
    \hfill
    \begin{subfigure}[t]{0.48\textwidth}
        \centering
        \includegraphics[width=\textwidth]{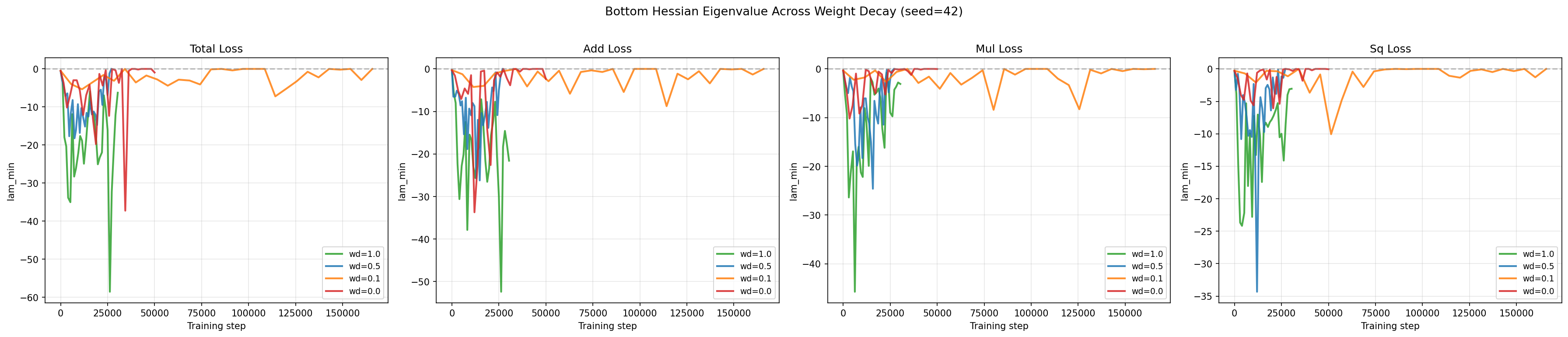}
        \caption{Tri-task cross-WD time series (seed~42): $\lambda = 1.0$ (green) occupies the most negative territory.}
        \label{fig:tri_wd_traces}
    \end{subfigure}
    \caption{Tri-task Hessian analysis replicates the dual-task pattern: stronger WD drives deeper negative curvature.}
    \label{fig:tri_hessian_wd}
\end{figure}

\paragraph{Tri-task Hessian.}
The tri-task Hessian analysis replicates the dual-task patterns.
At $\lambda = 1.0$, the total-loss bottom eigenvalue reaches $-10$ to $-19$; at $\lambda = 0.1$, it reaches only $-1.5$ to $-4.3$.
The WD-to-curvature monotonic relationship is preserved, and cross-WD comparisons (\Cref{fig:tri_hessian_wd}) show the same qualitative structure as dual-task.

\section{Weight Decay Phase Diagram}
\label{sec:wd_phase}

Collecting results across five WD values reveals a structured phase diagram governing grokking dynamics.

\begin{figure}[t]
    \centering
    \includegraphics[width=0.7\textwidth]{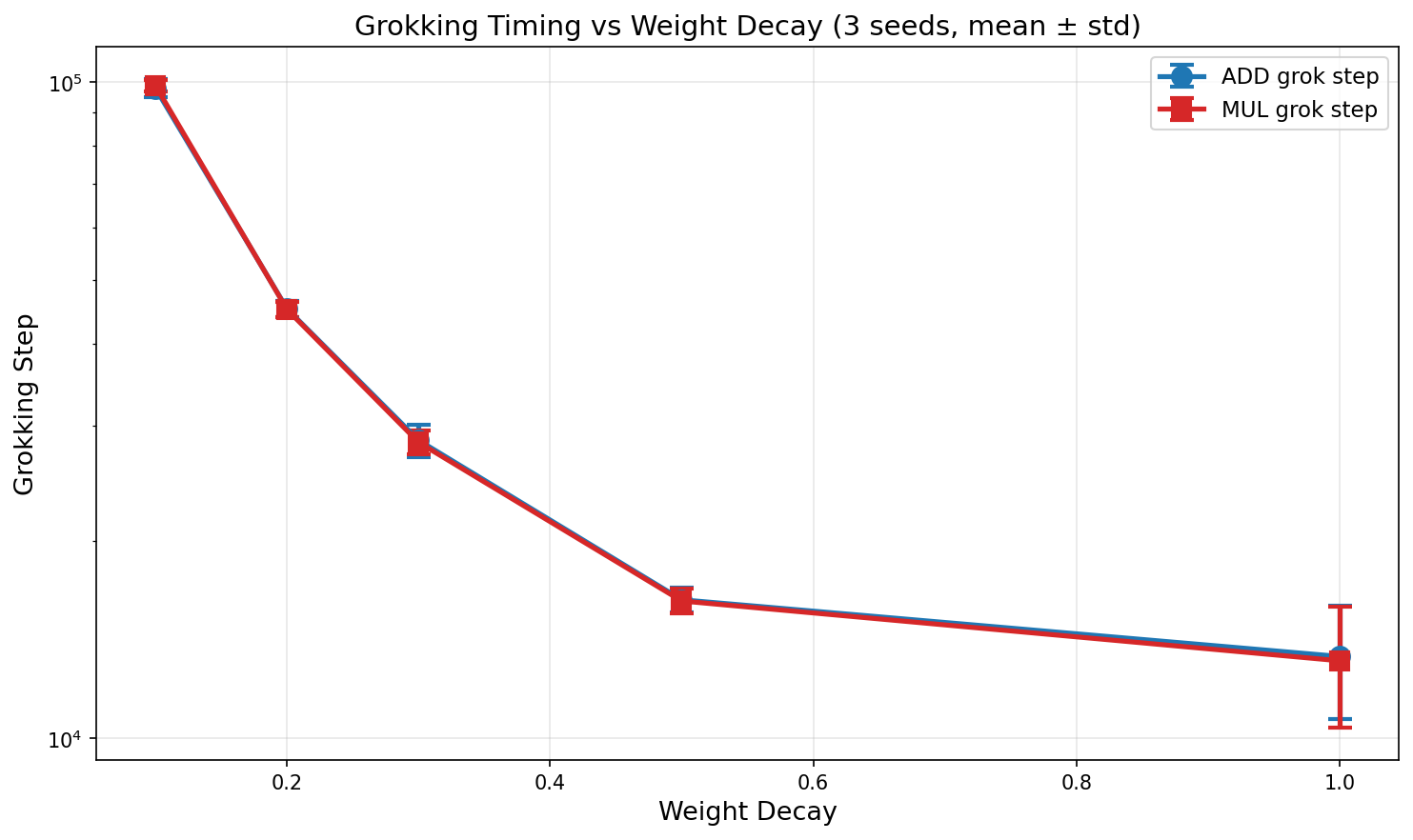}
    \caption{Grokking timescale vs.\ weight decay (dual-task, 3 seeds, log scale). The relationship is nearly log-linear for $\lambda > 0$, with a sharp jump between $\lambda = 0.5$ and $\lambda = 0.1$.}
    \label{fig:wd_timing}
\end{figure}

\paragraph{Grokking timescale.}
Grokking time scales roughly as $t_\text{grok} \propto \lambda^{-\alpha}$, with the relationship appearing nearly log-linear on the WD axis for $\lambda > 0$ (\Cref{fig:wd_timing}).
Two regimes emerge:
\begin{itemize}
    \item $\lambda \geq 0.5$: fast grokking (13--16k steps), with curvature depth scaling strongly ($-36$ to $-63$);
    \item $\lambda \leq 0.3$: slow grokking (28--98k steps), with curvature depth plateauing around $-15$.
\end{itemize}
The transition between regimes occurs near $\lambda \approx 0.3$, suggesting a critical weight decay value.

\paragraph{Defect onset lead time (dual-task).}
We measure defect onset lead time in the dual-task setting across all five WD values and three seeds (15 conditions total).
Across all 15 dual-task conditions, commutator defect onset \emph{always} precedes grokking (\Cref{tab:defect_onset}).
However, the lead fraction follows a V-shaped curve with a minimum at $\lambda = 0.5$ (4.9\%), where defect onset and grokking nearly coincide.
At lower WD, grok time stretches out much faster than defect onset, creating long incubation periods (lead fraction up to 70\% at $\lambda = 0.1$).

\begin{table}[t]
\centering
\caption{Defect onset lead over grokking (dual-task, mean across 3 seeds per WD).}
\label{tab:defect_onset}
\begin{tabular}{@{}rrrrr@{}}
\toprule
$\lambda$ & Mean Grok Step & Mean Onset Step & Mean Lead (steps) & Lead \% \\
\midrule
1.0 & 13,133 & 6,600 & 6,533 & 52.9\% \\
0.5 & 16,200 & 15,400 & 800 & 4.9\% \\
0.3 & 28,200 & 19,133 & 9,067 & 32.0\% \\
0.2 & 45,067 & 17,667 & 27,400 & 60.8\% \\
0.1 & 97,933 & 29,467 & 68,467 & 69.9\% \\
\bottomrule
\end{tabular}
\end{table}

\begin{figure}[t]
    \centering
    \begin{subfigure}[t]{0.48\textwidth}
        \centering
        \includegraphics[width=\textwidth]{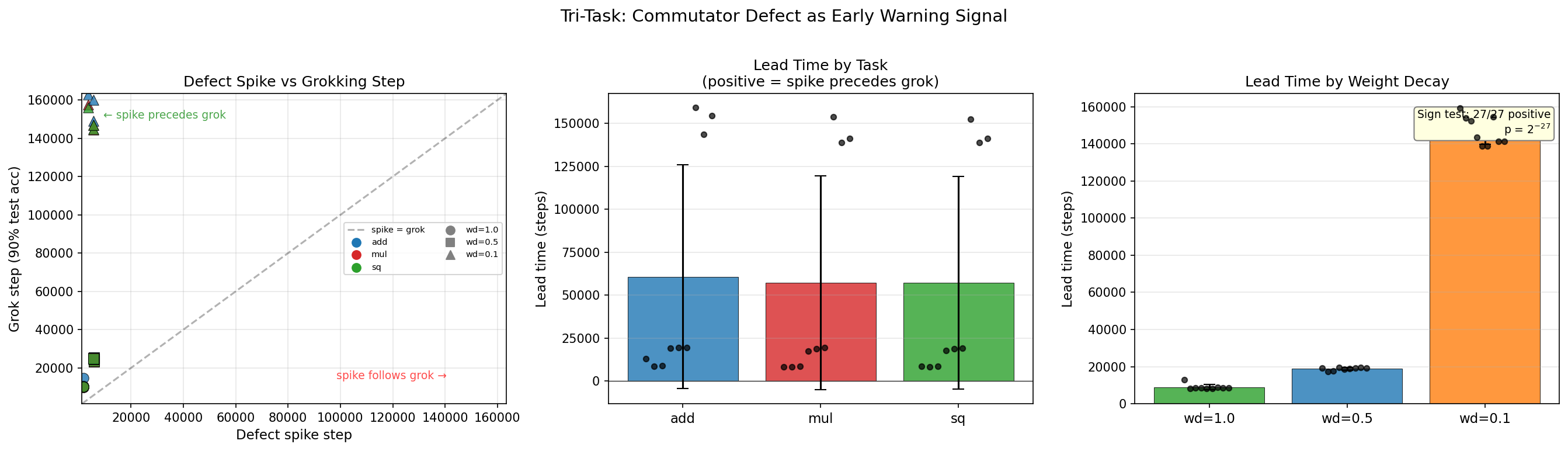}
        \caption{Tri-task lead-time analysis.
        Left: spike step vs.\ grok step scatter (all 27 points above the diagonal).
        Center: lead time by task (all three tasks show similar distributions).
        Right: lead time by WD (monotonically increasing).}
        \label{fig:tritask_lead}
    \end{subfigure}
    \hfill
    \begin{subfigure}[t]{0.48\textwidth}
        \centering
        \includegraphics[width=\textwidth]{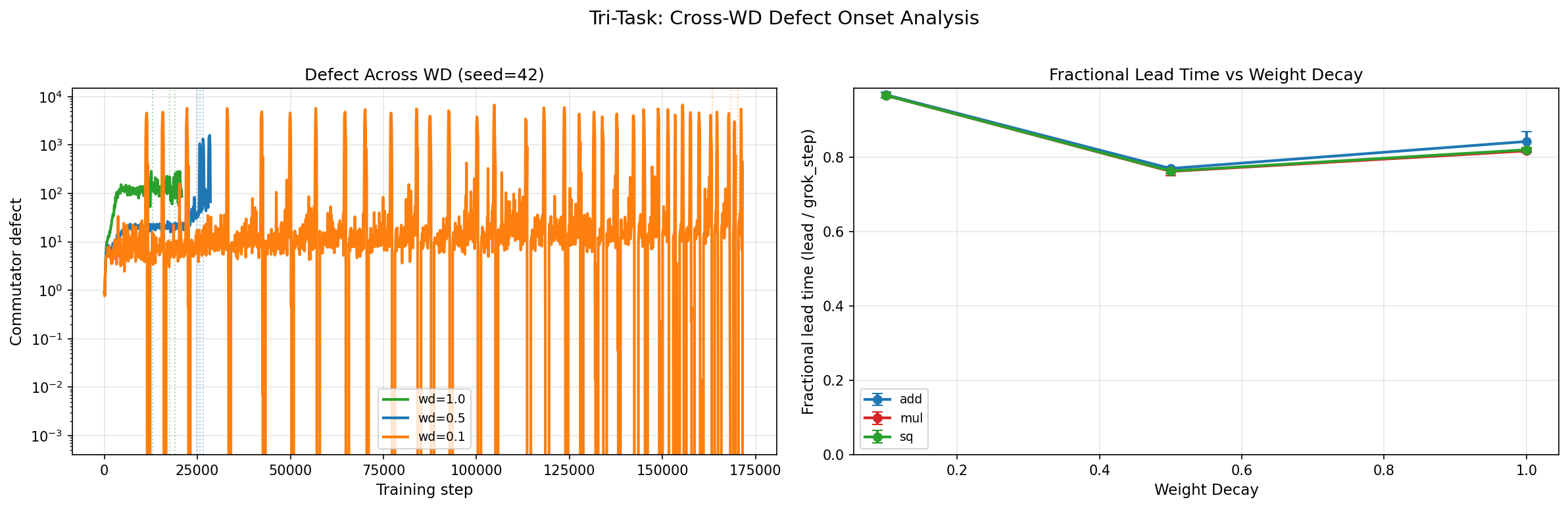}
        \caption{Left: defect time-series across WD (seed~42), showing the defect explosion at different timescales. Right: fractional lead time vs.\ WD---monotonically increasing for all three tasks.}
        \label{fig:tritask_lead_wd}
    \end{subfigure}
    \caption{Tri-task defect onset always precedes grokking ($27/27$, $p = 2^{-27}$). Unlike the dual-task V-shape, the tri-task lead fraction increases monotonically with decreasing WD, reaching ${\sim}97\%$ at $\lambda = 0.1$.}
    \label{fig:tritask_defect_lead}
\end{figure}

\paragraph{Defect onset lead time (tri-task).}
We extend the defect onset analysis to the tri-task setting, training fresh models with inline commutator defect measurement every 100 steps across $\lambda \in \{1.0, 0.5, 0.1\}$ and three seeds (27 task$\times$seed$\times$WD conditions), plus $\lambda = 0$ controls (\Cref{fig:tritask_lead}).
In all 27 grokking conditions, defect onset precedes generalization (sign test: $27/27$ positive, $p = 2^{-27} \approx 7.5 \times 10^{-9}$).
Per-task lead times are remarkably uniform: the mean lead is $60{,}622$ steps for addition, $57{,}144$ for multiplication, and $57{,}078$ for the quadratic sum, indicating that the defect spike is a \emph{shared} signal of the entire model approaching the grokking transition, rather than a task-specific indicator.

The tri-task lead fraction is monotonically increasing with decreasing WD (\Cref{tab:tritask_onset,fig:tritask_lead_wd}), in contrast to the dual-task V-shaped curve:
\begin{itemize}
    \item $\lambda = 1.0$: mean lead $= 8{,}900$ steps, lead fraction $\approx 82\%$;
    \item $\lambda = 0.5$: mean lead $= 18{,}800$ steps, lead fraction $\approx 77\%$;
    \item $\lambda = 0.1$: mean lead $= 147{,}144$ steps, lead fraction $\approx 97\%$.
\end{itemize}
The defect spike occurs very early in training relative to grokking at all WD values, and the relative earliness \emph{increases} at lower WD---meaning the defect is an even more valuable early warning signal when grokking is slow \citep[see also][for power-law scaling of lead time across task families]{xu2026earlywarning}.
At $\lambda = 0.1$, the spike occurs within the first ${\sim}3\%$ of training, leaving a long incubation period before generalization emerges.

\begin{table}[t]
\centering
\caption{Tri-task defect onset lead over grokking (mean across 3 tasks $\times$ 3 seeds = 9 conditions per WD).}
\label{tab:tritask_onset}
\begin{tabular}{@{}rrrrr@{}}
\toprule
$\lambda$ & Mean Grok Step & Mean Onset Step & Mean Lead (steps) & Lead \% \\
\midrule
1.0 & 12,089 & 1,833 & 8,900 & 82.2\% \\
0.5 & 25,333 & 5,767 & 18,800 & 76.5\% \\
0.1 & 152,356 & 5,000 & 147,144 & 96.7\% \\
\bottomrule
\end{tabular}
\end{table}

\begin{figure}[t]
    \centering
    \begin{subfigure}[t]{0.48\textwidth}
        \centering
        \includegraphics[width=\textwidth]{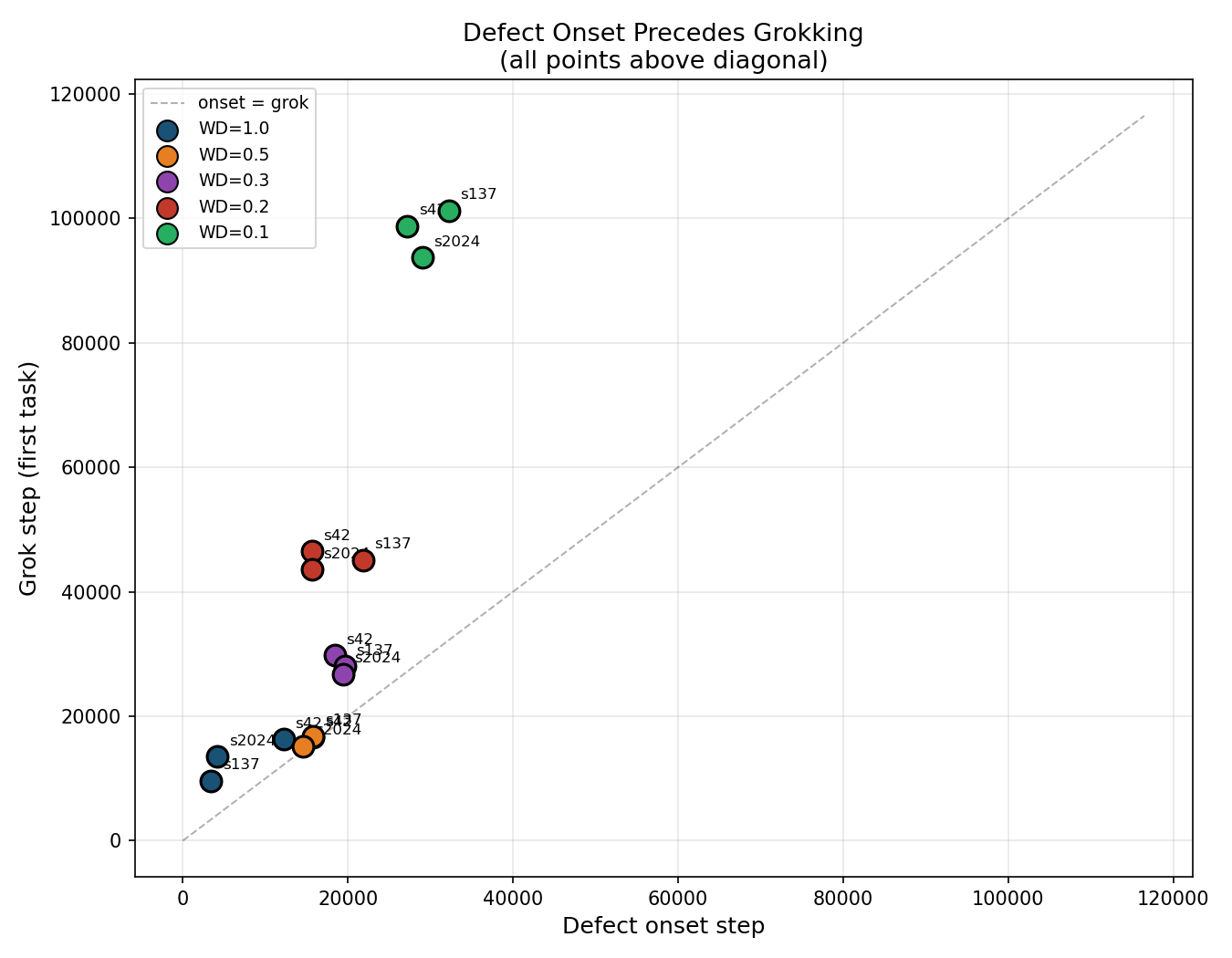}
        \caption{Onset step vs.\ grok step: all 15 points lie above the diagonal (onset precedes grokking in every condition).}
        \label{fig:onset_scatter}
    \end{subfigure}
    \hfill
    \begin{subfigure}[t]{0.48\textwidth}
        \centering
        \includegraphics[width=\textwidth]{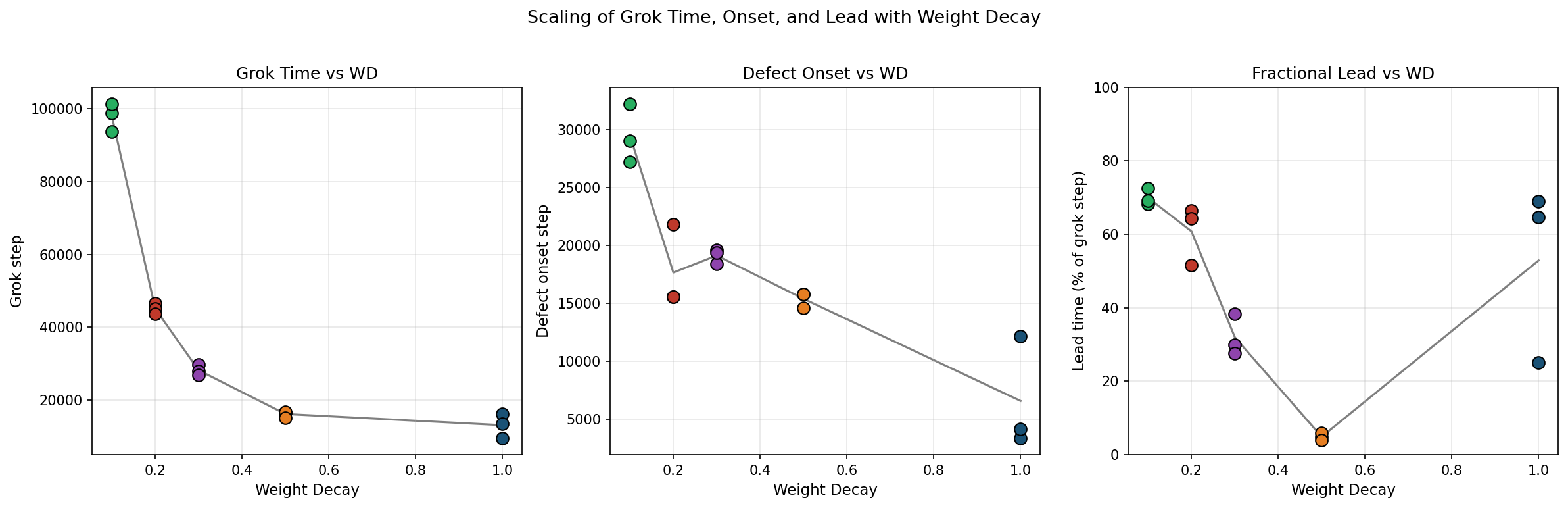}
        \caption{WD scaling: grokking time (left) and lead fraction (right) vs.\ $\lambda$.}
        \label{fig:wd_scaling}
    \end{subfigure}
    \caption{Dual-task defect onset always precedes grokking (15/15 conditions), with a non-monotonic lead fraction (minimum at $\lambda = 0.5$).}
    \label{fig:defect_lead}
\end{figure}

\begin{figure}[t]
    \centering
    \begin{subfigure}[t]{0.48\textwidth}
        \centering
        \includegraphics[width=\textwidth]{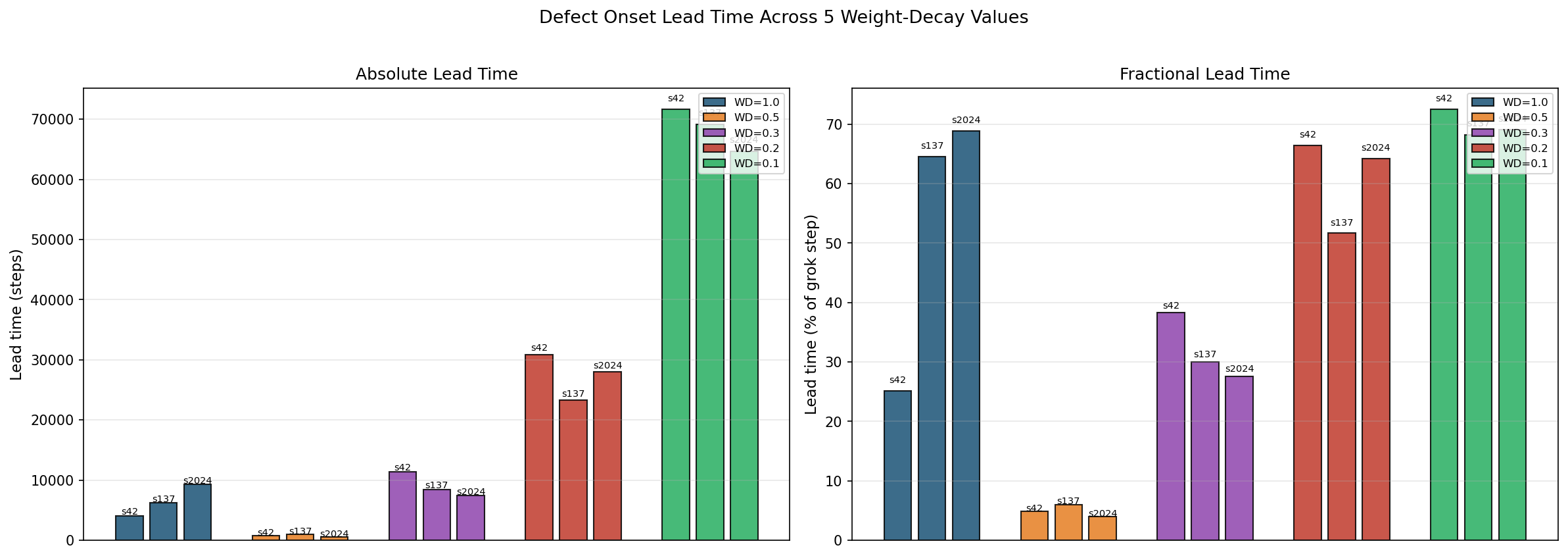}
        \caption{Absolute and fractional lead time by WD (per-seed bars). Fractional lead (right) shows a clear V-shape with minimum at $\lambda = 0.5$.}
        \label{fig:lead_bars}
    \end{subfigure}
    \hfill
    \begin{subfigure}[t]{0.48\textwidth}
        \centering
        \includegraphics[width=\textwidth]{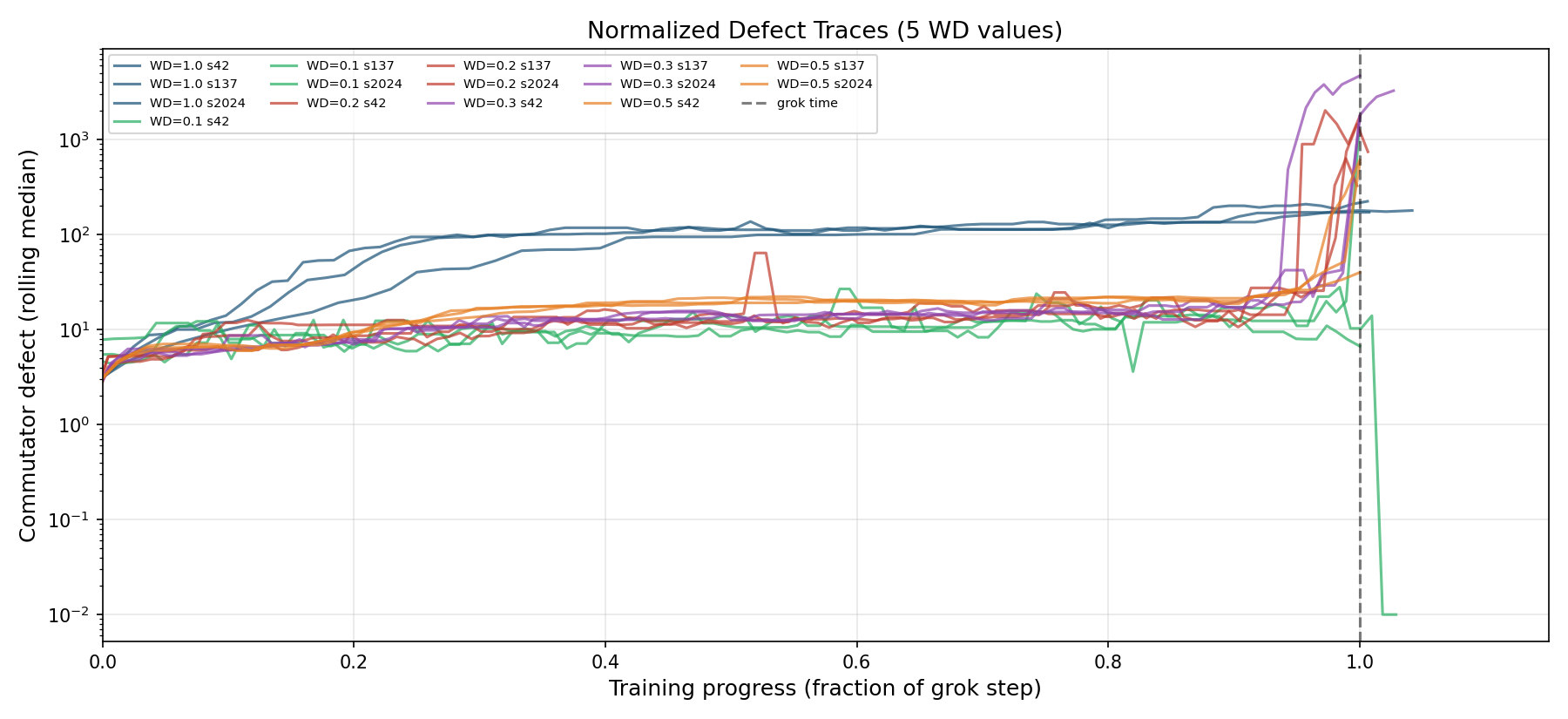}
        \caption{Normalized defect traces (all 15 conditions). Training progress is measured as fraction of grok step; all traces converge near $t/t_\text{grok} = 1$.}
        \label{fig:normalized_traces}
    \end{subfigure}
    \caption{V-shaped lead fraction across weight decay (dual-task only). \textbf{(a)}~Per-seed lead times: $\lambda = 0.5$ has the smallest lead (${\sim}5\%$), with larger leads at both higher and lower WD. \textbf{(b)}~All 15 dual-task defect traces on a common normalized time axis.}
    \label{fig:lead_vshaped}
\end{figure}

\paragraph{V-shaped lead fraction (dual-task) vs.\ monotonic (tri-task).}
The dual-task lead fraction (lead time as a percentage of grok step) follows a striking V-shaped curve across WD (\Cref{fig:lead_vshaped}).
At $\lambda = 1.0$, lead fractions are 25--69\% (mean 53\%); they collapse to ${\sim}5\%$ at $\lambda = 0.5$, then rise monotonically back to ${\sim}70\%$ at $\lambda = 0.1$.
This non-monotonicity is visible per-seed (\Cref{fig:lead_bars}): all three seeds agree that $\lambda = 0.5$ has the smallest lead.
The normalized defect traces (\Cref{fig:normalized_traces}) show all 15 dual-task conditions on a common x-axis (fraction of grok step), making the V-shape visually apparent: the $\lambda = 0.5$ traces (orange) reach the grokking transition earliest in relative time, while the $\lambda = 1.0$ and $\lambda = 0.1$ traces have long incubation periods.

The tri-task setting shows a qualitatively different pattern: the lead fraction is monotonically increasing as WD decreases ($82\% \to 77\% \to 97\%$), without the dual-task's dip at $\lambda = 0.5$ (\Cref{fig:tritask_lead_wd}).
We hypothesize that the additional task constraints in the tri-task setting prevent the ``efficient'' trajectory compression seen at $\lambda = 0.5$ in dual-task; with three tasks competing for shared capacity, the defect onset remains early relative to grokking at all WD values.
This is consistent with the broader pattern that tri-task grokking is more fragile---requiring more of the available parameter-space capacity, leaving less room for geometric shortcuts.

\begin{figure}[t]
    \centering
    \includegraphics[width=\textwidth]{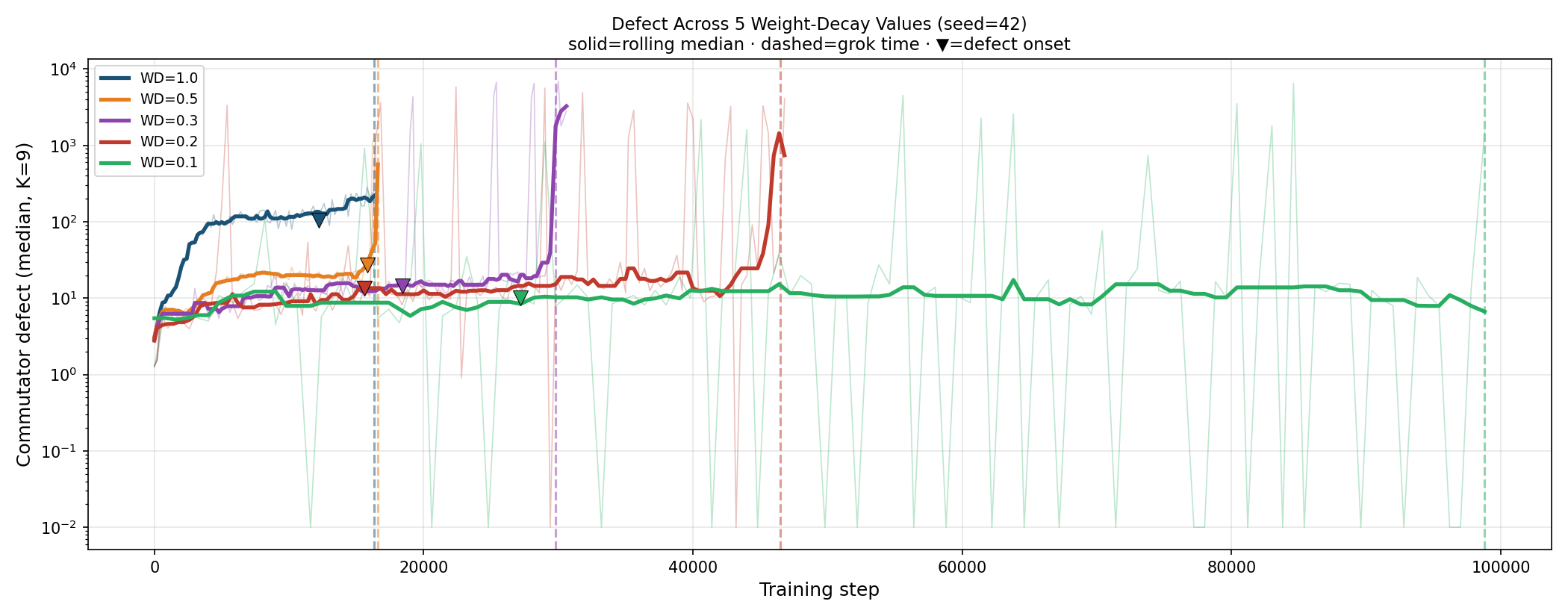}
    \caption{Defect traces across five WD values (seed~42). $\lambda \geq 0.5$: smooth monotonic rise. $\lambda \leq 0.3$: intermittent spiky pattern with transient explosions.}
    \label{fig:defect_traces}
\end{figure}

\paragraph{Defect dynamics vary qualitatively with WD.}
At $\lambda \geq 0.5$, defect traces show smooth monotonic rises.
At $\lambda \leq 0.3$, defect traces become intermittent and spiky, with transient explosions ($100$--$5{,}000\times$ baseline) followed by collapses back to baseline (\Cref{fig:defect_traces}).
The rolling median stays flat until just before grokking, when it finally sustains high values.

\paragraph{Non-monotonic defect magnitude.}
At the final checkpoint, the defect magnitude is non-monotonic in WD (\Cref{tab:defect_wd_final}).
$\lambda = 0.5$ shows the largest final defect ($837$--$1{,}288$), exceeding $\lambda = 1.0$ by ${\sim}10\times$, because the analysis captures the model mid-transition.
$\lambda = 0.1$ shows near-zero defect at its final checkpoint because the model has fully converged post-grokking.

\begin{table}[t]
\centering
\caption{Dual-task final-checkpoint defect and integrability across WD.}
\label{tab:defect_wd_final}
\begin{tabular}{@{}lrrr@{}}
\toprule
$\lambda$ & Final Defect Range & Integrability ($\rho$) & Grad Cosine (same-task) \\
\midrule
1.0 & 82--144 & 1.000 & 0.77--0.84 \\
0.5 & 837--1,288 & 1.000 & 0.47--0.76 \\
0.1 & ${\sim}0$ & 1.000 & 0.14--0.49 \\
0.0 & 0--52 & varies & 0.04--0.42 \\
\bottomrule
\end{tabular}
\end{table}

\paragraph{WD $= 0.5$ is the ``efficient'' regime (dual-task).}
In the dual-task setting, $\lambda = 0.5$ emerges as a distinguished regime: it produces the tightest reconstruction threshold ($k^* = 5$, all seeds), the minimum defect lead (4.9\%), and the most compressed grokking trajectory.
This suggests that at $\lambda = 0.5$, the regularization pressure is optimally matched to the task complexity, producing direct trajectories from memorization to generalization with minimal detour.
Notably, this efficiency is specific to the dual-task setting; the tri-task lead fraction is $77\%$ at $\lambda = 0.5$ (vs.\ $82\%$ at $\lambda = 1.0$), indicating that the additional task constraints consume the geometric slack that enables efficient trajectories.

\section{Holographic Incompressibility}
\label{sec:incompress}

\subsection{Reconstruction Threshold $k^*$}

We reconstruct the grokked model as $\theta_\text{recon} = \theta_\text{init} + \sum_{i=1}^{k} (\mathbf{v}_i^\top \Delta\theta)\,\mathbf{v}_i$ using the top-$k$ uncentered PCA components and measure test accuracy on both tasks simultaneously.
The threshold $k^*$ is the minimum $k$ where all tasks exceed 90\%.

\paragraph{Dual-task.}
The threshold ranges from $k^* = 5$ ($\lambda = 0.5$, all seeds unanimous) to $k^* = 10$ ($\lambda = 0.2$, seed~137), with a clear monotonic trend: lower WD requires more components (\Cref{tab:dual_kstar}).
The transition from chance to near-perfect accuracy is sharp at high WD (chance at $k$, 99\% at $k+1$) and more gradual at low WD.

\begin{table}[t]
\centering
\caption{Reconstruction threshold $k^*$ for dual-task (both tasks $>90\%$).}
\label{tab:dual_kstar}
\begin{tabular}{@{}lccccc@{}}
\toprule
$\lambda$ & Seed 42 & Seed 137 & Seed 2024 & Mean & Var.\ at $k^*$ \\
\midrule
1.0 & 7 & 6 & 6 & 6.3 & 99.7--99.9\% \\
0.5 & 5 & 5 & 5 & 5.0 & 99.6--99.7\% \\
0.3 & 5 & 6 & 6 & 5.7 & 99.4--99.7\% \\
0.2 & 6 & 10 & 7 & 7.7 & 99.5--99.7\% \\
0.1 & 9 & 7 & 9 & 8.3 & 99.2--99.6\% \\
\bottomrule
\end{tabular}
\end{table}

\begin{figure}[t]
    \centering
    \begin{subfigure}[t]{0.48\textwidth}
        \centering
        \includegraphics[width=\textwidth]{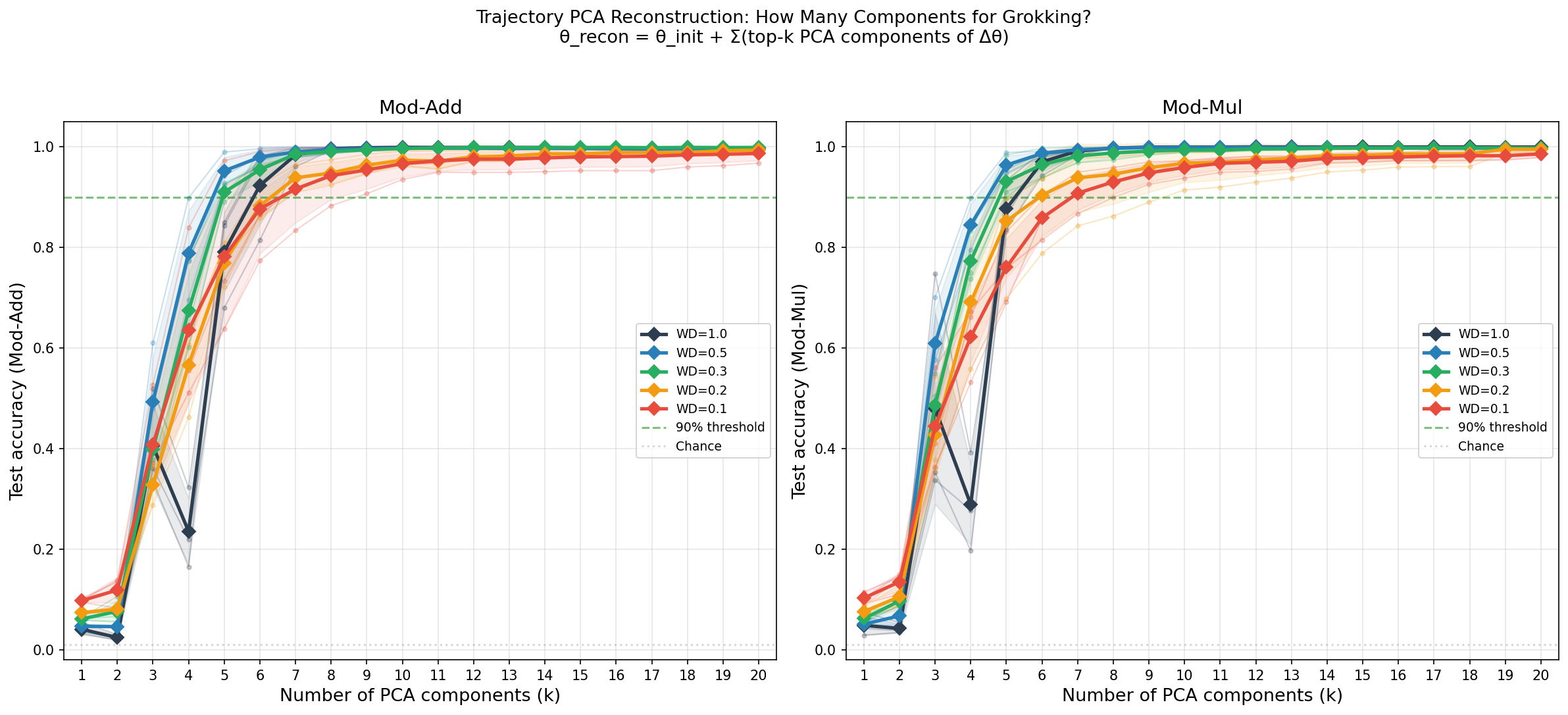}
        \caption{Accuracy vs.\ $k$ for dual-task across WD values: sharp transition from chance to $>99\%$.}
        \label{fig:acc_vs_k}
    \end{subfigure}
    \hfill
    \begin{subfigure}[t]{0.48\textwidth}
        \centering
        \includegraphics[width=\textwidth]{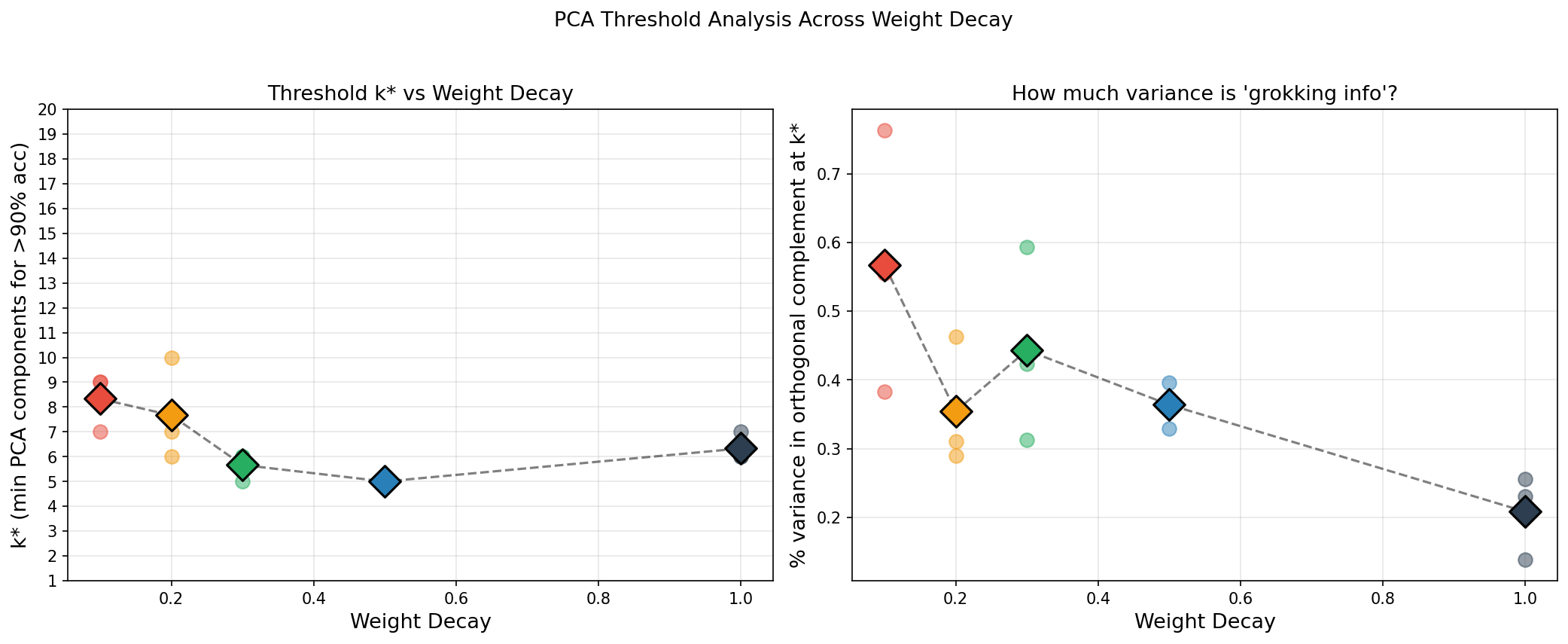}
        \caption{$k^*$ vs.\ WD: monotonically increasing at lower WD.}
        \label{fig:kstar_vs_wd}
    \end{subfigure}
    \caption{Reconstruction threshold. The grokking solution requires 5--10 PCA directions, with $k^*$ increasing at lower WD.}
    \label{fig:reconstruction}
\end{figure}

\paragraph{Tri-task.}
Across all five non-zero WD values, tri-task models require comparable or \emph{fewer} PCA components ($k^* \approx 4$--$6$, mean across seeds) than dual-task models ($k^* \approx 5$--$8$; \Cref{tab:tri_kstar}), despite having more parameters and an additional task.
The highest WD values ($\lambda \geq 0.5$) show the tightest thresholds ($k^* \approx 4.3$), while lower WD values ($\lambda \leq 0.3$) require slightly more components ($k^* \approx 5.3$--$6.3$).
This suggests that multi-task learning provides additional constraints that concentrate the generalizing solution into fewer directions.

\begin{table}[t]
\centering
\caption{Reconstruction threshold $k^*$ for tri-task (all three tasks $>90\%$).}
\label{tab:tri_kstar}
\begin{tabular}{@{}lcccc@{}}
\toprule
$\lambda$ & Seed 42 & Seed 137 & Seed 2024 & Mean \\
\midrule
1.0 & 5 & 5 & 3 & 4.3 \\
0.5 & 4 & 4 & 5 & 4.3 \\
0.3 & 7 & 5 & 7 & 6.3 \\
0.2 & 5 & 5 & 6 & 5.3 \\
0.1 & 5 & 5 & 8 & 6.0 \\
\bottomrule
\end{tabular}
\end{table}

\subsection{Post-Hoc Compression Tests}

We test whether the grokked model can be compressed using only the initial and final parameter snapshots (no trajectory information).

\paragraph{Per-layer SVD.}
For each weight matrix $W$, we compute $\Delta W = W_\text{grok} - W_\text{init}$ and keep only the top-$r$ singular values.
Results are all-or-nothing: rank-64 gives chance ($0.8$--$1.1\%$), while rank-128 (full rank $= d_\text{model}$) gives near-perfect accuracy ($99.6\%$ add, $99.9\%$ mul).
There is no low-rank structure in the weight matrix deltas.

\paragraph{Magnitude pruning.}
Zeroing out entries of $\Delta\theta$ below a magnitude threshold: keeping 50\% of the largest entries (99.3\% of $\norm{\Delta\theta}^2$) gives chance.
Partial recovery begins only at 80\% retention.

\paragraph{Uniform scaling.}
Scaling $\Delta\theta$ by a constant: $\theta_s = \theta_\text{init} + s \cdot \Delta\theta$.
At $s = 0.95$ and $s = 1.05$, accuracy drops to chance ($0.7$--$1.5\%$).
The grokking solution occupies an extremely precise point in parameter space; even a $\pm 5\%$ perturbation destroys it.

\begin{figure}[t]
    \centering
    \begin{subfigure}[t]{0.48\textwidth}
        \centering
        \includegraphics[width=\textwidth]{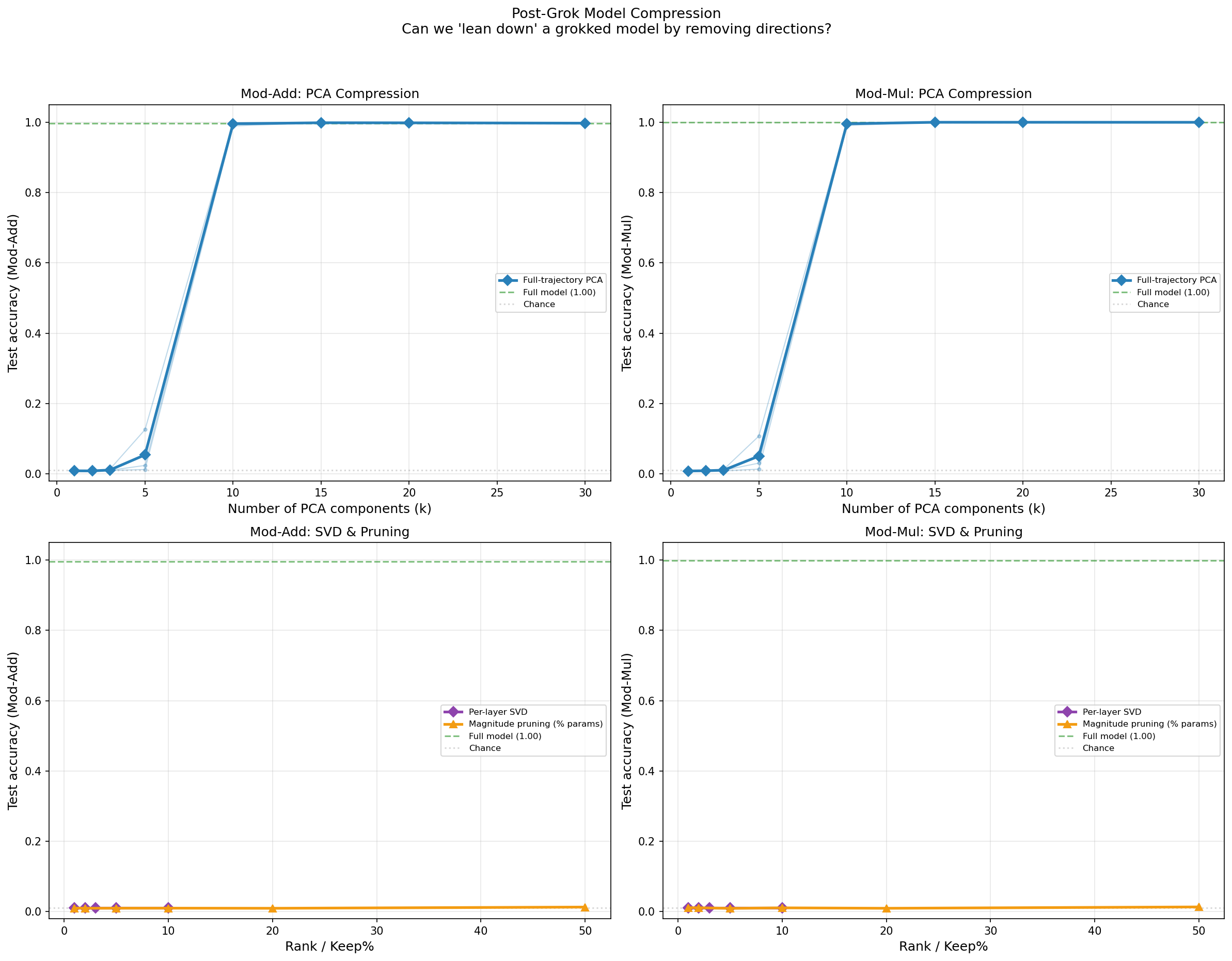}
        \caption{Post-hoc compression: PCA trajectory reconstruction (green), per-layer SVD (blue), magnitude pruning (orange). Only PCA with $k \geq k^*$ succeeds.}
        \label{fig:compression}
    \end{subfigure}
    \hfill
    \begin{subfigure}[t]{0.48\textwidth}
        \centering
        \includegraphics[width=\textwidth]{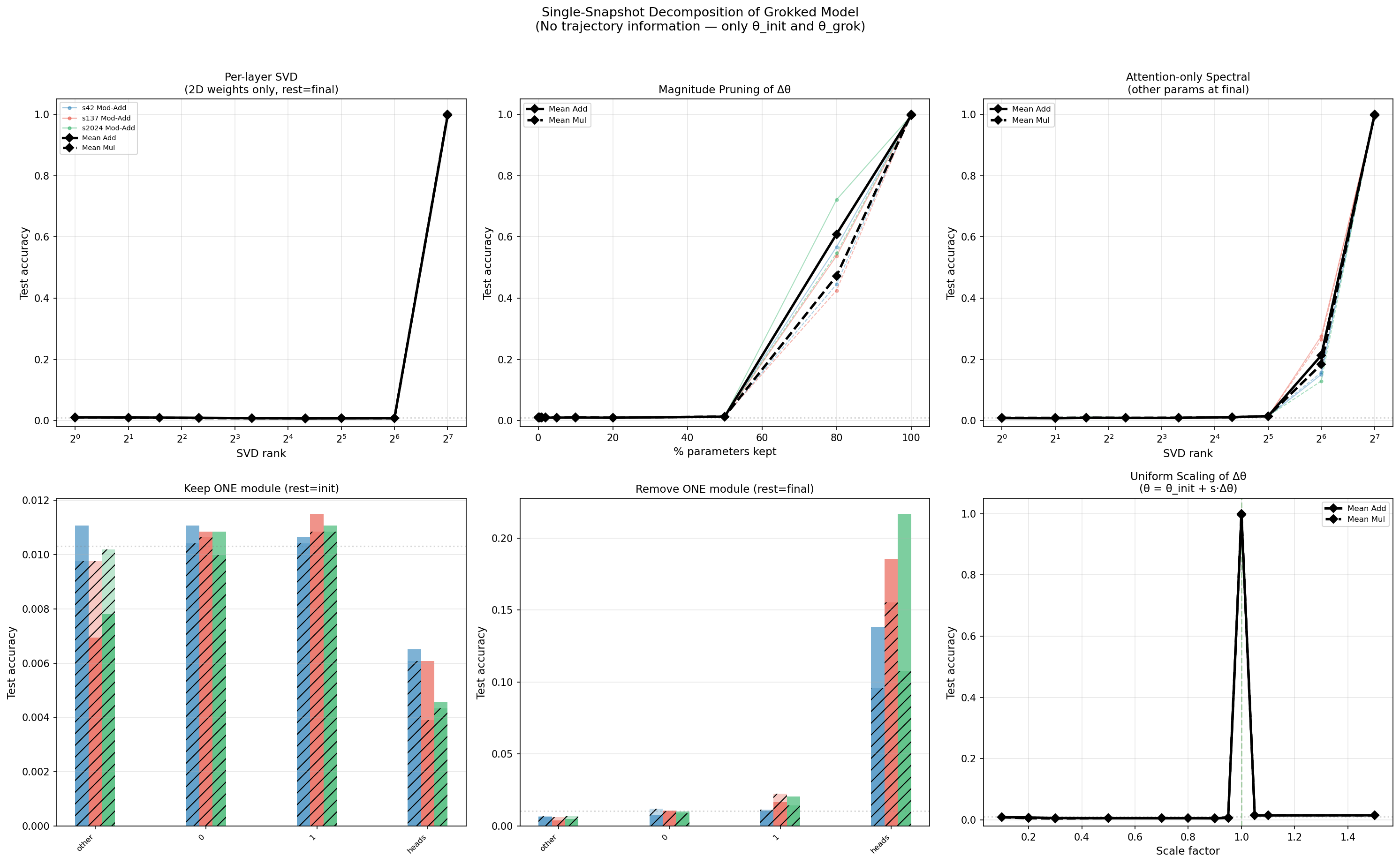}
        \caption{Snapshot decomposition: per-layer SVD rank, magnitude pruning threshold, and uniform scaling. All fail except full preservation.}
        \label{fig:snapshot}
    \end{subfigure}
    \caption{The grokking solution is incompressible by any post-hoc method. Only trajectory-informed PCA succeeds.}
    \label{fig:incompressibility}
\end{figure}

\subsection{Orthogonal Fragility}

We test the causal necessity of the orthogonal complement by removing a fraction $f$ of the gradient's component orthogonal to the PCA manifold during training:
\begin{equation}
    \mathbf{g}_\text{new} = \mathbf{g} - f \cdot \mathbf{g}_\perp, \qquad \mathbf{g}_\perp = \mathbf{g} - B(B^\top \mathbf{g}).
\end{equation}

\paragraph{Dual-task fine-grained dose-response.}
PCA projection at strength $s = 0.25$ delays grokking by 70--140\% ($13.9$k $\to$ $23.7$k steps at seed~42), while random projection at the same strength has no effect (\Cref{fig:app_dual_ablation}).
A fine-grained dose-response study (\Cref{tab:ortho_dual}) reveals monotonic delay from 0\% to 7\% deletion, followed by complete failure at 10--25\%.
Interestingly, 50\% deletion recovers grokking at a ${\sim}163\%$ delay, suggesting that large deletions alter the optimization dynamics sufficiently to find alternative routes to generalization.

\begin{table}[t]
\centering
\caption{Dual-task orthogonal deletion dose-response (WD=1.0, seed~42, 50k step budget).}
\label{tab:ortho_dual}
\begin{tabular}{@{}rccl@{}}
\toprule
Delete \% & Add Grok & Mul Grok & Delay \\
\midrule
0\% & 14,100 & 14,000 & (baseline) \\
1\% & 18,100 & 18,100 & $+29\%$ \\
2\% & 21,200 & 20,900 & $+50\%$ \\
3\% & 23,500 & 23,300 & $+67\%$ \\
5\% & 37,000 & 36,800 & $+163\%$ \\
7\% & 42,600 & 42,400 & $+203\%$ \\
10\% & FAIL & FAIL & --- \\
15\% & FAIL & FAIL & --- \\
25\% & FAIL & FAIL & --- \\
50\% & 36,900 & 36,800 & $+163\%$ \\
\bottomrule
\end{tabular}
\end{table}

\paragraph{Tri-task fine-grained dose-response.}
The tri-task setting is even more fragile.
Fine-grained deletion (\Cref{tab:ortho_tri}) shows that even 1\% orthogonal deletion delays grokking by ${\sim}37\%$ (mean across tasks), with monotonically increasing delay up to 7\% (where the delay exceeds $120\%$).
At 10\% deletion and above, grokking fails entirely within 60k steps---a sharp cliff between 7\% and 10\%.
This establishes that $<1\%$ of the gradient variance (orthogonal to the PCA manifold) is causally necessary for grokking to occur.

\begin{table}[t]
\centering
\caption{Tri-task orthogonal deletion dose-response (WD=1.0, seed~42, 60k step budget).}
\label{tab:ortho_tri}
\begin{tabular}{@{}rccccl@{}}
\toprule
Delete \% & Add Grok & Mul Grok & Sq Grok & Mean & Delay \\
\midrule
0\% & 25,600 & 18,200 & 23,400 & 22,400 & (baseline) \\
1\% & 35,200 & 21,200 & 28,800 & 28,400 & $+27\%$ \\
2\% & 42,400 & 30,200 & 36,000 & 36,200 & $+62\%$ \\
3\% & 35,000 & 29,000 & 33,600 & 32,500 & $+45\%$ \\
5\% & 35,400 & 26,800 & 33,800 & 32,000 & $+43\%$ \\
7\% & 56,000 & 44,600 & 49,600 & 50,100 & $+124\%$ \\
10\% & FAIL & FAIL & FAIL & --- & --- \\
15\% & FAIL & FAIL & FAIL & --- & --- \\
25\% & FAIL & FAIL & FAIL & --- & --- \\
\bottomrule
\end{tabular}
\end{table}

\begin{figure}[t]
    \centering
    \includegraphics[width=0.75\textwidth]{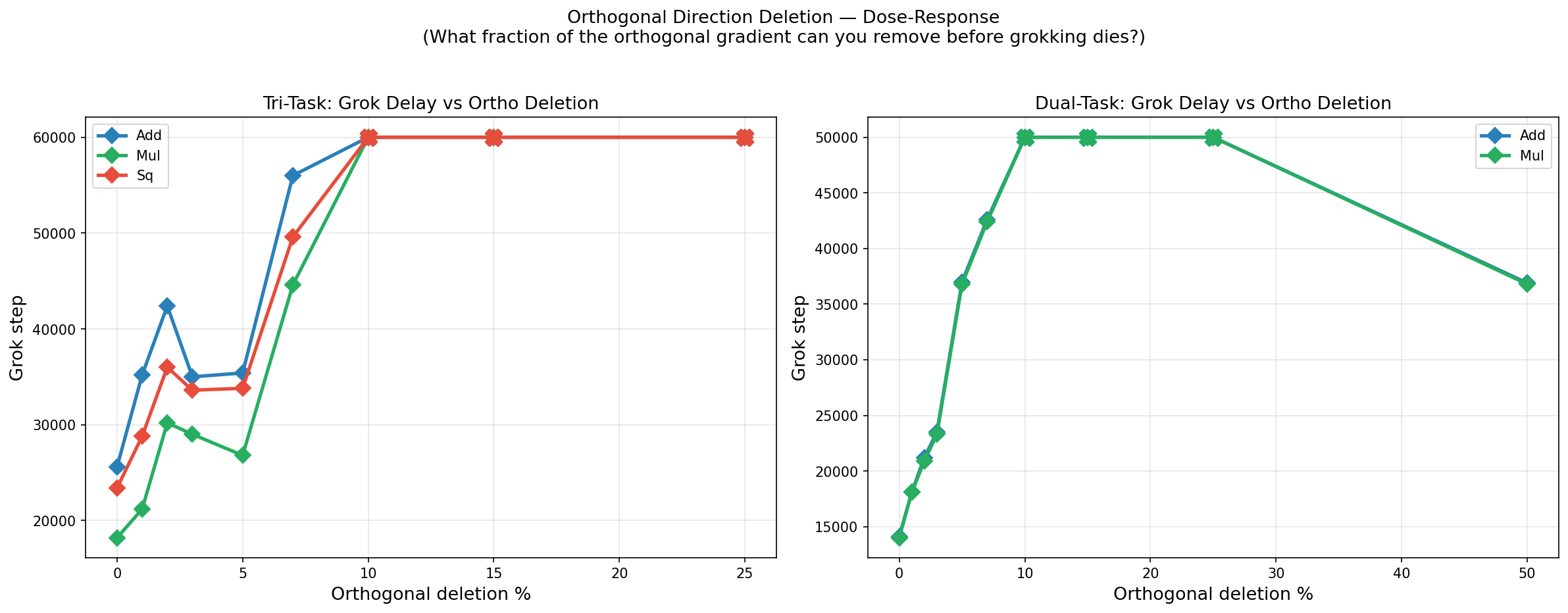}
    \caption{Orthogonal deletion dose-response for both tri-task and dual-task. Grok delay increases monotonically with deletion fraction until a sharp cliff at ${\sim}10\%$, beyond which grokking fails entirely. X-marks indicate failure to grok within the step budget.}
    \label{fig:ortho_fine}
\end{figure}

\paragraph{Synthesis.}
Together, these results show that generalization-relevant information is simultaneously low-dimensional ($k^*$ principal trajectory directions suffice for reconstruction) and globally distributed (full-rank weights, with every module necessary).
This dual character explains why compression and partial ablation fail simultaneously: the solution occupies a thin subspace of parameter space, but that subspace is woven through all weight matrices at full rank.
Any method that removes dimensions (SVD truncation, magnitude pruning) or perturbs the global structure (uniform scaling, orthogonal deletion) destroys the delicate encoding.

\section{Spectral Geometry of the Attention Operator Transition}
\label{sec:spectral_geometry}

The preceding sections characterize grokking through trajectory PCA, commutator defects, and Hessian curvature.
We now ask: \emph{what happens inside the attention operators during the commutator transition?}
To answer this, we analyze the singular value spectrum of the attention weight matrices $W_Q$ and $W_K$ at each training checkpoint for both dual-task and tri-task models, complementing the single-task spectral analysis of a companion study \citep{xu2026integrability}.

\paragraph{Geometric picture.}
These observations suggest that grokking corresponds to a geometric phase transition in the attention operators.
Early in training, the dominant singular directions of $W_Q$ and $W_K$ are nearly degenerate and the operators exhibit strong non-commutativity.
As training proceeds, a symmetry-breaking event occurs in which one singular direction becomes dominant, the commutator defect collapses, and the attention operators align into a shared eigenbasis.
Generalization emerges only after this alignment is achieved.
The spectral gap, the matrix commutator, and grokking are thus linked by a single dynamical narrative: \emph{degeneracy breeds instability, symmetry breaking resolves it, and generalization follows}.

\subsection{Spectral Symmetry Breaking Precedes Grokking}
\label{sec:fact_symmetry}

Define the leading spectral gap of the query matrix at layer $\ell$:
\begin{equation}
    g_{12}^{(\ell)} = \sigma_1(W_Q^{(\ell)}) - \sigma_2(W_Q^{(\ell)}).
\end{equation}
In every grokking run we examine---dual-task (3~seeds) and tri-task (3~seeds)---the spectral gap opens during or immediately before commutator collapse (\Cref{fig:multitask_portrait}).
The attention operator transitions from a near-degenerate regime ($\sigma_1 \approx \sigma_2$) to a rank-1-dominated regime ($\sigma_1 \gg \sigma_2$), and this transition is a necessary precursor to generalization.
Memorizing (no--weight-decay) runs show no such symmetry-breaking event: their spectral gaps remain small and the commutator wanders diffusively (\Cref{fig:multitask_control}).

\subsection{Grokking Follows a Universal Phase Trajectory}
\label{sec:fact_trajectory}

\begin{figure}[t]
    \centering
    \includegraphics[width=\textwidth]{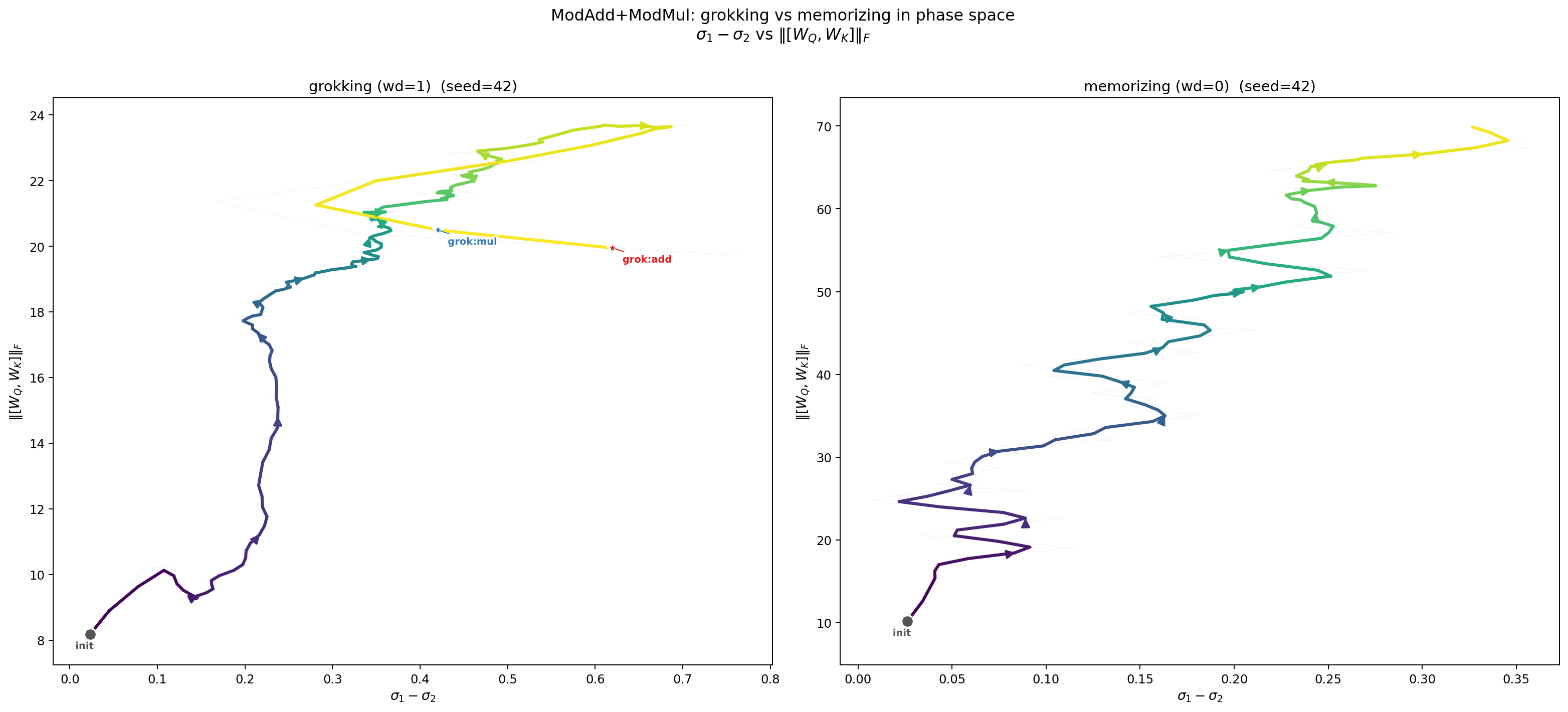}
    \caption{Grokking vs.\ memorizing phase portraits for dual-task arithmetic (modular addition + multiplication, seed~42, layer~0).
    Left: grokking run ($\mathrm{wd}=1$) shows the characteristic competition--instability--alignment loop, with both tasks grokking near the transition to alignment.
    Right: memorizing run ($\mathrm{wd}=0$) shows a diffuse random walk with $\sim\!3\times$ higher commutator values and no directed escape.}
    \label{fig:multitask_control}
\end{figure}

In the $(\sigma_1 - \sigma_2,\;\norm{[W_Q, W_K]}_F)$ phase plane, every grokking run traces a characteristic loop through three regimes:
\begin{enumerate}
    \item \textbf{Competition}: small spectral gap, rising commutator;
    \item \textbf{Instability}: spectral gap opening, commutator at peak;
    \item \textbf{Alignment}: large spectral gap, collapsing commutator; generalization appears.
\end{enumerate}
This loop structure is observed in all single-task operations (companion study), in dual-task arithmetic, and in tri-task arithmetic (\Cref{fig:multitask_portrait,fig:multitask_control}).
Memorizing runs exhibit no loop structure: their trajectories are diffuse random walks with substantially higher commutator values and no directed escape.
The universality of the loop suggests that grokking reflects a generic geometric transition in the attention operator rather than a task-specific phenomenon.

\subsection{Multi-Task Grokking Occurs Along a Shared Trajectory}
\label{sec:fact_multitask}

\begin{figure}[t]
    \centering
    \includegraphics[width=\textwidth]{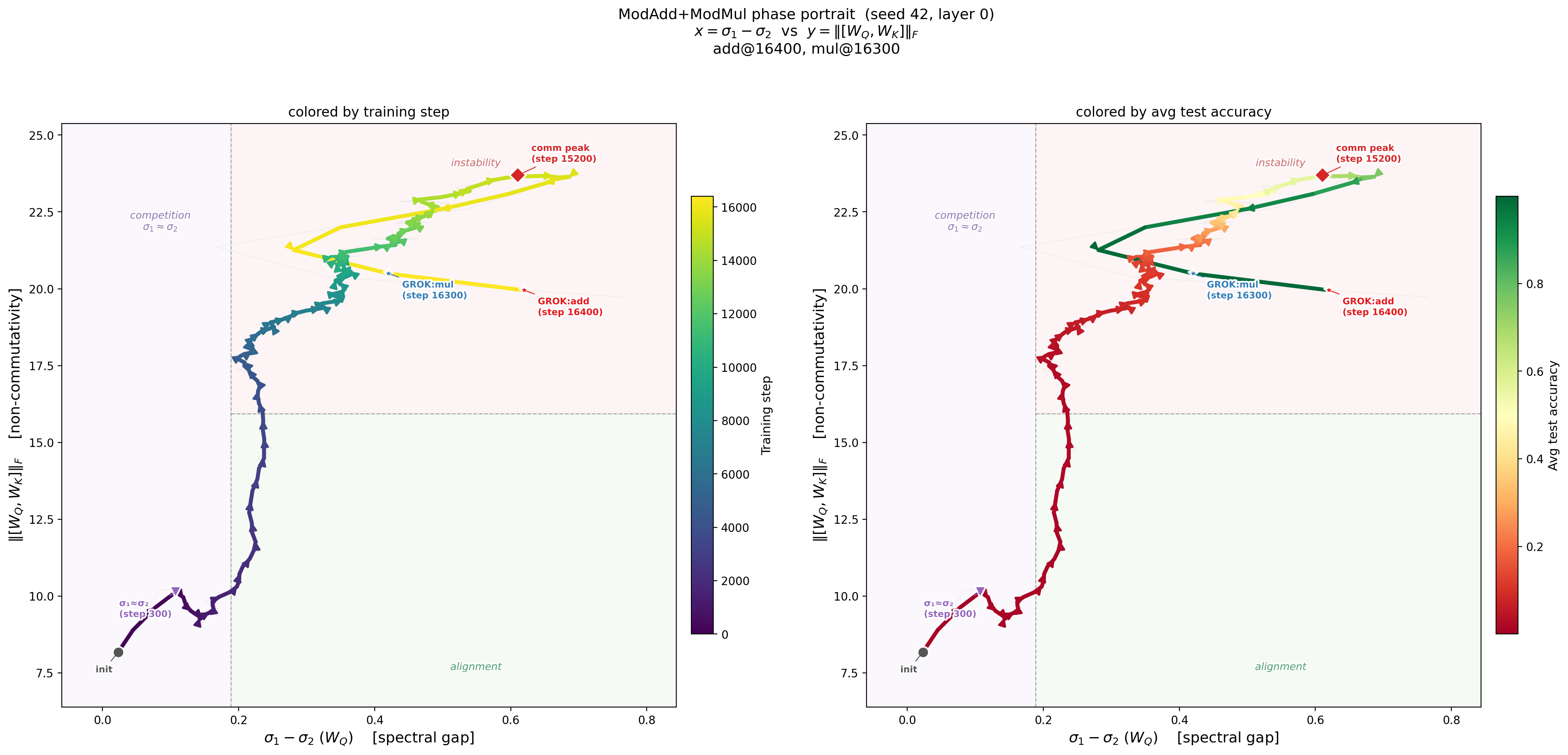}
    \caption{Phase portrait for tri-task arithmetic (seed~42, layer~0) in the $(\sigma_1 - \sigma_2,\;\norm{[W_Q, W_K]}_F)$ plane.
    Left: colored by training step. Right: colored by mean test accuracy across the three tasks.
    The three tasks grok at different positions along a single shared trajectory: multiplication (step~17{,}900) near the commutator peak, $a^2\!+\!b^2$ (step~24{,}800) during descent, and addition (step~30{,}500) in the alignment region.
    The competition--instability--alignment structure is preserved from the single-task case.}
    \label{fig:multitask_portrait}
\end{figure}

In the tri-task setting, all three tasks grok at different training steps, yet the shared-trunk attention operators trace a \emph{single} trajectory in phase space.
Each task generalizes at a different position along this shared path (\Cref{fig:multitask_portrait}):
\begin{equation}
    \underbrace{\text{comm.\ peak}}_{\text{step 17{,}600}}
    \;\to\;
    \underbrace{\text{grok:mul}}_{\text{step 17{,}900}}
    \;\to\;
    \underbrace{\text{grok:}a^2\!+\!b^2}_{\text{step 24{,}800}}
    \;\to\;
    \underbrace{\text{grok:add}}_{\text{step 30{,}500}}
\end{equation}
In the dual-task setting, the two tasks grok within 100--400 steps of each other (\Cref{tab:dual_grok}), consistent with the shorter trajectory and the near-simultaneity of operator alignment for two tasks.
This result implies that multi-task grokking does not require separate per-task circuits or independent phase transitions.
Instead, training reorganizes the shared geometric structure of the attention operators, and individual tasks become solvable as the trajectory passes through their respective alignment thresholds.
The commutator transition is a property of the trunk, not of any individual task head.

\subsection{Layer Asymmetry in Instability Magnitude}
\label{sec:fact_layer}

We decompose the matrix commutator by layer and examine the peak magnitude of $\norm{[W_Q^{(\ell)}, W_K^{(\ell)}]}_F$ during training.
Across all conditions tested (\Cref{tab:layer_asymmetry}), the bottom layer (Layer~0) develops substantially stronger non-commutativity than the top layer (Layer~1):
\begin{equation}
    \max_t \norm{[W_Q^{(0)}, W_K^{(0)}]}_F \;>\; \max_t \norm{[W_Q^{(1)}, W_K^{(1)}]}_F,
    \qquad \text{ratio} \approx 1.3\text{--}2.4\times.
\end{equation}
This amplitude asymmetry is robust across all seeds and datasets; the peak \emph{timing} order between layers is not consistent across seeds and should not be regarded as an invariant.

\begin{table}[t]
\centering
\caption{Layer-wise peak commutator magnitude.
Layer~0 (bottom) consistently develops $1.3$--$2.4\times$ stronger non-commutativity than Layer~1 (top).}
\label{tab:layer_asymmetry}
\begin{tabular}{@{}lccccc@{}}
\toprule
Dataset & Seed & L0 peak & L1 peak & Ratio \\
\midrule
Dual-task (add+mul) & 42 & 23.7 & 14.3 & 1.7$\times$ \\
 & 137 & 20.5 & 12.2 & 1.7$\times$ \\
 & 2024 & 21.3 & 13.1 & 1.6$\times$ \\
\midrule
Tri-task (add+mul+sq) & 42 & 24.7 & 11.3 & 2.2$\times$ \\
 & 137 & 27.3 & 11.6 & 2.4$\times$ \\
 & 2024 & 24.6 & 12.8 & 1.9$\times$ \\
\bottomrule
\end{tabular}
\end{table}

\begin{figure}[t]
    \centering
    \includegraphics[width=0.85\textwidth]{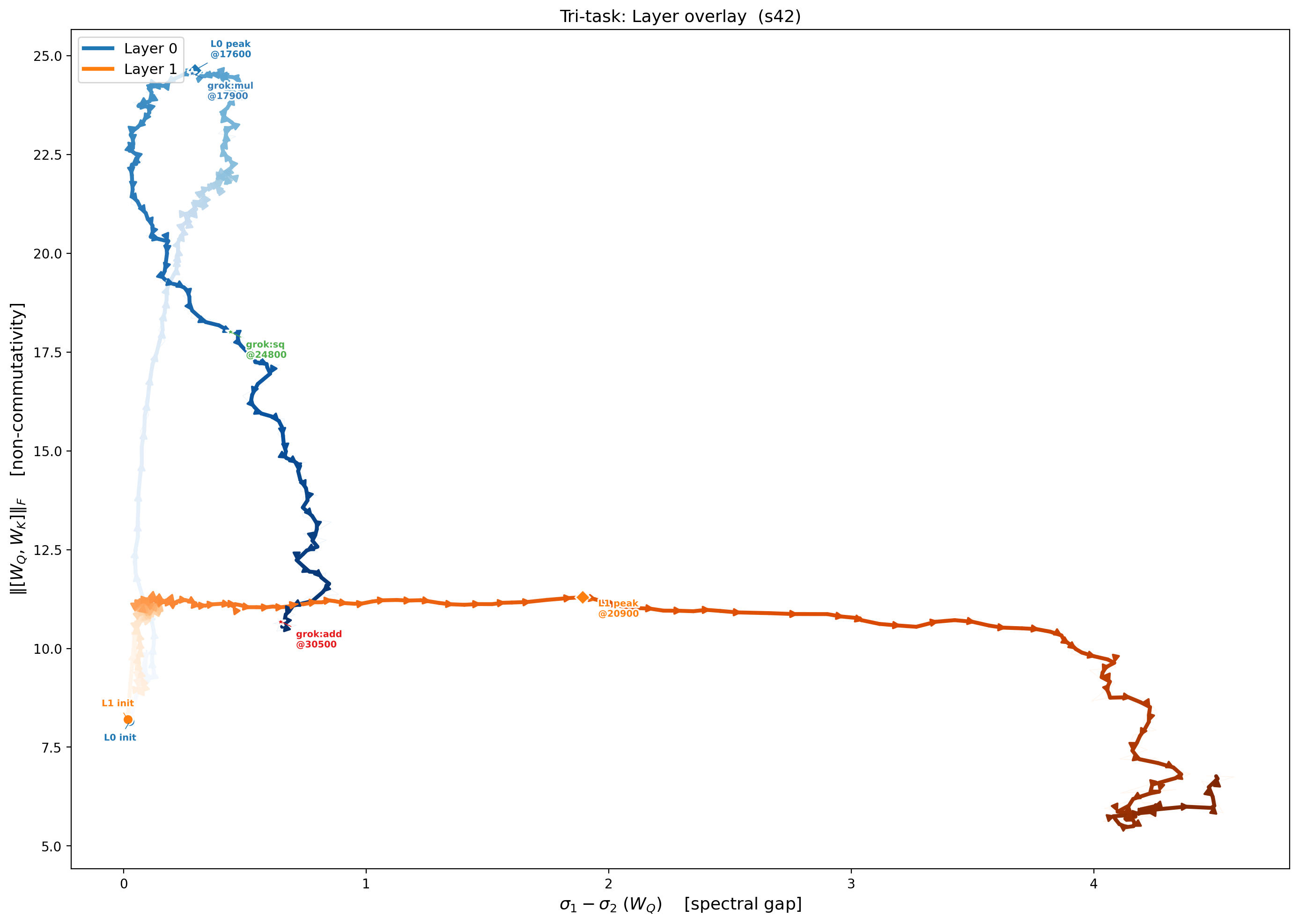}
    \caption{Layer-wise phase portrait overlay for tri-task arithmetic (seed~42).
    Layer~0 (blue) traces a large loop reaching peak commutator $\approx 24.7$ at step~17{,}600, while Layer~1 (orange) remains below $\approx 11.3$.
    The bottom layer bears the dominant geometric instability.
    The tri-task amplifies the asymmetry relative to the single-task case ($2.2\times$ vs.\ $1.1\times$).}
    \label{fig:layer_overlay}
\end{figure}

The per-layer phase portrait (\Cref{fig:layer_overlay}) provides a geometric interpretation: Layer~0 traces a large loop in the $(\sigma_1 - \sigma_2,\;\norm{[W_Q, W_K]}_F)$ plane, extending to high commutator values and wide spectral gaps before collapsing near the grokking transition.
Layer~1 traces a smaller loop that peaks earlier and collapses sooner.
The bottom layer bears the dominant geometric instability during grokking, while the top layer stabilizes first.
In the multi-task setting, this asymmetry is amplified: the ratio increases from $\sim\!1.1\times$ (single-task) to $\sim\!1.7$--$2.4\times$ (dual- and tri-task), suggesting that the additional representational demands of multiple tasks are absorbed primarily by the lower layer.

\subsection{Gram Matrix Eigenvalue Gap Dynamics}
\label{sec:gram_matrix}

The weight-matrix spectral analysis above examines individual attention operators $W_Q$ and $W_K$.
A complementary view comes from the \emph{rolling-window Gram matrix} $\boldsymbol{G}(t) = \boldsymbol{X}(t)\boldsymbol{X}(t)^\top$, formed from $W$ consecutive parameter updates $\boldsymbol{\delta}_s = \boldsymbol{\theta}_{s+1} - \boldsymbol{\theta}_s$ (all attention weights flattened into $\mathbb{R}^p$), which captures the spectral structure of the optimisation trajectory itself~\citep{xu2026spectral_edge}.
Three quantities are extracted via the SVD of the $W \times p$ update matrix $\boldsymbol{X}$ with singular values $\sigma_1 \ge \sigma_2 \ge \cdots \ge \sigma_W$:
\begin{itemize}
  \item[(i)] The \textbf{signal rank} $k^* = \arg\max_j\, \omega_j \cdot (\sigma_j/\sigma_{j+1})$, where $\omega_j = \sigma_j/\sum_i \sigma_i$ weights the ratio by spectral mass at position~$j$, suppressing spurious tail ratios outside the extreme aspect ratio regime.
  \item[(ii)] The \textbf{gap ratio} $R = \sigma_{k^*}/\sigma_{k^*+1}$, measuring how sharply the signal separates from sub-dominant modes.
  \item[(iii)] The \textbf{sub-leading eigenvalue gap} $g_{23} = \sigma_2^2 - \sigma_3^2$, tracking the separation between the second and third eigenvalues of $\boldsymbol{G}$.
\end{itemize}
We compute all three quantities with $W = 10$ across 18 grokking runs (12~single-task from 4~operations $\times$ 3~seeds, plus 3~dual-task and 3~tri-task, all with $\lambda > 0$) and 18 matched controls ($\lambda = 0$).
\Cref{tab:gram_spectral} reports the full results.

\begin{table}[t]
\centering
\caption{Rolling-window Gram matrix spectral analysis ($W = 10$).
$g_{23}^{\mathrm{early}}$: peak $g_{23}$ before grokking;
$g_{23}^{\mathrm{grok}}$: $g_{23}$ at the grokking step;
Decline $= g_{23}^{\mathrm{early}} / g_{23}^{\mathrm{grok}}$;
$R$: gap ratio at signal rank $k^*$.
Single-task operations that do not grok within $2 \times 10^5$ steps (x2\_xy\_y2, x3\_xy) are omitted.}
\label{tab:gram_spectral}
\small
\begin{tabular}{llrrrrrrc}
\toprule
\textbf{Setting} & $\boldsymbol{\lambda}$ & \textbf{Grok step} & $g_{23}^{\mathrm{early}}$ & $g_{23}^{\mathrm{grok}}$ & \textbf{Decline} & $R_{\mathrm{early}}$ & $k^*_{\mathrm{term}}$ & \textbf{Decl.} \\
\midrule
\multicolumn{9}{l}{\textit{Single-task (mean $\pm$ std over 3 seeds)}} \\
add          & 1.0 & 2500--3100  & $14.6 \pm 0.2$ & $0.56 \pm 0.4$ & $50\times$  & $1.40 \pm 0.05$ & 1 & \cmark \\
             & 0.0 & ---         & $33.0 \pm 9.3$ & ---            & ---         & $2.72 \pm 0.22$ & --- & --- \\
mul          & 1.0 & 2600--3000  & $14.7 \pm 0.1$ & $0.39 \pm 0.2$ & $49\times$  & $1.41 \pm 0.05$ & 1 & \cmark \\
             & 0.0 & ---         & $23.6 \pm 10.7$ & ---           & ---         & $2.85 \pm 0.11$ & --- & --- \\
sub          & 1.0 & 3300--3700  & $14.4 \pm 0.1$ & $0.48 \pm 0.2$ & $35\times$  & $1.31 \pm 0.02$ & 1 & \cmark \\
             & 0.0 & ---         & $33.2 \pm 2.8$ & ---            & ---         & $2.62 \pm 0.23$ & --- & --- \\
$a^2{+}b^2$  & 1.0 & 2000--2500  & $18.9 \pm 0.5$ & $0.94 \pm 0.5$ & $28\times$  & $1.47 \pm 0.08$ & 1 & \cmark \\
             & 0.0 & ---         & $28.3 \pm 13.1$ & ---           & ---         & $3.12 \pm 0.64$ & --- & --- \\
\midrule
\multicolumn{9}{l}{\textit{Dual-task: add + mul}} \\
             & 1.0 & 9700--16100 & $30.8 \pm 0.2$ & $2.94 \pm 0.8$ & $11\times$  & $1.35 \pm 0.10$ & 1 & \cmark \\
             & 0.0 & ---         & $18.6 \pm 4.0$ & ---            & ---         & $1.96 \pm 0.35$ & --- & --- \\
\midrule
\multicolumn{9}{l}{\textit{Tri-task: add + mul + sub}} \\
             & 1.0 & 13100--26400 & $31.3 \pm 0.3$ & $0.53 \pm 0.2$ & $63\times$ & $1.41 \pm 0.11$ & 1 & \cmark \\
             & 0.0 & ---          & $32.3 \pm 0.7$ & ---            & ---        & $1.77 \pm 0.31$ & --- & --- \\
\midrule
\multicolumn{9}{l}{\textit{Aggregate}} \\
All $\lambda > 0$ & & & & & $39\times$ mean & $1.39 \pm 0.08$ & 1 (15/18) & 18/18 \\
All $\lambda = 0$ & & & & & ---            & $2.51 \pm 0.56$ & --- & 1/18 \\
\bottomrule
\end{tabular}
\end{table}

The results are unambiguous: $g_{23}$ declines before grokking in 18 of 18 runs with weight decay (mean $39\times$, range $8$--$111\times$) and in only 1 of 18 matched controls.
The gap ratio $R$ separates the two conditions with non-overlapping distributions: $\lambda = 0$ runs show \emph{higher} $R = 2.51 \pm 0.56$ (memorization updates are rank-1 concentrated, reflecting highly aligned gradients during rote memorization) compared to $R = 1.39 \pm 0.08$ for $\lambda > 0$ (circuit-formation updates spread across spectral modes as the generalising representation assembles).
The signal rank stabilises at $k^*_{\mathrm{terminal}} = 1$ in 15 of 18 grokking runs (83\%), matching the 10/12 reported by \citet{xu2026spectral_edge} and confirming single-mode dominance at convergence.

Several patterns emerge across the single- and multi-task settings.
First, the decline magnitude scales with task complexity: single-task operations show $28$--$50\times$ mean decline, dual-task $11\times$, and tri-task $63\times$.
The tri-task result is particularly striking: despite having the longest grokking timescale (steps 13{,}100--26{,}400 vs.\ 2{,}000--3{,}700 for single-task), $g_{23}$ drops from ${\sim}31$ to ${\sim}0.5$, indicating that multi-operation circuit formation involves dramatic spectral compression.
Second, the early-phase $g_{23}$ level itself differentiates: dual-task $\lambda = 1.0$ shows $g_{23}^{\mathrm{early}} = 30.8$ vs.\ $18.6$ for $\lambda = 0$ ($1.7\times$), while tri-task converges ($31.3$ vs.\ $32.3$), suggesting that the early spectral structure depends on task multiplicity.
Third, two single-task operations ($a^2 + ab + b^2$ and $a^3 + ab$, both mod~97) do not grok within $2 \times 10^5$ steps even at $\lambda = 1.0$; these runs show $R_{\mathrm{early}} \approx 3.4$ (matching $\lambda = 0$ controls) and no $g_{23}$ decline, confirming that the spectral signature is tied to the grokking transition itself, not merely to the presence of weight decay.

The $\lambda = 0$ controls display the \emph{predicted} spectral signature of the grokking-absent regime~\citep[Remark~12.22]{xu2026spectral_edge}: $g_{23}$ remains high throughout all training steps with no decline, confirming that without the curvature floor $\omega$ the spectral symmetry-breaking transition never initiates.

The Gram matrix analysis thus provides a complementary perspective to the weight-matrix SVD of the preceding subsections.
While the weight-matrix spectral gap $g_{12} = \sigma_1(W_Q) - \sigma_2(W_Q)$ tracks the internal structure of individual attention operators, the Gram matrix $g_{23}$ tracks the coherence of the \emph{optimisation trajectory} itself.
Both converge on the same mechanistic picture: grokking is a spectral phase transition in which concentrated update dynamics give way to distributed circuit formation, with weight decay providing the driving force.

\section{Superposition Dynamics in Parameter Space}
\label{sec:superposition}

A central question in mechanistic interpretability is how multiple algorithmic features are represented simultaneously in overparameterized neural networks.
Prior work has emphasized activation-space superposition, in which multiple features are encoded in overlapping neuron subspaces.
Here, we provide complementary evidence for superposition in \emph{parameter space}, and characterize its emergence dynamically during grokking.

\paragraph{Terminology.}
We use \emph{parameter-space superposition} to denote the coexistence of multiple algorithmic solutions within overlapping low-dimensional trajectory subspaces, distinct from the activation-space superposition studied in prior interpretability work \citep{elhage2022superposition}.
Concretely, ``superposition'' in our context refers to three related but distinct phenomena: (i)~parameter-space overlap, where weight matrices encode multiple circuits at full rank; (ii)~PCA compression, where $k^* \ll P$ trajectory directions suffice for reconstruction; and (iii)~multi-task coexistence, where distinct algorithms share the same subspace without destructive interference.
These are complementary facets of the same underlying geometric structure.

\subsection{Trajectory PCA and Intrinsic Dimensionality}

Let $\Delta\theta(t) = \theta(t) - \theta(0)$ denote the parameter displacement from initialization.
For each training run, we perform an uncentered SVD decomposition of the full training trajectory,
\begin{equation}
    \Delta\theta(T) = \sum_{i=1}^{r} \sigma_i \mathbf{u}_i \mathbf{v}_i^\top,
\end{equation}
and define the rank-$k$ approximation $\Delta\theta_k = \sum_{i=1}^{k} \sigma_i \mathbf{u}_i \mathbf{v}_i^\top$.
We reconstruct compressed models via $\theta_k = \theta(0) + \Delta\theta_k$ and evaluate generalization performance as a function of $k$.

We define $k^*$ as the minimal number of principal components required for \emph{all} tasks to exceed 90\% test accuracy.
Across three-task grokking experiments and five weight decay values, we find $k^* \in [4, 8]$, despite models containing over $3 \times 10^5$ parameters.
Thus, the final generalizing solution occupies an extremely low-dimensional subspace of parameter space.

\subsection{Sharp Reconstruction Phase Transition}
\label{sec:sharp_transition}

Reconstruction accuracy as a function of $k$ exhibits a sharp phase transition \citep[cf.][]{montanari2026phase} (see \Cref{fig:app_kstar_comparison,fig:app_heatmap_comparison} and \Cref{tab:app_tritask_acc,tab:app_dualtask_acc} in the Appendix for detailed per-$k$ accuracy breakdowns):
\begin{itemize}
    \item For $k < k^*$, performance is near chance.
    \item For $k \geq k^*$, accuracy jumps abruptly to near-optimal levels.
\end{itemize}
No gradual improvement regime is observed.
Instead, generalization emerges once a critical subspace dimension is reached.

Notably, the leading principal component (PC1) typically captures 70--90\% of trajectory variance yet carries \emph{no} generalization capability.
This direction corresponds primarily to memorization and shortcut fitting.
Algorithmic information resides in subdominant components.
This separation implies that \textbf{variance explained is not a reliable proxy for functional relevance}.

\subsection{Superposition of Algorithmic Features}
\label{sec:algo_superposition}

Each task in our multi-task setting (addition, multiplication, squaring) admits a distinct algorithmic solution.
The observation that all three tasks are recovered within $k^* \approx 5$--$8$ directions implies that multiple algorithms coexist in a shared low-dimensional subspace.

Geometrically, the learned solution satisfies
\begin{equation}
    \theta(T) \in \theta(0) + \mathrm{span}\{\mathbf{u}_1, \dots, \mathbf{u}_{k^*}\},
\end{equation}
where each task corresponds to a partially overlapping linear combination of basis directions.
We interpret this as \textbf{algorithmic superposition in parameter space}: multiple computational programs are implemented via overlapping parameter directions rather than disjoint modules.

\subsection{Role of Regularization: Compression Pressure}
\label{sec:compression_pressure}

The superposition dimensionality $k^*$ depends systematically on weight decay (\Cref{tab:tri_kstar,tab:dual_kstar}):
higher weight decay yields smaller $k^*$, forcing solutions into fewer directions;
lower decay permits higher-dimensional representations.
Concretely, WD $\geq 0.5$ produces $k^* \approx 4.3$, while WD $= 0.3$ requires $k^* \approx 6.3$.

Thus, weight decay acts as a \textbf{compression pressure} that regulates the density of superposed features.
Strong regularization enforces tighter packing of algorithms, while weak regularization allows partial separation.
This mirrors activation-space superposition results, but here emerges directly in parameter space.

\subsection{Constraint-Induced Compression: Dual-Task vs.\ Tri-Task}
\label{sec:constraint_compression}

\begin{figure}[t]
    \centering
    \begin{subfigure}[t]{0.48\textwidth}
        \centering
        \includegraphics[width=\textwidth]{figTHR_B_threshold_vs_wd.png}
        \caption{Dual-task $k^*$ vs.\ WD: monotonically increasing at lower WD, with seed-level scatter.}
        \label{fig:dual_kstar_wd}
    \end{subfigure}
    \hfill
    \begin{subfigure}[t]{0.48\textwidth}
        \centering
        \includegraphics[width=\textwidth]{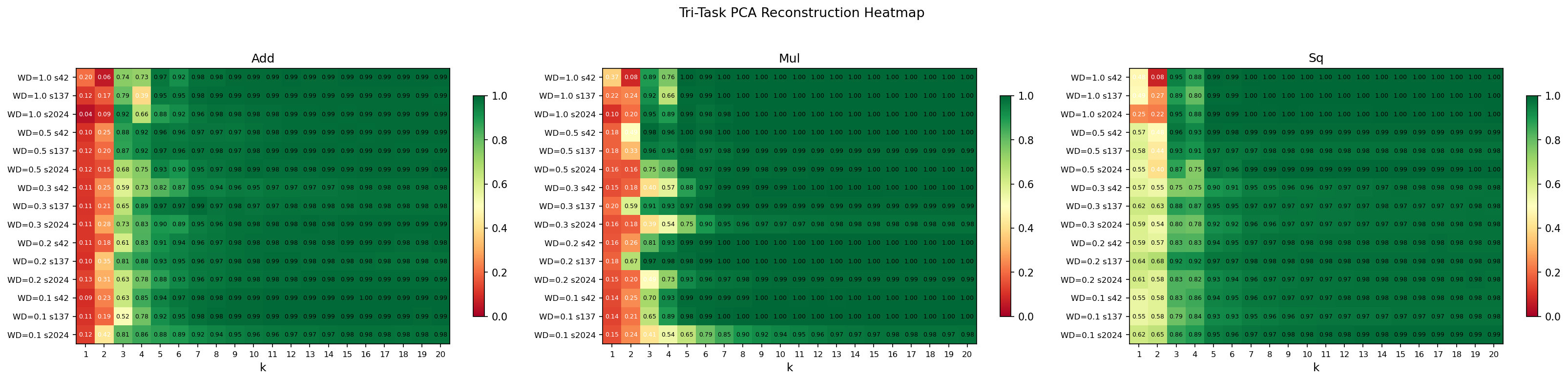}
        \caption{Tri-task reconstruction heatmap: fewer components needed, with a sharper chance-to-perfect transition.}
        \label{fig:tri_kstar_heatmap}
    \end{subfigure}
    \caption{Constraint-induced compression. \textbf{(a)}~Dual-task models require $k^* = 5$--$10$ PCA directions depending on WD, with the orthogonal complement carrying 0.2--0.8\% of trajectory variance. \textbf{(b)}~Tri-task models require fewer components ($k^* = 3$--$8$), consistent with stronger superposition pressure from additional task constraints. See \Cref{tab:app_dualtask_acc,tab:app_tritask_acc} for per-$k$ accuracy breakdowns.}
    \label{fig:kstar_comparison}
\end{figure}

Dual-task models systematically require more PCA components than tri-task models under comparable weight decay (\Cref{tab:dual_kstar,tab:tri_kstar,fig:kstar_comparison}).
For $\lambda \in [0.1, 1.0]$, tri-task models typically require $k^* \approx 4$--$6$ components, while dual-task models require $k^* \approx 5$--$8$.
Lower weight decay further increases $k^*$ in both settings.
This indicates that solution dimensionality increases when task constraints are relaxed.

With fewer concurrent objectives, the optimizer faces weaker representational constraints and can distribute algorithmic structure across additional parameter directions.
In contrast, tri-task training compresses solutions into fewer directions, suggesting stronger superposition pressure.
Geometrically, each additional task eliminates degrees of freedom in the execution subspace, forcing tighter packing of the constituent algorithms.
The result is a more compact representation that concentrates generalization-relevant information into fewer principal components.

This difference in compression explains the robustness contrast observed in orthogonal-gradient ablation (\Cref{sec:transverse_ablation}).
Dual-task models exhibit partial recovery under extreme deletion ($\alpha = 0.50$), consistent with manifold redundancy: the solution is spread across enough directions that alternative pathways remain accessible even under severe perturbation.
Tri-task models do not recover at any deletion level $\geq 10\%$, indicating tighter packing and reduced slack.
This supports a \textbf{constraint-induced compression hypothesis}: stronger task coupling produces more compressed representations, but at the cost of increased fragility.

Despite differing solution dimensionality, both regimes exhibit persistent non-convex curvature in the orthogonal complement (\Cref{sec:hessian}), indicating that compression does not eliminate saddle structure.

\subsection{Dynamic Emergence of Superposition}
\label{sec:dynamic_emergence}

Combining trajectory PCA with defect and Hessian analyses reveals a consistent developmental picture:

\begin{enumerate}
    \item \textbf{Early phase (memorization):} Training proceeds primarily along PC1. The trajectory is effectively rank-1 and dominated by shortcut fitting.
    \item \textbf{Intermediate phase (feature expansion):} Additional directions gradually activate. Commutator defect and negative curvature accumulate orthogonally to the main trajectory.
    \item \textbf{Late phase (superposition completion):} Once the trajectory spans $k^*$ directions, the system escapes the memorization saddle and generalization emerges.
\end{enumerate}
Thus, superposition does not arise at initialization but is \emph{constructed dynamically} during training through gradual dimensional expansion.

\subsection{Relation to Execution Manifold and Integrability}
\label{sec:relation_manifold}

Earlier sections showed that optimization trajectories remain confined to a low-dimensional execution manifold and exhibit near-integrable dynamics.
The present results clarify an apparent paradox:
\begin{itemize}
    \item The training \emph{path} is effectively one-dimensional.
    \item The terminal \emph{solution} has a $k^*$-dimensional tangent structure.
\end{itemize}
Training follows a narrow channel through parameter space, but converges to a multi-feature basin whose local geometry supports several overlapping algorithms.
Commutator defects remain orthogonal to the execution manifold, indicating that feature interactions occur primarily in transverse directions.

\subsection{Holographic Incompressibility Revisited}
\label{sec:holo_revisited}

Although the solution lies in a low-dimensional subspace, it cannot be compressed below $k^*$ without catastrophic loss of performance.
Below this threshold, generalization collapses abruptly (\Cref{sec:sharp_transition}).

We refer to this phenomenon as \textbf{holographic incompressibility}: the essential algorithmic information appears to be near-maximally compressed within the execution subspace, and further dimensional reduction destroys global function.
This contrasts with classical low-rank compression results, which often preserve performance under much stronger reductions.

\subsection{Implications for Feature Learning}
\label{sec:implications_features}

Our results suggest a unified view of feature formation:
\begin{itemize}
    \item Features correspond to directions in parameter space.
    \item Multiple features overlap within a small subspace.
    \item Regularization controls feature packing density.
    \item Grokking corresponds to completion of a minimal superposition basis.
\end{itemize}
This provides a dynamical foundation for superposition theory and connects feature geometry to optimization dynamics.

\section{Overparameterization as Manifold Redundancy}
\label{sec:overparam}

A central empirical observation in deep learning is that heavily overparameterized models train more reliably, generalize better, and exhibit greater robustness to perturbations \citep{neyshabur2017exploring, belkin2019reconciling, kaplan2020scaling}.
While this phenomenon is well documented, mechanistic explanations remain incomplete.
We propose that overparameterization confers robustness by inducing redundancy in center manifold structure.

\subsection{Transverse Gradient Ablation}
\label{sec:transverse_ablation}

To probe the role of transverse directions in optimization, we selectively removed a fraction of gradient components orthogonal to the execution manifold,
\begin{equation}
    \mathbf{g} \;\mapsto\; P_{\mathcal{M}_c}\,\mathbf{g} + (1-\alpha)\,P_{\mathcal{M}_c^\perp}\,\mathbf{g},
\end{equation}
where $\alpha \in [0,1]$ controls deletion strength.
We then measured grokking time under varying $\alpha$ (see \Cref{tab:ortho_dual,tab:ortho_tri} for full results, and \Cref{tab:app_ortho_dual,tab:app_ortho_tri} in the Appendix for additional detail).

Across both dual-task and tri-task settings, we observe:
\begin{itemize}
    \item Small deletions (1--7\%) induce large delays (up to $+203\%$ in dual-task, $+124\%$ in tri-task).
    \item Deletions $\geq 10\%$ eliminate grokking entirely.
    \item Dual-task models exhibit partial recovery at extreme deletion ($\alpha = 0.50$).
    \item Tri-task models do not recover at any deletion level $\geq 10\%$.
\end{itemize}
These results demonstrate that grokking depends critically on rare transverse updates that constitute $<1\%$ of gradient variance.

\subsection{Fragility Threshold and Escape Connectivity}
\label{sec:fragility_threshold}

The sharp transition near $\alpha = 0.10$ suggests a fragility boundary in parameter space.
Below this threshold, narrow corridors connecting memorization saddles to generalization basins remain intact.
Above it, these corridors collapse, preventing escape.

We interpret this as a percolation-like phenomenon in optimization geometry, analogous to the mode connectivity observed in loss landscapes \citep{draxler2018essentially, garipov2018loss}: viable descent paths exist only when sufficient transverse connectivity is preserved.
The consistency of the ${\sim}10\%$ threshold across both dual-task and tri-task settings---despite different parameter counts, task complexities, and grokking timescales---suggests this boundary reflects a consistent property of the optimization landscape across tested regimes, rather than an artifact of any particular configuration.

\subsection{Redundant Center Manifolds in Overparameterized Models}
\label{sec:redundant_manifolds}

Dual-task models possess significantly more degrees of freedom than required to represent two algorithms.
As a result, multiple geometrically distinct center manifolds coexist:
\begin{equation}
    \mathcal{M}_c^{(1)},\;\mathcal{M}_c^{(2)},\;\dots
\end{equation}
When the primary escape route is disrupted (e.g., by 50\% orthogonal deletion), optimization can transition to an alternative manifold.
This explains the non-monotonic recovery observed in dual-task experiments: deletions of 10--25\% are sufficient to block the primary pathway but insufficient to open alternatives, while 50\% deletion forces a qualitative reorganization that reveals a secondary manifold.

In contrast, tri-task models require a larger minimal superposition dimension ($k^* \approx 5$--$8$ vs.\ $k^* \approx 5$--$9$ for dual-task, but with tighter packing per task).
The increased constraint eliminates redundant manifolds, yielding a single fragile generalization pathway.
Thus, excess parameters provide redundancy in center manifold structure.

\subsection{Relation to Curvature Spectrum}
\label{sec:curvature_redundancy}

Hessian analysis (\Cref{sec:hessian}) reveals that overparameterized models possess many unstable directions (negative eigenvalues).
Only a subset are utilized in constructing the dominant center manifold.

In dual-task models, additional negative curvature modes remain available after partial ablation.
These modes enable reconfiguration of the escape pathway, consistent with the observed recovery at $\alpha = 0.50$.

In tri-task models, nearly all usable modes are required to support the three-task superposition.
Ablation therefore destroys the entire escape mechanism, with no fallback.

Overparameterization thus supplies surplus unstable directions that can be repurposed during training---a geometric buffer against perturbation.

\subsection{Compression-Induced Reconfiguration}
\label{sec:compression_reconfig}

The non-monotonic recovery observed under extreme deletion in dual-task models suggests that strong projection can force optimization into simpler subspaces that still admit valid solutions.
This resembles implicit low-rank regularization \citep{li2018measuring}: excessive dimensional restriction eliminates complex manifolds but may reveal simpler alternatives.

Specifically, at $\alpha = 0.50$, the retained gradient lies almost entirely within the PCA manifold plus half the transverse component.
This constrained optimization landscape may preferentially select for solutions with lower effective dimensionality---solutions that happen to be reachable through the reduced subspace.
Tri-task models lack such simplified embeddings, explaining the absence of recovery.

\subsection{Implications for Generalization and Robustness}
\label{sec:implications_robustness}

Our results suggest that robustness in overparameterized models arises not merely from flat minima \citep{keskar2017large}, but from \textbf{geometric redundancy in optimization pathways}.
Specifically:
\begin{itemize}
    \item Multiple center manifolds provide alternative generalization routes.
    \item Redundancy buffers against noise, gradient perturbations, and partial information loss.
    \item Loss of redundancy (through increased task load or reduced parameters) induces brittle training dynamics.
\end{itemize}
This reframes overparameterization as a mechanism for \emph{structural stability}: the model is robust not because any single solution is flat, but because many viable solutions exist in the neighborhood of the optimization trajectory.

\subsection{Connections to Practical Training Methods}
\label{sec:practical_connections}

This perspective clarifies the empirical behavior of several widely used techniques:
\begin{itemize}
    \item \textbf{Pruning and sparsification} \citep{frankle2019lottery}: Remove transverse connectivity, risking collapse of escape corridors.
    Our results predict that pruning beyond ${\sim}10\%$ of transverse capacity should degrade generalization sharply.
    \item \textbf{Low-rank adaptation} \citep[LoRA;][]{hu2022lora}: Restricts the manifold dimension available for fine-tuning, limiting redundancy.
    The success of LoRA may depend on pre-trained models having already traversed the critical manifold expansion phase.
    \item \textbf{Gradient clipping}: Suppresses unstable modes that may be essential for saddle escape.
    Too aggressive clipping could eliminate the transverse signal needed for grokking-like generalization.
    \item \textbf{Adaptive optimizers} \citep{thilak2022slingshot}: Reshape the curvature spectrum, potentially redistributing gradient energy between manifold and transverse directions.
\end{itemize}
Our framework predicts that such methods succeed only when sufficient manifold redundancy remains.

\subsection{Summary}

We conclude that overparameterization enhances learning by generating redundant center manifolds that provide multiple geometrically distinct generalization pathways.
Transverse gradient ablation reveals a sharp fragility threshold ($\alpha \approx 0.10$) separating robust and brittle regimes.
Task complexity determines whether alternative manifolds exist, explaining the differential recovery behavior between dual-task (recovery at $\alpha = 0.50$) and tri-task (no recovery) settings.

\section{Discussion}
\label{sec:discussion}

\subsection{The Scaffold--Solution Duality}

\emph{Thesis: the dominant PCA directions encode the memorization scaffold, while the generalizing solution lives in the orthogonal complement.}

Our results reveal a striking duality.
The PCA manifold captures the \emph{scaffold} of the training trajectory: the dominant directions of weight change during memorization.
PC1--PC3 account for $>98\%$ of $\norm{\Delta\theta}^2$ but contribute nothing to generalization (chance-level accuracy when used for reconstruction).
The solution lives in PC$_{k^*}$--PC$_{k^*+\epsilon}$, which contribute $<1\%$ of variance but are overwhelmingly responsible for generalization.
The fine-grained orthogonal deletion experiment makes this quantitative: removing just $1\%$ of the orthogonal gradient component delays grokking by ${\sim}30\%$, while removing ${\geq}10\%$ prevents it entirely.
The dose-response relationship is remarkably consistent between dual-task and tri-task settings, with a sharp cliff at $f \approx 0.07$--$0.10$.

This is consistent with the integrability result $\rho \approx 1.000$: the commutator vectors (probing loss-landscape curvature) are predominantly orthogonal to the scaffold manifold, indicating that the curvature structure relevant to generalization resides predominantly outside the dominant trajectory.
Causal interventions in a companion study confirm this picture: amplifying non-commutativity accelerates grokking by 32--60\% across task families, while suppressing orthogonal gradients delays or prevents it \citep{xu2026earlywarning}.

\subsection{Multi-Task Geometry: Orthogonal Task Embedding}

\emph{Thesis: shared-trunk models accommodate multiple circuits via geometric separation rather than superposition.}

The near-orthogonality of task head weights ($|\cos| < 0.08$) and the low cross-task gradient cosine (0.15--0.28) reveal that the shared trunk learns a representation in which different tasks occupy non-overlapping subspaces.
This geometric separation explains the absence of destructive interference: the tasks do not compete for the same directions.

The grokking order (mul $\to$ sq $\to$ add) reflects a natural hierarchy: multiplication, as a nonlinear operation, establishes its circuit first, and the quadratic sum $x^2 + y^2$ shares enough multiplicative structure to follow closely.
Addition, algebraically distinct, is last.

The drop in PC1\% from single-task (70--94\%) to dual-task (55--77\%) to tri-task (49--65\%) is consistent with the manifold needing additional dimensions to accommodate each circuit.
Yet the manifold remains strongly low-dimensional---far from the full parameter space---and the integrability property is preserved exactly.

\subsection{Weight Decay as a Phase Parameter}

\emph{Thesis: weight decay controls grokking through a phase diagram with distinct dynamical regimes.}

The five-point WD sweep reveals that weight decay is not merely a regularization knob but a \emph{phase parameter} governing qualitatively different grokking dynamics:
\begin{itemize}
    \item \textbf{High WD ($\lambda \geq 0.5$)}: Fast grokking (${\sim}16$k steps), deep Hessian curvature ($\lambda_\text{min} \approx -36$ to $-63$), smooth defect accumulation, small dual-task lead (5--53\%) but large tri-task lead ($77$--$82\%$), tight reconstruction ($k^* = 5$--$6$).
    \item \textbf{Intermediate WD ($\lambda = 0.2$--$0.3$)}: Moderate grokking (${\sim}28$--$45$k), curvature plateau ($\lambda_\text{min} \approx -15$), intermittent defect dynamics, larger lead (32--61\%), moderate $k^*$ ($6$--$8$).
    \item \textbf{Low WD ($\lambda = 0.1$)}: Slow grokking (${\sim}98$k dual-task, ${\sim}152$k tri-task), shallow curvature ($\lambda_\text{min} \approx -17$), spiky transient defect patterns, very large lead (70\% dual-task, 97\% tri-task), high $k^*$ ($7$--$9$).
    \item \textbf{No WD ($\lambda = 0$)}: No grokking despite substantial negative curvature ($\lambda_\text{min} \approx -50$).
\end{itemize}

The $\lambda = 0$ outlier is critical: it demonstrates that saddle curvature alone is insufficient.
Weight decay provides the regularization pressure---the force that pushes the trajectory off the memorization saddle and toward lower-norm generalizing basins.
Without it, the model sits at the saddle indefinitely.

The intra-signal gap framework~\citep{xu2026spectral_edge} provides a spectral account of this mechanism: weight decay promotes the spectral gap opening at $k^* = 1$ that stabilises the dominant attention direction, while no-weight-decay runs remain in the degenerate regime with no gap dynamics and no generalisation (see \Cref{sec:gram_matrix} for the full quantitative analysis).

\subsection{Incompressibility and the Nature of Algorithmic Solutions}

\emph{Thesis: grokking solutions are holographic---distributed across all parameters at full rank with no redundancy.}

The failure of every compression method (per-layer SVD, magnitude pruning, uniform scaling, module ablation) establishes that the grokking solution has qualitatively different structure from typical neural network solutions.
In standard deep learning, low-rank approximations and pruning routinely preserve accuracy \citep{frankle2019lottery, hu2022lora}.
Here, the modular arithmetic circuits use the \emph{full capacity} of each weight matrix (rank-128 required, rank-64 fails), with every module necessary and extreme sensitivity to perturbation ($\pm 5\%$ scaling $\to$ chance).

The apparent paradox---that the training trajectory is low-rank (PC1 captures 94\% of variance) while the solution is full-rank---resolves through the scaffold--solution duality.
The trajectory's dominant directions are the path the optimizer takes (the memorization scaffold), not the destination.
The generalizing solution is encoded in the tiny, seemingly negligible residual orthogonal to this scaffold.

\paragraph{A curious non-monotonicity.}
The dual-task orthogonal deletion experiment reveals a surprising non-monotone pattern: deletions of 10--25\% kill grokking, but 50\% deletion recovers it (at a $163\%$ delay).
One interpretation is that large deletions so drastically change the optimization landscape that the model finds an alternative route to generalization---one not relying on the specific orthogonal structure of the original trajectory.
This non-monotonicity was not observed in the tri-task setting (where all deletions $\geq 10\%$ fail), suggesting that the additional task constraints eliminate alternative grokking pathways.

\subsection{Relation to Feature-Based Interpretability}
\label{sec:feature_interp}

Recent work studies superposition via sparse autoencoders (SAEs) and feature dictionaries in activation space \citep{elhage2022superposition, bricken2023monosemanticity}, decomposing internal representations at fixed checkpoints into interpretable directions.
Our results are complementary: rather than decomposing representations at fixed snapshots, we analyze how the parameter trajectory dynamically constructs a low-dimensional basis over the course of training.
The execution manifold we identify is a \emph{parameter-space} object, while SAE features are \emph{activation-space} objects; the two perspectives address different facets of superposition.

A natural question is whether the PCA trajectory directions we identify correspond to interpretable features in the SAE sense.
Preliminary observations suggest partial alignment: the reconstruction threshold $k^*$ increases with task count (from ${\sim}5$ in dual-task to ${\sim}7$ in tri-task), consistent with each task contributing additional feature directions.
However, the dominant PCA directions encode the memorization scaffold rather than the generalizing solution, complicating direct correspondence.
Future work could connect these approaches by applying SAE decomposition along the execution manifold, tracking feature emergence and dissolution as a function of training progress.

\subsection{Limitations}

Our multi-task experiments are limited to 2--3 tasks on modular arithmetic mod~97 with small Transformers (${\sim}300$k parameters).
Whether the scaling of manifold rank with task count, the holographic incompressibility, and the WD phase diagram generalize to larger models and real-world tasks remains open.
We do not study models beyond ${\sim}300$k parameters; larger architectures may exhibit qualitatively different manifold structure \citep{kaplan2020scaling}, and the sharp fragility threshold we observe could shift or broaden with scale.
We do not study natural language or vision tasks, where optimization noise, data heterogeneity, and the absence of clean algorithmic structure may obscure or eliminate the low-dimensional manifold geometry reported here, though a companion study \citep{xu2026earlywarning} finds that commutator defect onset reliably precedes grokking in Dyck languages and the SCAN compositional-generalization benchmark, suggesting the geometric mechanism extends beyond modular arithmetic.
The Hessian analysis uses power iteration (sensitive to initialization and convergence), and the per-task eigenvalue comparison does not account for cross-task Hessian interactions.
The orthogonal deletion experiment uses a single seed (42) and a single WD value (1.0); extending to multiple seeds and WD values would strengthen the result, though the dose-response consistency across both dual-task and tri-task settings is encouraging.

\section{Conclusion}
\label{sec:conclusion}

We have extended the geometric analysis of grokking from single-task to multi-task modular arithmetic, revealing several new phenomena.
\textbf{First}, multi-task grokking produces a staggered ordering (mul $\to$ sq $\to$ add) with task-specific heads that are nearly orthogonal, showing that shared-trunk models accommodate multiple circuits via geometric separation.
\textbf{Second}, the execution manifold is empirically integrable ($\rho \approx 1.000$) across all task counts, weight decay values, and seeds---the most consistent invariant we have identified.
\textbf{Third}, weight decay controls grokking through a structured phase diagram: saddle curvature depth, grokking timescale, defect lead, and reconstruction threshold all covary systematically with $\lambda$, with a critical regime near $\lambda \approx 0.3$ and a decoupled $\lambda = 0$ outlier that has curvature but no grokking.
Defect onset precedes grokking in 42/42 conditions across both dual-task and tri-task settings ($p = 2^{-27}$ for tri-task alone), with the tri-task lead fraction reaching $97\%$ at $\lambda = 0.1$.
\textbf{Fourth}, the grokking solution is holographic---distributed across all parameters at full rank, incompressible by any post-hoc method, and exquisitely sensitive to orthogonal gradient perturbation.
Fine-grained dose-response experiments reveal a continuous delay function culminating in a sharp cliff at ${\sim}10\%$ deletion, with the tri-task setting being strictly more fragile than dual-task.
The dominant training trajectory is a scaffold; the solution is the tiny residual.
\textbf{Fifth}, multiple algorithmic solutions coexist within a shared subspace of only 4--8 PCA directions, revealing parameter-space superposition whose density is controlled by weight decay.
\textbf{Sixth}, the differential response to transverse ablation---dual-task models recover at extreme deletion via alternative center manifolds, while tri-task models do not---reveals that overparameterization confers robustness through geometric redundancy in optimization pathways.
Together, these results paint a picture of multi-task grokking as a geometrically structured phase transition, governed by the interplay of saddle curvature, weight decay pressure, holographic solution encoding, and manifold redundancy.

\paragraph{Reproducibility.}
All code and figures are available at \url{https://github.com/skydancerosel/grokking-integrability}.

\bibliographystyle{plainnat}

\appendix
\section{Additional Figures}
\label{app:figures}

\begin{figure}[ht]
    \centering
    \begin{subfigure}[t]{0.48\textwidth}
        \centering
        \includegraphics[width=\textwidth]{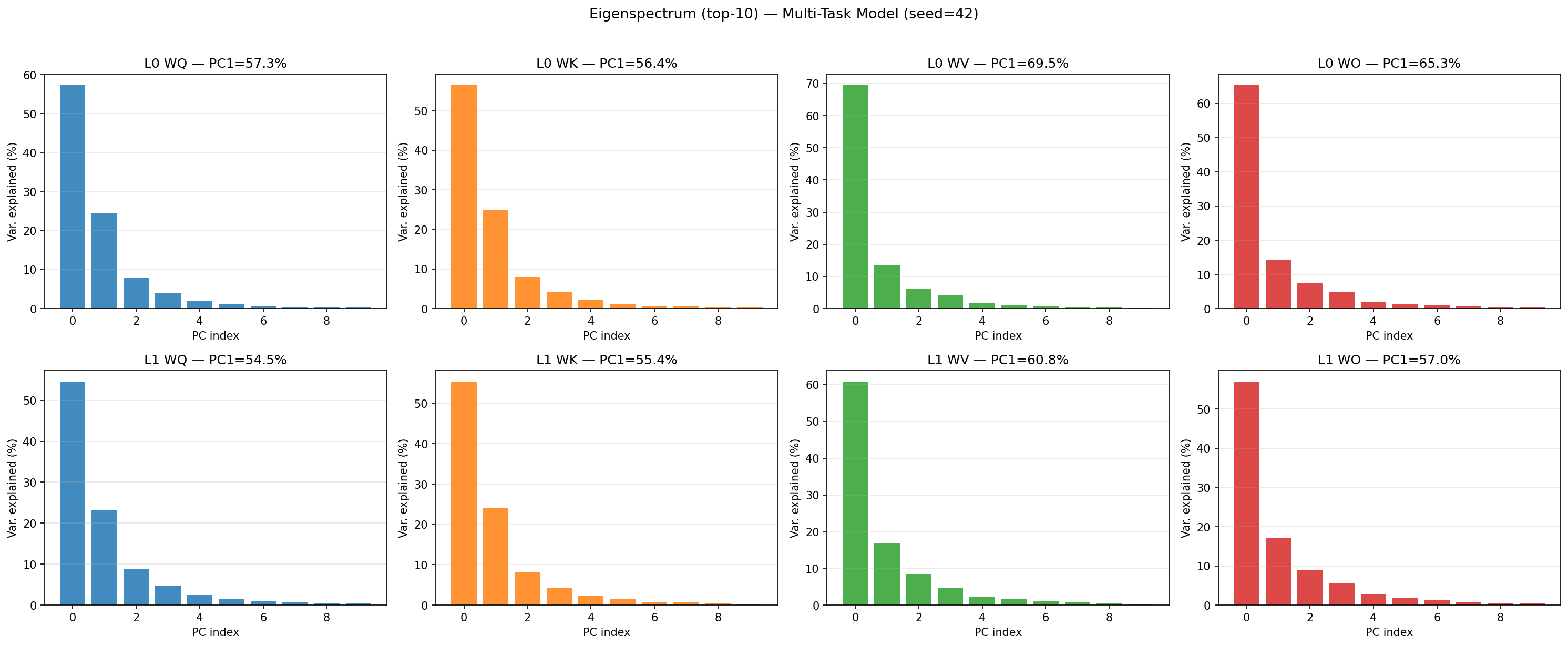}
        \caption{Dual-task eigenspectrum (seed~42).}
        \label{fig:app_dual_eigen}
    \end{subfigure}
    \hfill
    \begin{subfigure}[t]{0.48\textwidth}
        \centering
        \includegraphics[width=\textwidth]{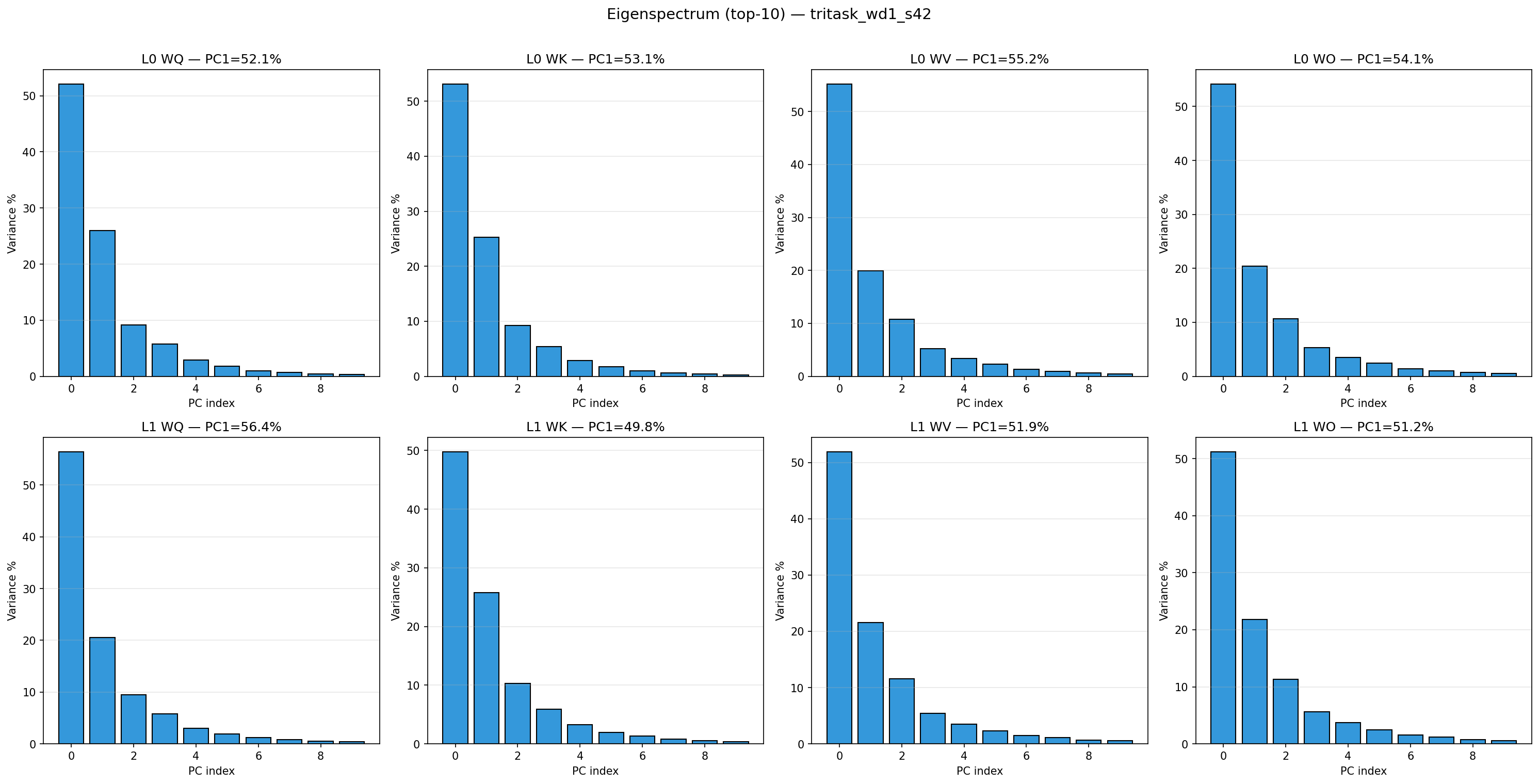}
        \caption{Tri-task eigenspectrum (seed~42).}
        \label{fig:app_tri_eigen}
    \end{subfigure}
    \caption{Top-10 eigenspectra for dual-task and tri-task. Both show a dominant first eigenvalue, but the gap is smaller in multi-task settings.}
    \label{fig:app_eigenspectra}
\end{figure}

\begin{figure}[ht]
    \centering
    \begin{subfigure}[t]{0.48\textwidth}
        \centering
        \includegraphics[width=\textwidth]{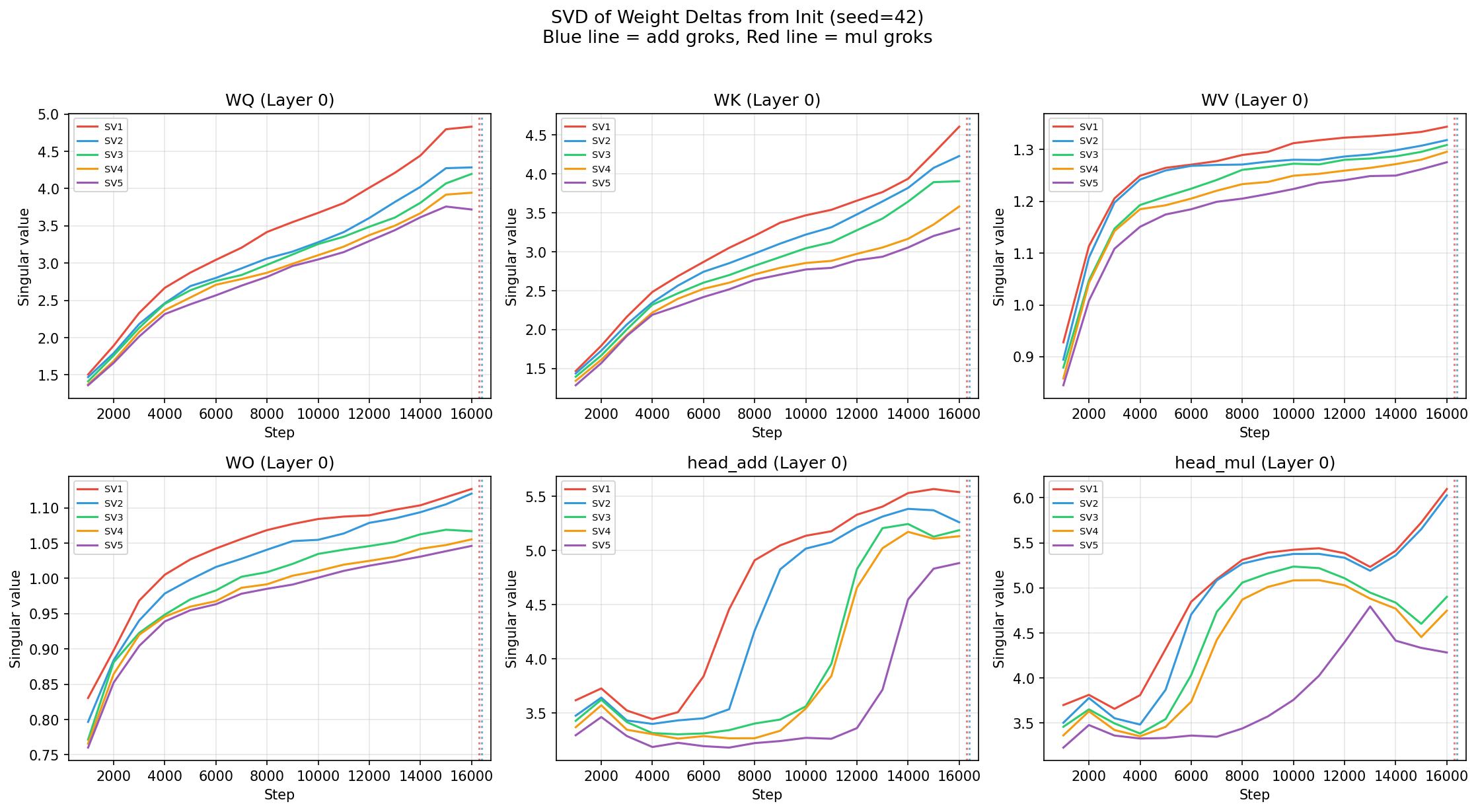}
        \caption{Dual-task SVD of weight deltas (seed~42).}
        \label{fig:app_dual_svd}
    \end{subfigure}
    \hfill
    \begin{subfigure}[t]{0.48\textwidth}
        \centering
        \includegraphics[width=\textwidth]{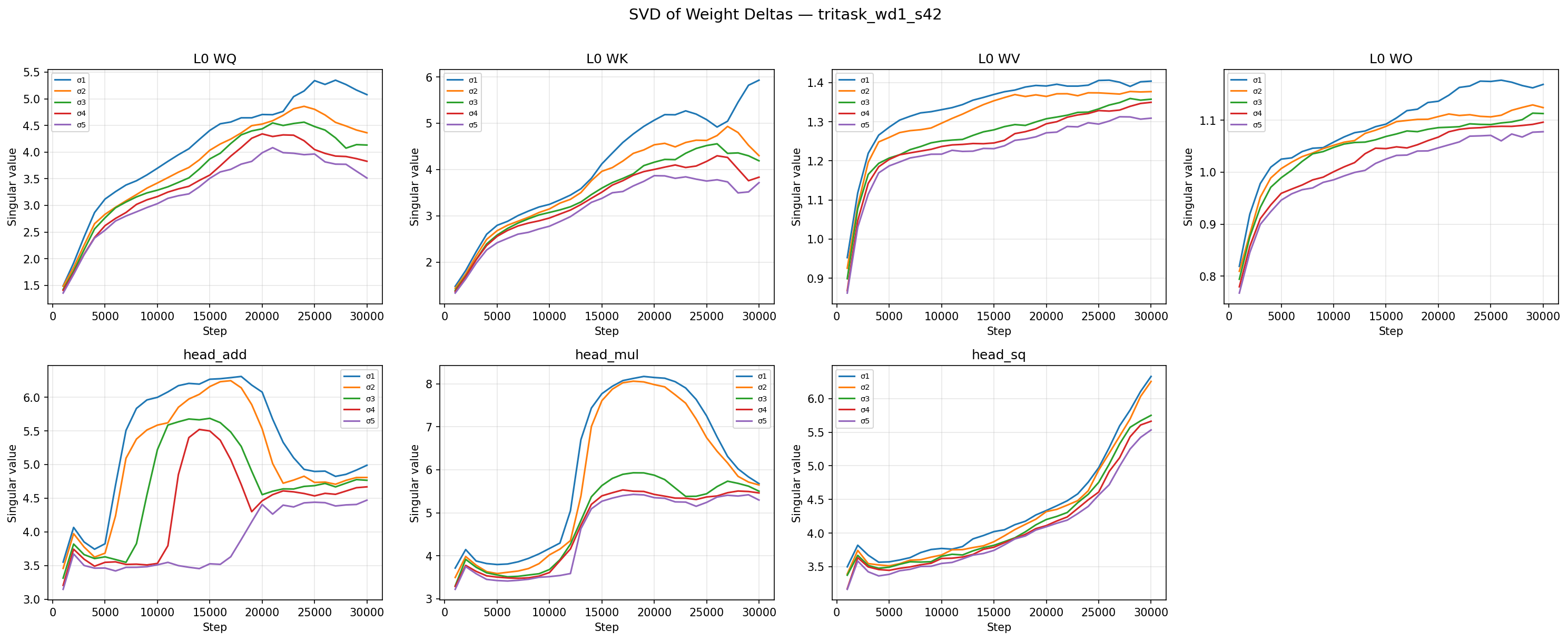}
        \caption{Tri-task SVD of weight deltas (seed~42).}
        \label{fig:app_tri_svd}
    \end{subfigure}
    \caption{SVD of weight deltas. Top-5 singular values grow concurrently in multi-task settings, with head-specific SVDs showing growth patterns at different timings consistent with staggered grokking.}
    \label{fig:app_svd}
\end{figure}

\begin{figure}[ht]
    \centering
    \begin{subfigure}[t]{0.48\textwidth}
        \centering
        \includegraphics[width=\textwidth]{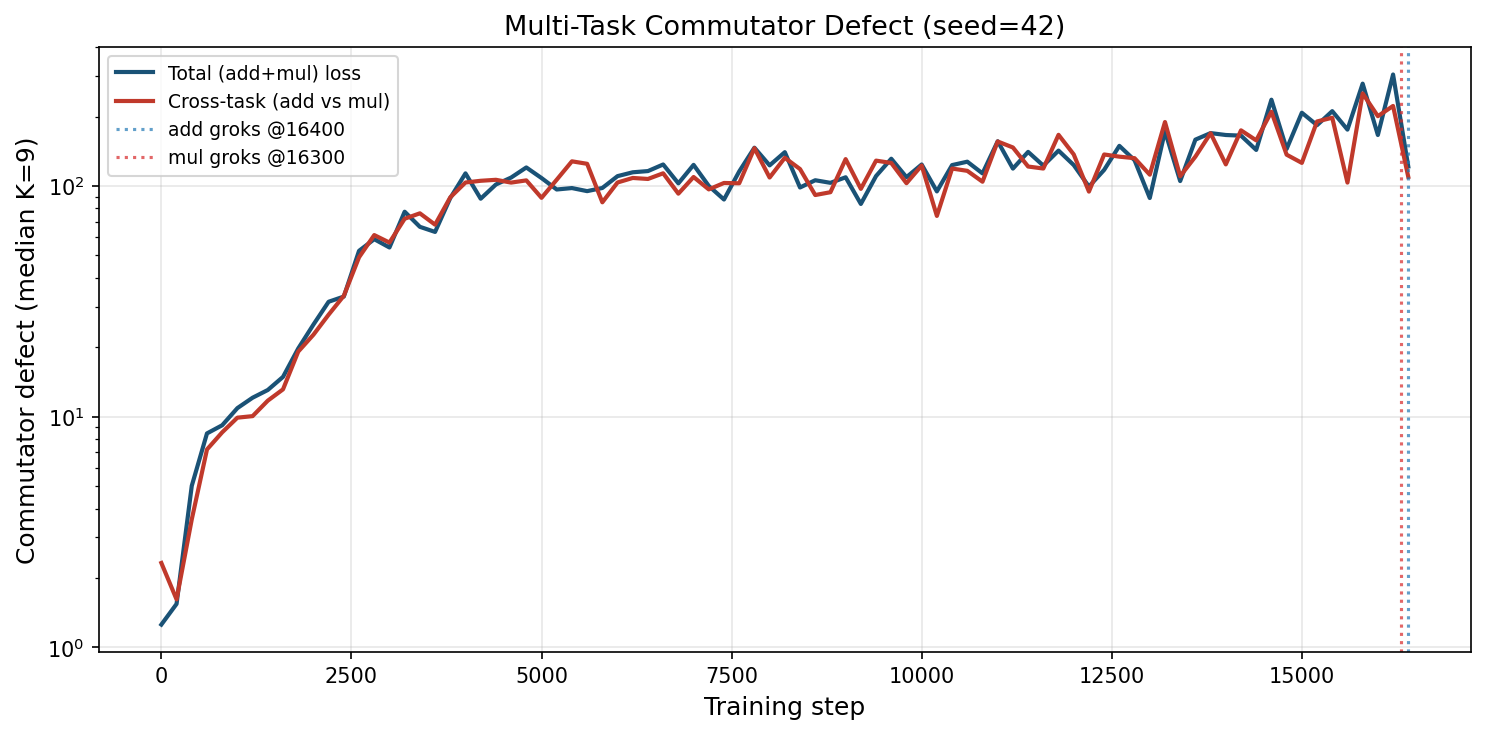}
        \caption{Dual-task commutator defect (seed~42, WD=1.0).}
        \label{fig:app_dual_defect}
    \end{subfigure}
    \hfill
    \begin{subfigure}[t]{0.48\textwidth}
        \centering
        \includegraphics[width=\textwidth]{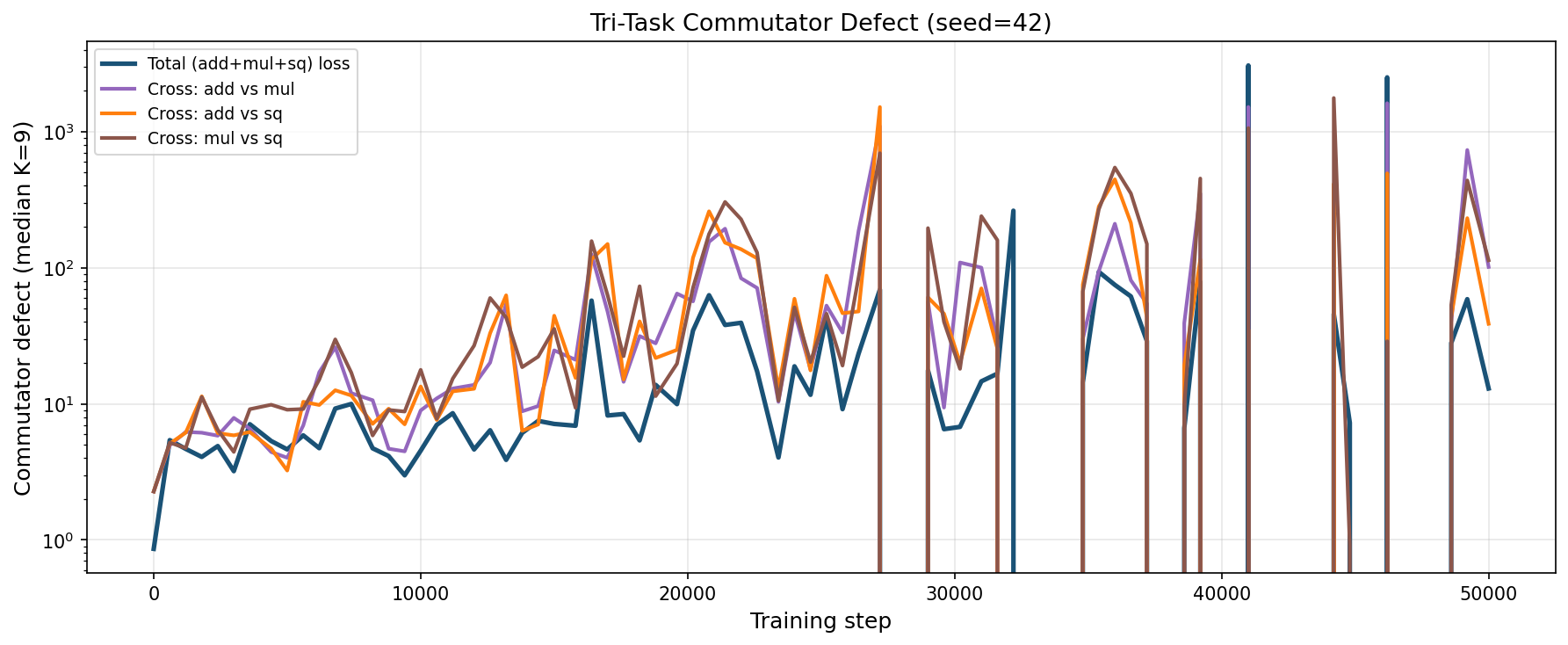}
        \caption{Tri-task commutator defect (seed~42, WD=1.0).}
        \label{fig:app_tri_defect}
    \end{subfigure}
    \caption{Commutator defect time series for both multi-task settings.}
    \label{fig:app_defect}
\end{figure}

\begin{figure}[ht]
    \centering
    \includegraphics[width=\textwidth]{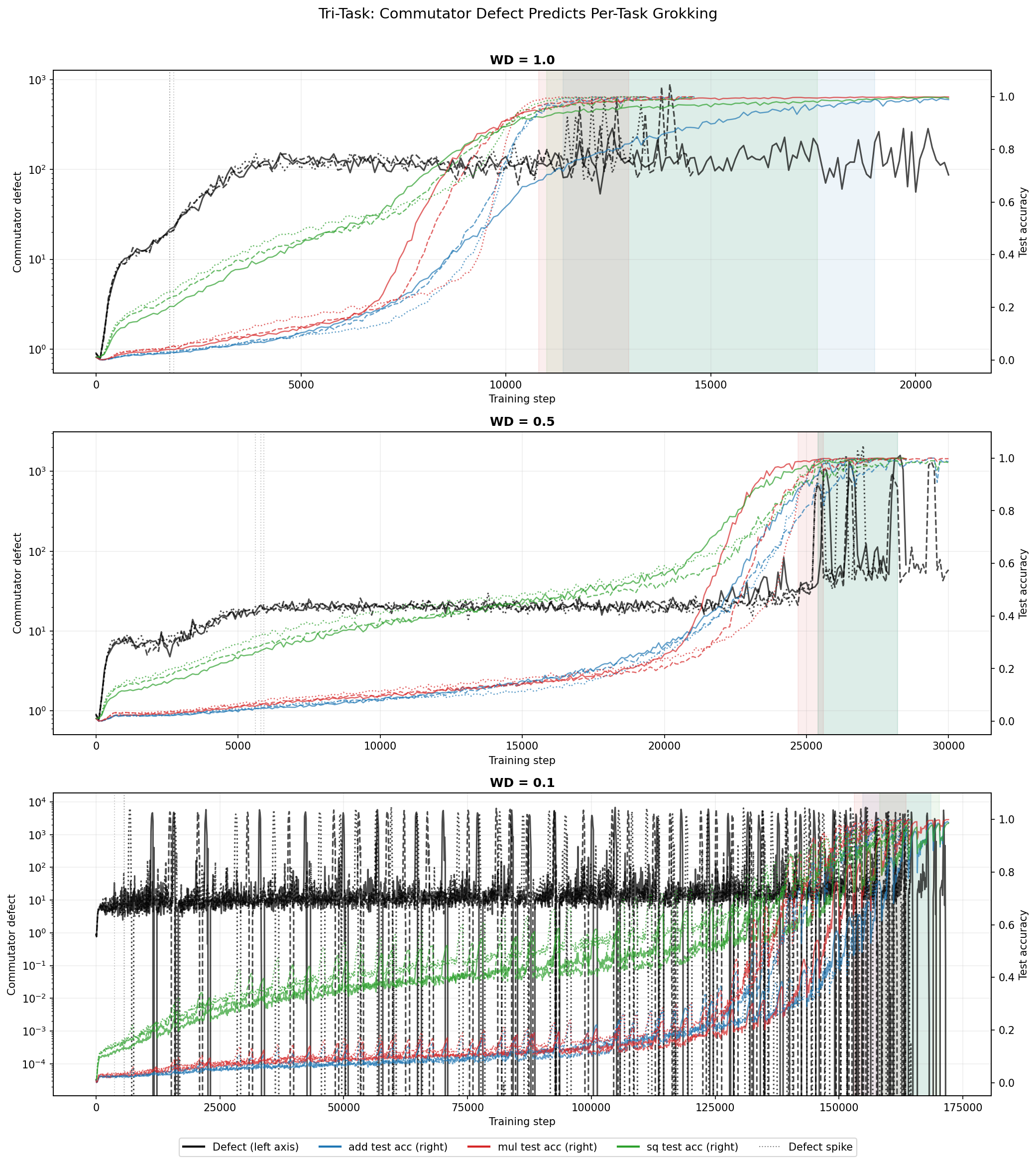}
    \caption{Tri-task defect vs.\ per-task test accuracy across three WD values ($\lambda = 1.0, 0.5, 0.1$), 3 seeds overlaid per panel. Black curves (left axis): commutator defect on log scale. Colored curves (right axis): per-task test accuracy (blue = add, red = mul, green = sq). Shaded bands mark the grok region for each task. Dotted verticals mark defect spike detection. The defect spike clearly precedes grokking in all panels, with the temporal gap increasing at lower WD.}
    \label{fig:app_tritask_defect_acc}
\end{figure}

\begin{figure}[ht]
    \centering
    \includegraphics[width=0.8\textwidth]{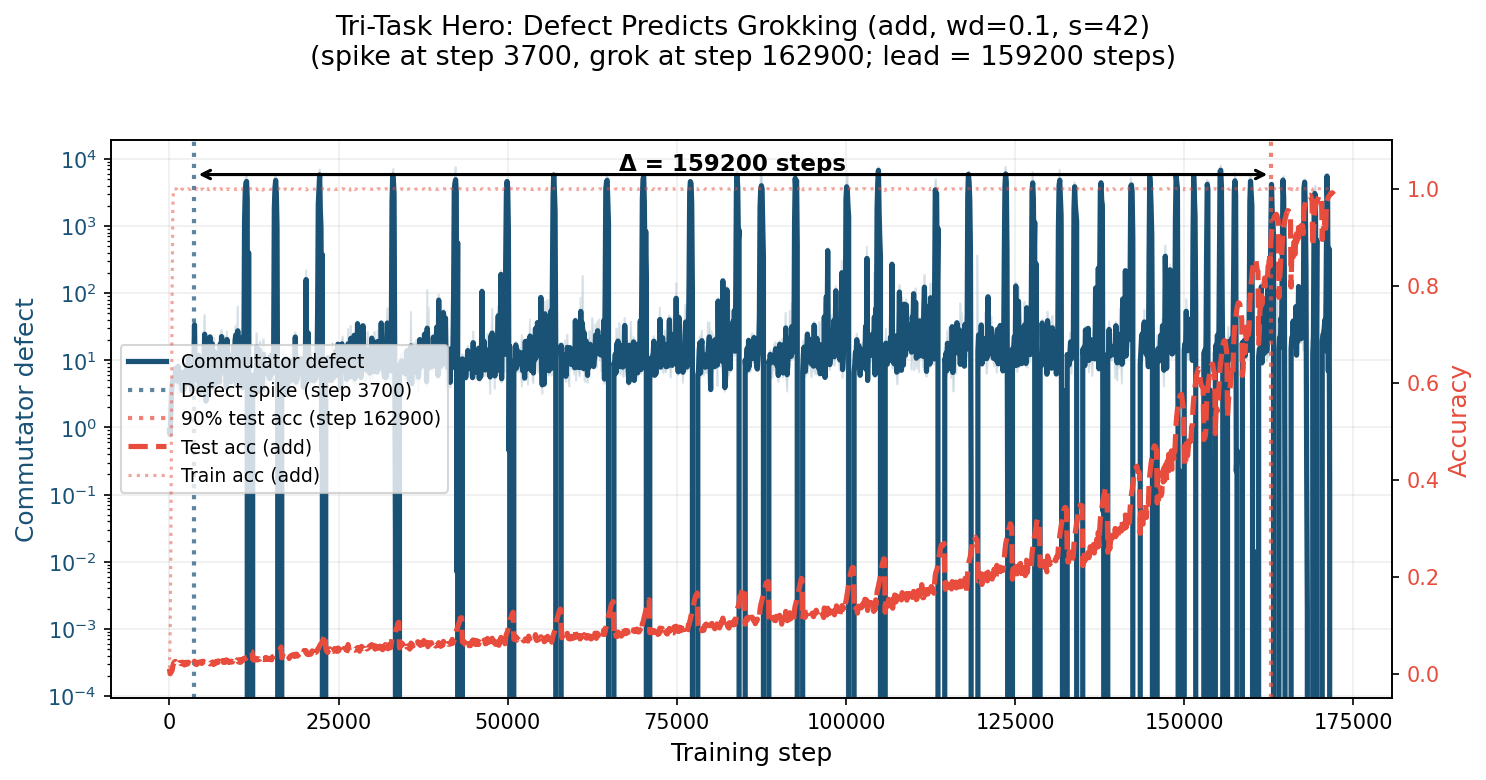}
    \caption{Hero figure: defect predicts grokking in the tri-task setting (add task, $\lambda = 0.1$, seed~42). Blue: commutator defect (left axis, log scale) with IQR ribbon. Red: test and train accuracy (right axis). The defect spike occurs at step~$3{,}700$, while 90\% test accuracy is reached at step~$162{,}900$---a lead time of $159{,}200$ steps ($97.7\%$ of the grokking time).}
    \label{fig:app_tritask_hero}
\end{figure}

\begin{figure}[ht]
    \centering
    \begin{subfigure}[t]{0.48\textwidth}
        \centering
        \includegraphics[width=\textwidth]{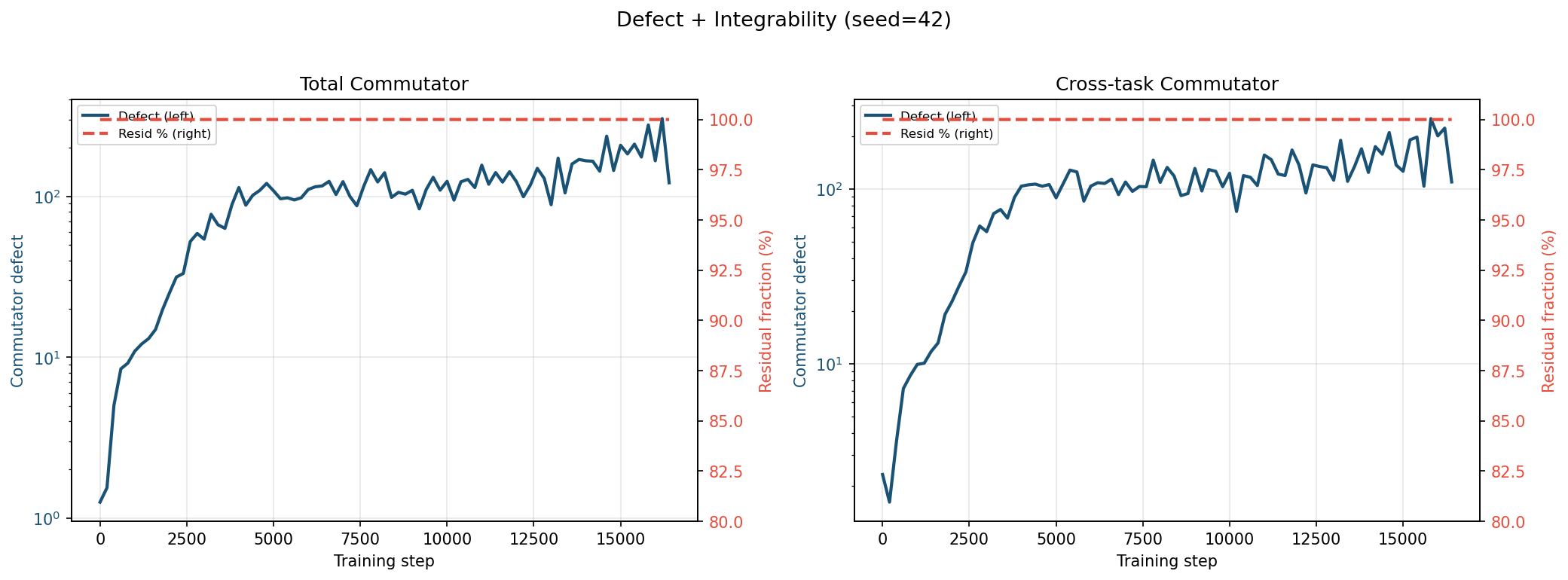}
        \caption{Dual-task combined: defect + integrability (seed~42).}
        \label{fig:app_dual_combined}
    \end{subfigure}
    \hfill
    \begin{subfigure}[t]{0.48\textwidth}
        \centering
        \includegraphics[width=\textwidth]{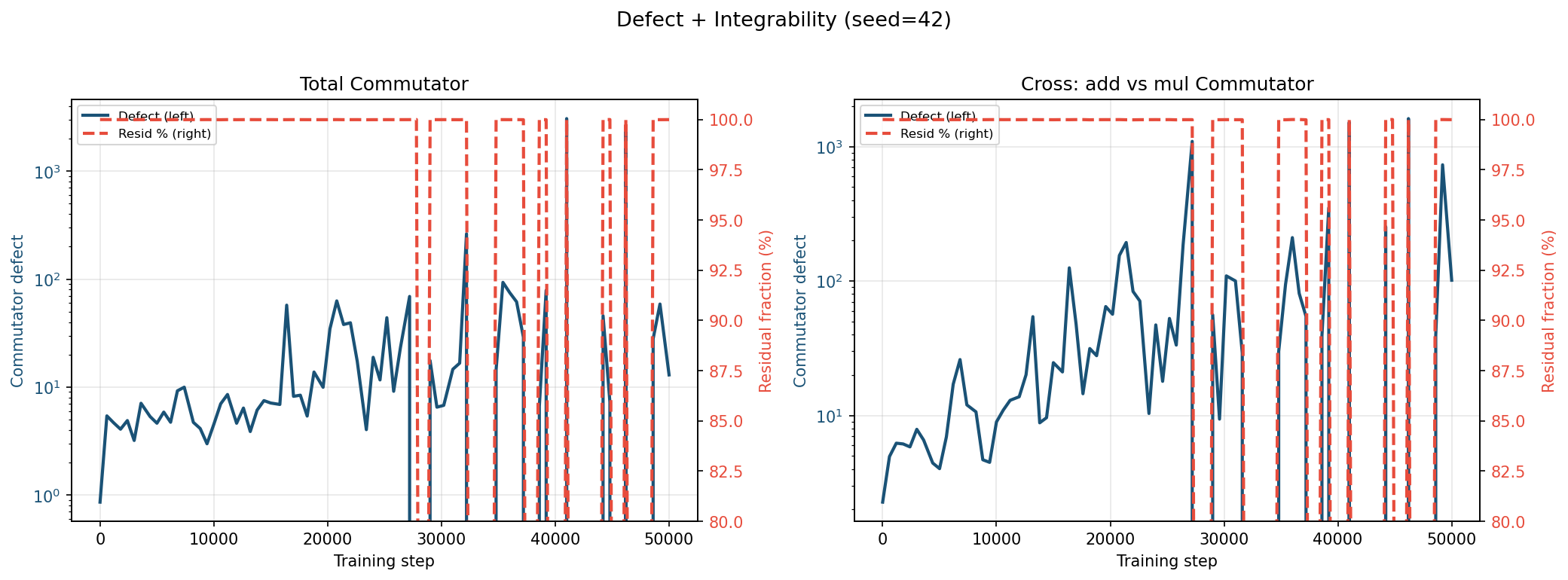}
        \caption{Tri-task combined: defect + integrability (seed~42).}
        \label{fig:app_tri_combined}
    \end{subfigure}
    \caption{Combined defect magnitude and integrability over training. Integrability remains at 1.000 throughout while defect varies.}
    \label{fig:app_combined}
\end{figure}

\begin{figure}[ht]
    \centering
    \includegraphics[width=0.7\textwidth]{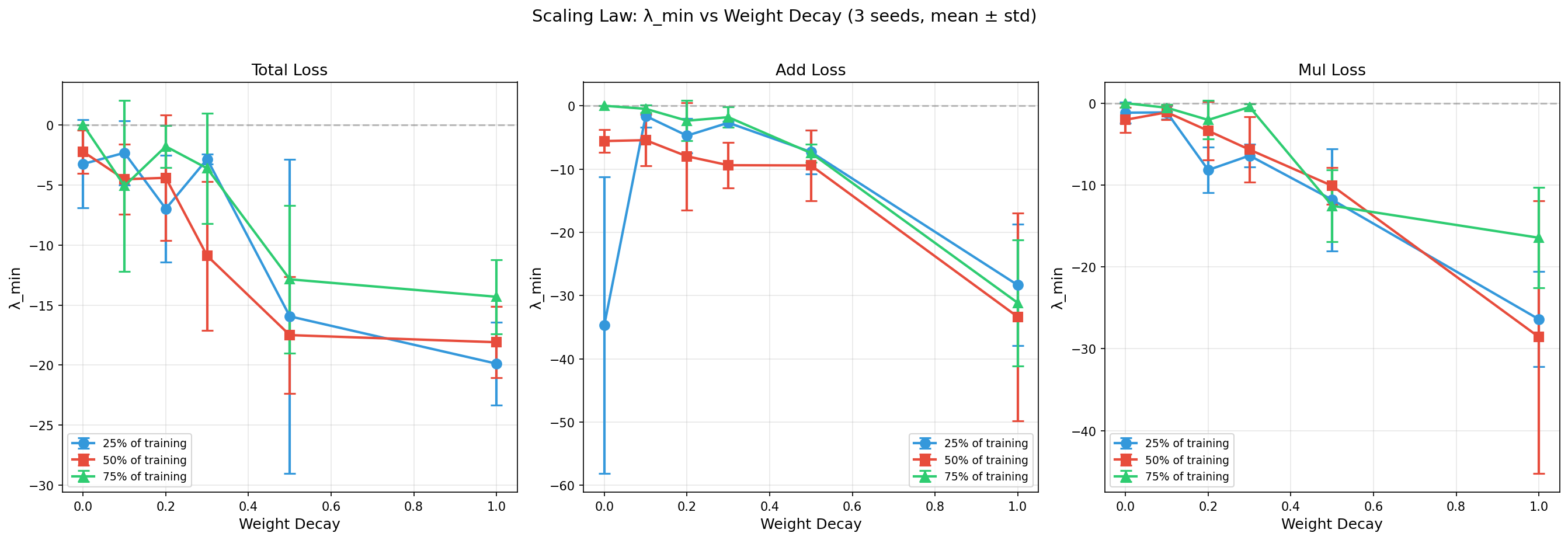}
    \caption{Hessian curvature scaling across WD values (multi-seed). Curvature depth increases monotonically with WD in the grokking regime.}
    \label{fig:app_hessian_scaling}
\end{figure}

\begin{figure}[ht]
    \centering
    \includegraphics[width=0.7\textwidth]{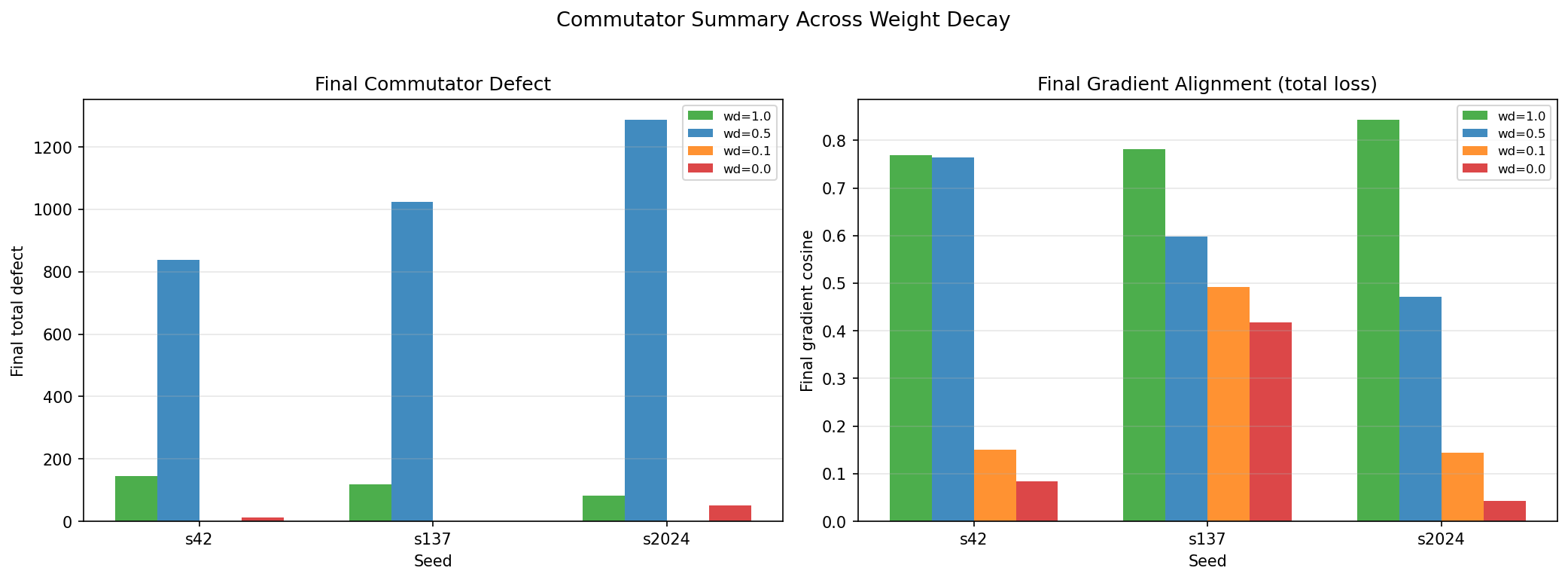}
    \caption{Tri-task cross-WD defect summary. WD=0.5 shows the largest final defect (analysis captures the model mid-transition), while WD=0.1 collapses to zero post-grokking.}
    \label{fig:app_tri_wd_defect}
\end{figure}

\begin{figure}[ht]
    \centering
    \includegraphics[width=0.7\textwidth]{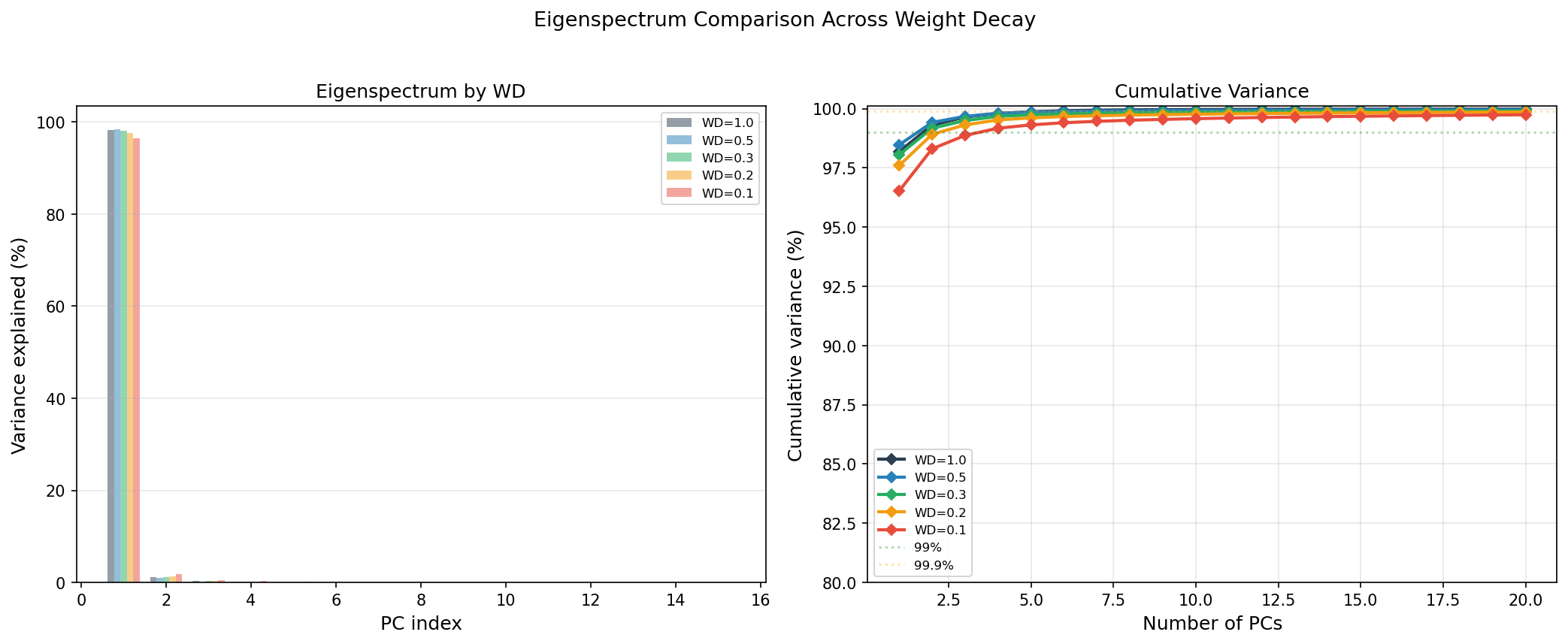}
    \caption{Dual-task PCA eigenspectrum across weight decay values.
    The top 5--10 eigenvalues capture $>$99\% of trajectory variance;
    the remaining eigenvalues form a rapidly decaying tail.
    The number of non-negligible components increases at lower WD,
    consistent with the $k^*$ trend reported in \Cref{sec:constraint_compression}.
    See \Cref{fig:app_heatmap_comparison} for the corresponding reconstruction heatmaps.}
    \label{fig:app_reconstruction}
\end{figure}

\begin{figure}[ht]
    \centering
    \begin{subfigure}[t]{0.48\textwidth}
        \centering
        \includegraphics[width=\textwidth]{figTHR_A_accuracy_vs_k.png}
        \caption{Dual-task: reconstruction accuracy vs.\ $k$ for each WD.
        Both add and mul require $k \geq 5$--$9$ components (depending on WD) to cross 90\%.}
        \label{fig:app_dual_acc_vs_k}
    \end{subfigure}
    \hfill
    \begin{subfigure}[t]{0.48\textwidth}
        \centering
        \includegraphics[width=\textwidth]{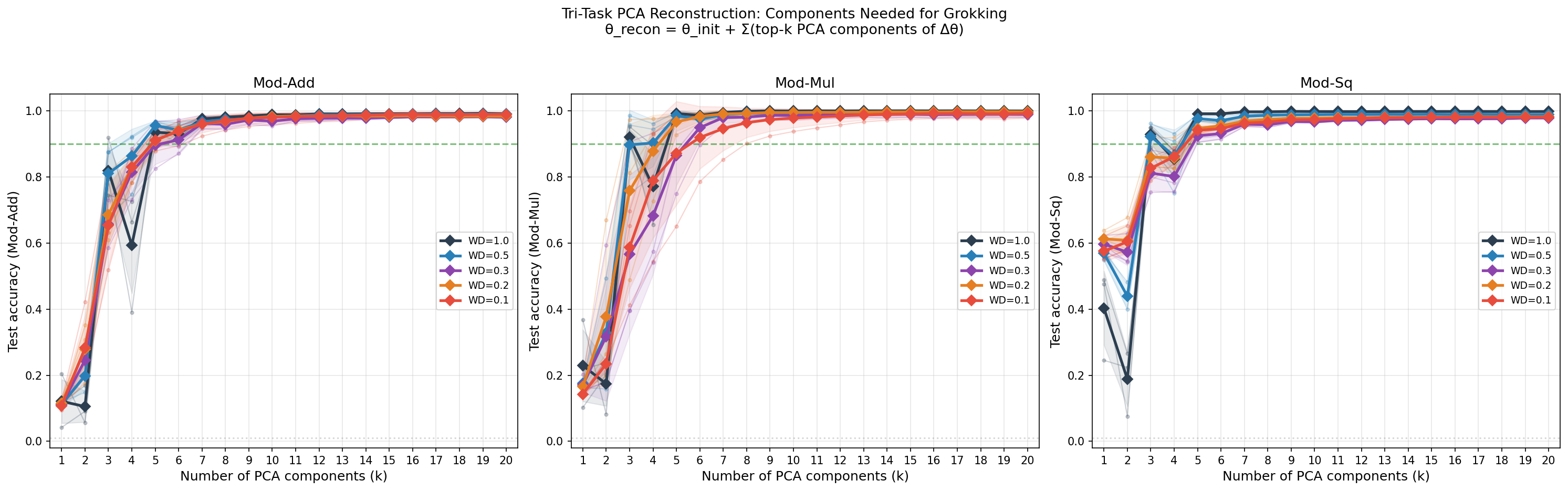}
        \caption{Tri-task: reconstruction accuracy vs.\ $k$ for each WD.
        All three tasks cross 90\% at $k^* \approx 3$--$8$, consistently lower than dual-task.}
        \label{fig:app_tri_acc_vs_k}
    \end{subfigure}
    \caption{Detailed comparison of reconstruction threshold $k^*$ between dual-task and tri-task settings
    (companion to \Cref{fig:kstar_comparison} in the main text).
    \textbf{(a)}~Dual-task models require more PCA directions; the chance-to-perfect transition is visible
    as a sigmoid-like jump.
    \textbf{(b)}~Tri-task models exhibit the same sharp transition but shifted leftward to lower $k$,
    indicating stronger constraint-induced compression.
    The systematic $k^*_{\text{dual}} > k^*_{\text{tri}}$ ordering holds across all five WD values
    and three random seeds (see \Cref{tab:app_dualtask_acc,tab:app_tritask_acc} for numerical details).}
    \label{fig:app_kstar_comparison}
\end{figure}

\begin{figure}[ht]
    \centering
    \begin{subfigure}[t]{0.48\textwidth}
        \centering
        \includegraphics[width=\textwidth]{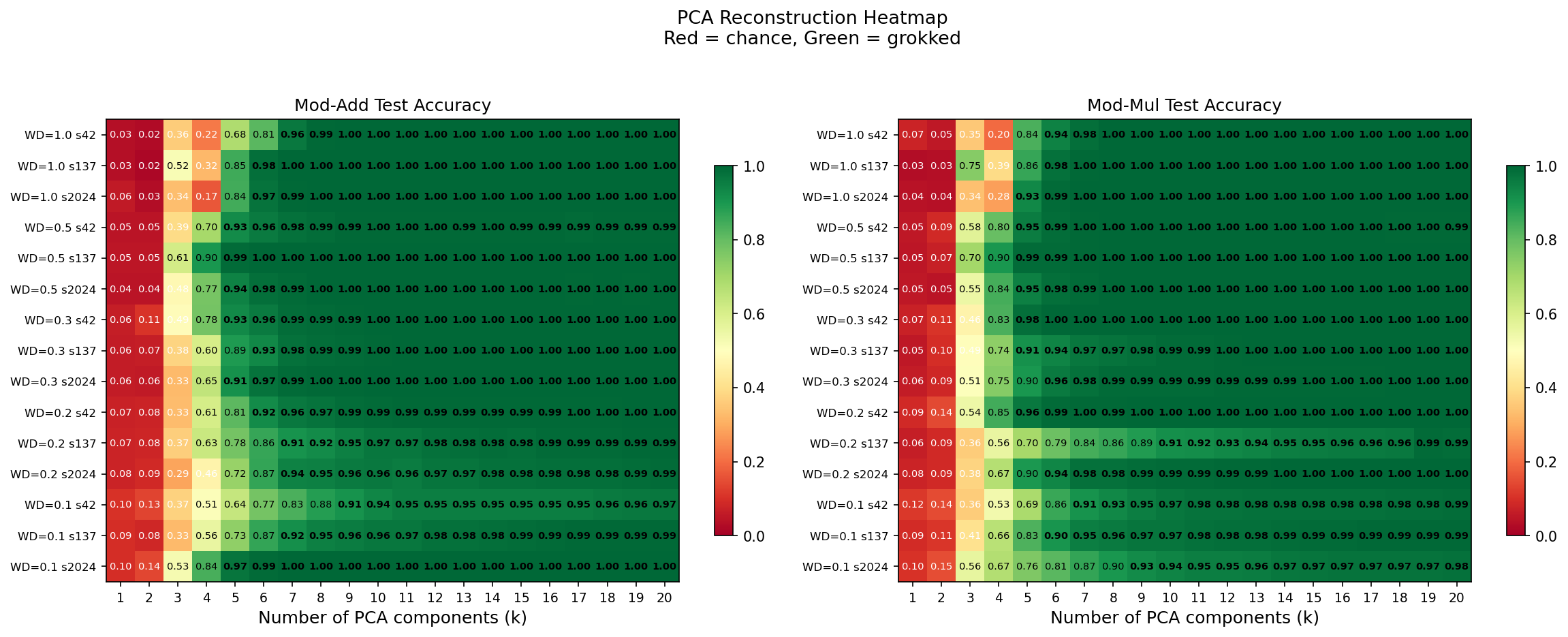}
        \caption{Dual-task: reconstruction accuracy heatmap (WD $\times$ $k$).
        The bright band shifts rightward at lower WD, reflecting increased $k^*$.}
        \label{fig:app_dual_heatmap_detail}
    \end{subfigure}
    \hfill
    \begin{subfigure}[t]{0.48\textwidth}
        \centering
        \includegraphics[width=\textwidth]{figTRI_THR_B_heatmap.png}
        \caption{Tri-task: reconstruction accuracy heatmap (WD $\times$ $k$).
        The transition band is shifted leftward relative to dual-task.}
        \label{fig:app_tri_heatmap_detail}
    \end{subfigure}
    \caption{Heatmap comparison of reconstruction accuracy across WD and $k$ for dual-task \textbf{(a)} and tri-task \textbf{(b)}.
    The boundary between chance-level (dark) and near-perfect (bright) accuracy occurs at lower $k$ for the tri-task setting, providing a
    visual summary of constraint-induced compression across the full WD phase diagram.}
    \label{fig:app_heatmap_comparison}
\end{figure}

\begin{figure}[ht]
    \centering
    \includegraphics[width=0.7\textwidth]{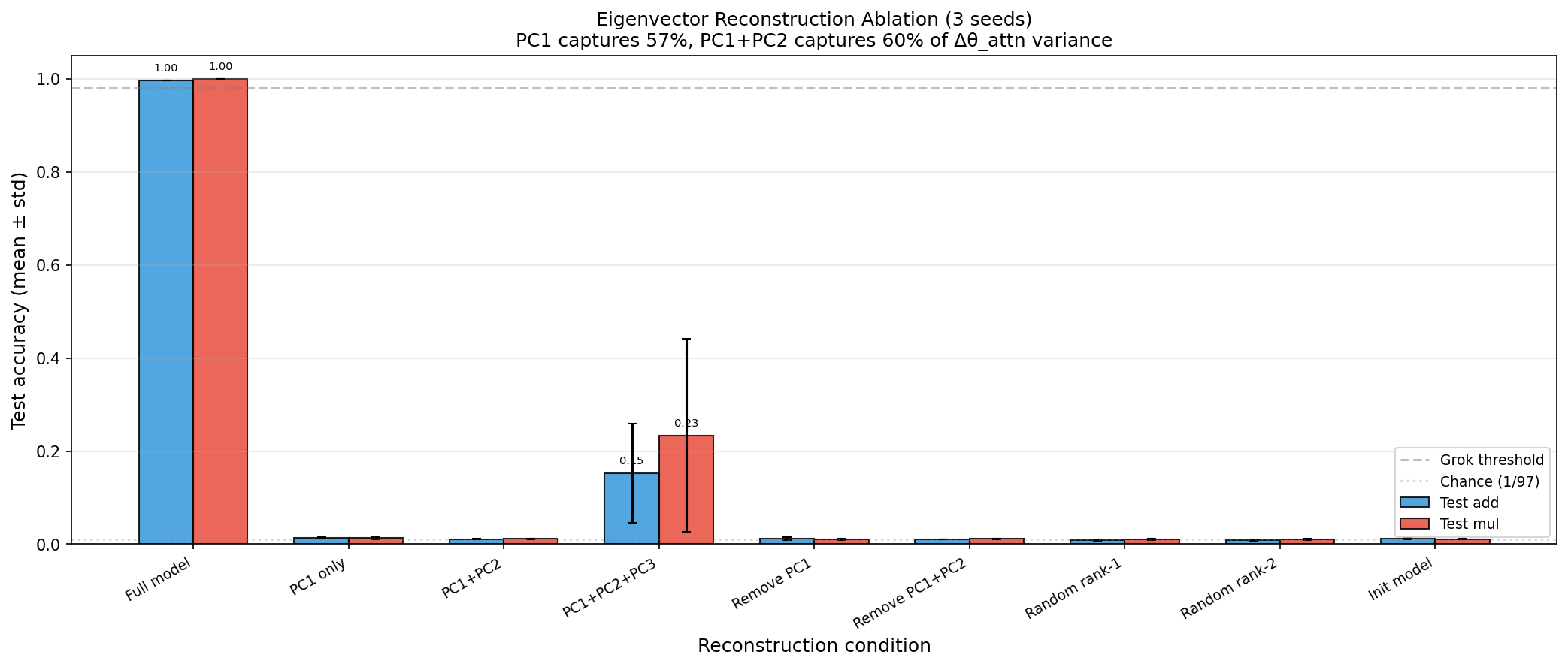}
    \caption{Dual-task gradient projection ablation (averaged across 3 seeds, WD=1.0). PCA projection at strength $s = 0.25$ delays grokking by 70--140\%, while random projection at the same strength has no effect.}
    \label{fig:app_dual_ablation}
\end{figure}

\section{Detailed PCA Reconstruction Tables}
\label{app:pca_tables}

The following tables provide complete per-$k$ reconstruction accuracy for both tri-task and dual-task settings across all weight decay values.
All data shown are for seed~42; results for other seeds are qualitatively identical (see $k^*$ summaries in \Cref{tab:tri_kstar,tab:dual_kstar}).
Bold rows indicate $k = k^*$, the minimum number of PCA components at which all tasks exceed 90\% test accuracy.

\begin{table}[ht]
\centering
\caption{Tri-task reconstruction accuracy vs.\ number of PCA components $k$ (seed~42).
  Each sub-table shows one weight decay value.
  \textbf{CumVar} is cumulative variance explained.
  Bold row = $k^*$ (first $k$ where all tasks $>90\%$).
  PC1 captures 89--94\% of variance but carries no generalization capability.}
\label{tab:app_tritask_acc}
\small
\begin{tabular}{@{}r r r r r | r r r r r@{}}
\toprule
\multicolumn{5}{c|}{\textbf{$\lambda = 1.0$ \quad ($k^* = 5$)}} & \multicolumn{5}{c}{\textbf{$\lambda = 0.5$ \quad ($k^* = 4$)}} \\
$k$ & CumVar & Add & Mul & Sq & $k$ & CumVar & Add & Mul & Sq \\
\midrule
 1 & 91.7\% &  20.4 &  36.7 &  47.5 &  1 & 93.9\% &   9.9 &  18.0 &  57.4 \\
 2 & 95.9\% &   5.6 &   8.2 &   7.5 &  2 & 97.1\% &  25.0 &  49.3 &  48.1 \\
 3 & 98.0\% &  74.4 &  89.2 &  94.8 &  3 & 98.8\% &  87.5 &  98.5 &  96.1 \\
 4 & 98.8\% &  72.5 &  76.2 &  87.7 &  \textbf{4} & \textbf{99.3\%} &  \textbf{91.8} &  \textbf{96.0} &  \textbf{93.0} \\
 \textbf{5} & \textbf{99.3\%} &  \textbf{96.5} &  \textbf{99.9} &  \textbf{99.3} &  5 & 99.5\% &  96.3 &  99.5 &  98.9 \\
 6 & 99.5\% &  91.7 &  99.2 &  99.2 &  6 & 99.7\% &  95.7 &  98.2 &  98.5 \\
 7 & 99.6\% &  97.8 &  99.8 &  99.7 &  7 & 99.7\% &  97.0 &  99.5 &  98.9 \\
 8 & 99.7\% &  97.8 &  99.9 &  99.7 &  8 & 99.8\% &  97.3 &  99.8 &  99.1 \\
10 & 99.8\% &  99.1 & 100.0 &  99.8 & 10 & 99.9\% &  97.9 &  99.9 &  99.0 \\
15 & 99.9\% &  99.3 & 100.0 &  99.9 & 15 & 99.9\% &  98.9 &  99.9 &  99.1 \\
\bottomrule
\end{tabular}

\vspace{0.5em}

\begin{tabular}{@{}r r r r r | r r r r r@{}}
\toprule
\multicolumn{5}{c|}{\textbf{$\lambda = 0.3$ \quad ($k^* = 7$)}} & \multicolumn{5}{c}{\textbf{$\lambda = 0.2$ \quad ($k^* = 5$)}} \\
$k$ & CumVar & Add & Mul & Sq & $k$ & CumVar & Add & Mul & Sq \\
\midrule
 1 & 93.5\% &  11.0 &  15.2 &  57.4 &  1 & 91.9\% &  11.0 &  16.1 &  58.7 \\
 2 & 96.1\% &  24.7 &  18.1 &  54.6 &  2 & 95.1\% &  17.6 &  26.5 &  56.6 \\
 3 & 97.6\% &  58.5 &  39.6 &  75.3 &  3 & 97.4\% &  60.9 &  81.2 &  83.0 \\
 4 & 98.5\% &  73.1 &  57.4 &  75.5 &  4 & 98.6\% &  82.9 &  93.1 &  82.6 \\
 5 & 99.0\% &  82.4 &  87.9 &  90.4 &  \textbf{5} & \textbf{99.0\%} &  \textbf{91.2} &  \textbf{98.6} &  \textbf{93.9} \\
 6 & 99.4\% &  87.1 &  96.7 &  91.4 &  6 & 99.3\% &  94.2 &  99.2 &  95.0 \\
 \textbf{7} & \textbf{99.6\%} &  \textbf{94.5} &  \textbf{99.1} &  \textbf{95.2} &  7 & 99.4\% &  95.8 &  99.7 &  96.7 \\
 8 & 99.7\% &  94.3 &  99.0 &  94.9 &  8 & 99.6\% &  96.6 &  99.7 &  97.3 \\
10 & 99.8\% &  95.3 &  99.4 &  96.2 & 10 & 99.7\% &  98.2 &  99.6 &  97.6 \\
15 & 99.9\% &  97.6 &  99.6 &  97.7 & 15 & 99.8\% &  98.5 &  99.8 &  97.9 \\
\bottomrule
\end{tabular}

\vspace{0.5em}

\begin{tabular}{@{}r r r r r@{}}
\toprule
\multicolumn{5}{c}{\textbf{$\lambda = 0.1$ \quad ($k^* = 5$)}} \\
$k$ & CumVar & Add & Mul & Sq \\
\midrule
 1 & 89.6\% &   9.5 &  14.2 &  55.1 \\
 2 & 93.7\% &  23.3 &  24.8 &  58.0 \\
 3 & 96.6\% &  63.1 &  69.6 &  82.9 \\
 4 & 98.0\% &  84.8 &  92.9 &  85.5 \\
 \textbf{5} & \textbf{98.8\%} &  \textbf{93.9} &  \textbf{98.8} &  \textbf{94.0} \\
 6 & 99.0\% &  96.7 &  98.5 &  94.6 \\
 7 & 99.2\% &  97.9 &  98.9 &  96.2 \\
 8 & 99.2\% &  98.4 &  99.2 &  96.6 \\
10 & 99.4\% &  99.1 &  99.7 &  97.4 \\
15 & 99.6\% &  99.3 &  99.6 &  97.8 \\
\bottomrule
\end{tabular}
\end{table}

\begin{table}[ht]
\centering
\caption{Dual-task reconstruction accuracy vs.\ number of PCA components $k$ (seed~42).
  Same format as \Cref{tab:app_tritask_acc}.
  The dual-task transition is equally sharp but requires more components ($k^* = 5$--$9$) than tri-task ($k^* = 3$--$8$).}
\label{tab:app_dualtask_acc}
\small
\begin{tabular}{@{}r r r r | r r r r@{}}
\toprule
\multicolumn{4}{c|}{\textbf{$\lambda = 1.0$ \quad ($k^* = 7$)}} & \multicolumn{4}{c}{\textbf{$\lambda = 0.5$ \quad ($k^* = 5$)}} \\
$k$ & CumVar & Add & Mul & $k$ & CumVar & Add & Mul \\
\midrule
 1 & 94.9\% &   3.5 &   7.4 &  1 & 96.8\% &   4.5 &   5.3 \\
 2 & 97.0\% &   2.4 &   5.4 &  2 & 98.2\% &   4.5 &   8.9 \\
 3 & 98.5\% &  36.1 &  35.3 &  3 & 98.8\% &  39.1 &  57.7 \\
 4 & 99.0\% &  21.9 &  19.7 &  4 & 99.5\% &  69.6 &  79.5 \\
 5 & 99.4\% &  68.0 &  83.5 &  \textbf{5} & \textbf{99.6\%} &  \textbf{92.5} &  \textbf{95.5} \\
 6 & 99.7\% &  81.4 &  94.3 &  6 & 99.7\% &  96.1 &  98.8 \\
 \textbf{7} & \textbf{99.8\%} &  \textbf{96.1} &  \textbf{98.4} &  7 & 99.8\% &  97.9 &  99.5 \\
 8 & 99.9\% &  99.3 &  99.8 &  8 & 99.9\% &  98.8 &  99.8 \\
10 & 99.9\% &  99.9 & 100.0 & 10 & 99.9\% &  99.5 &  99.9 \\
15 & 100.0\% &  99.9 & 100.0 & 15 & 100.0\% &  99.5 &  99.7 \\
\bottomrule
\end{tabular}

\vspace{0.5em}

\begin{tabular}{@{}r r r r | r r r r@{}}
\toprule
\multicolumn{4}{c|}{\textbf{$\lambda = 0.3$ \quad ($k^* = 5$)}} & \multicolumn{4}{c}{\textbf{$\lambda = 0.2$ \quad ($k^* = 6$)}} \\
$k$ & CumVar & Add & Mul & $k$ & CumVar & Add & Mul \\
\midrule
 1 & 95.7\% &   6.3 &   7.1 &  1 & 95.1\% &   7.1 &   8.8 \\
 2 & 97.5\% &  10.6 &  10.9 &  2 & 97.2\% &   7.6 &  14.1 \\
 3 & 98.3\% &  49.0 &  46.3 &  3 & 98.0\% &  33.0 &  54.3 \\
 4 & 99.2\% &  77.6 &  83.3 &  4 & 99.0\% &  60.7 &  84.7 \\
 \textbf{5} & \textbf{99.4\%} &  \textbf{92.8} &  \textbf{98.4} &  5 & 99.3\% &  80.9 &  96.2 \\
 6 & 99.6\% &  95.9 &  99.8 &  \textbf{6} & \textbf{99.5\%} &  \textbf{92.3} &  \textbf{98.8} \\
 7 & 99.7\% &  98.7 &  99.9 &  7 & 99.6\% &  96.4 &  99.5 \\
 8 & 99.8\% &  99.2 &  99.9 &  8 & 99.7\% &  97.5 &  99.5 \\
10 & 99.8\% &  99.8 & 100.0 & 10 & 99.8\% &  99.2 &  99.9 \\
15 & 99.9\% &  99.9 & 100.0 & 15 & 99.8\% &  99.4 & 100.0 \\
\bottomrule
\end{tabular}

\vspace{0.5em}

\begin{tabular}{@{}r r r r@{}}
\toprule
\multicolumn{4}{c}{\textbf{$\lambda = 0.1$ \quad ($k^* = 9$)}} \\
$k$ & CumVar & Add & Mul \\
\midrule
 1 & 93.8\% &  10.2 &  11.5 \\
 2 & 96.6\% &  13.5 &  14.5 \\
 3 & 97.7\% &  37.2 &  36.3 \\
 4 & 98.6\% &  51.1 &  53.2 \\
 5 & 98.8\% &  63.8 &  69.1 \\
 6 & 99.2\% &  77.4 &  85.8 \\
 7 & 99.3\% &  83.4 &  90.7 \\
 8 & 99.4\% &  88.3 &  93.3 \\
 \textbf{9} & \textbf{99.5\%} &  \textbf{90.7} &  \textbf{94.9} \\
10 & 99.5\% &  93.5 &  96.7 \\
15 & 99.6\% &  95.3 &  97.8 \\
\bottomrule
\end{tabular}
\end{table}

\section{Detailed Transverse Gradient Ablation Tables}
\label{app:ablation_tables}

The following tables provide extended detail for the orthogonal deletion dose-response experiments described in \Cref{sec:transverse_ablation}.
All experiments use WD $= 1.0$, seed~42.
``Delay'' is computed relative to the baseline ($\alpha = 0$) mean grokking step.
FAIL indicates that one or more tasks did not reach 90\% test accuracy within the step budget.

\begin{table}[ht]
\centering
\caption{Dual-task transverse gradient ablation: detailed results (WD=1.0, seed~42, 50k step budget).
  Columns show per-task grokking steps, the mean across tasks, absolute delay, and percentage delay.
  The sharp cliff at $\alpha = 0.10$ and non-monotonic recovery at $\alpha = 0.50$ are clearly visible.}
\label{tab:app_ortho_dual}
\small
\begin{tabular}{@{}r r r r r l@{}}
\toprule
$\alpha$ (\%) & Add Grok & Mul Grok & Mean & $\Delta t$ (steps) & Delay \\
\midrule
0 (baseline) & 14,100 & 14,000 & 14,050 & --- & --- \\
1 & 18,100 & 18,100 & 18,100 & $+4{,}050$ & $+29\%$ \\
2 & 21,200 & 20,900 & 21,050 & $+7{,}000$ & $+50\%$ \\
3 & 23,500 & 23,300 & 23,400 & $+9{,}350$ & $+67\%$ \\
5 & 37,000 & 36,800 & 36,900 & $+22{,}850$ & $+163\%$ \\
7 & 42,600 & 42,400 & 42,500 & $+28{,}450$ & $+203\%$ \\
10 & FAIL & FAIL & --- & --- & $\infty$ \\
15 & FAIL & FAIL & --- & --- & $\infty$ \\
25 & FAIL & FAIL & --- & --- & $\infty$ \\
\textbf{50} & \textbf{36,900} & \textbf{36,800} & \textbf{36,850} & $+22{,}800$ & $+163\%$ \\
\bottomrule
\end{tabular}

\vspace{0.8em}

\small
\textbf{Key observations:}
The delay is approximately linear in $\alpha$ for small deletions ($\Delta t \approx 4{,}000 \cdot \alpha$ for $\alpha \leq 0.03$), then accelerates super-linearly.
At $\alpha = 0.07$, both tasks are delayed to nearly identical steps (42,400--42,600), suggesting the tasks are tightly coupled in their response to transverse perturbation.
The add--mul gap, which is ${\sim}100$ steps at baseline, remains $<200$ steps across all successful conditions.
At $\alpha = 0.50$, grokking recovers at a delay comparable to $\alpha = 0.05$, consistent with the manifold redundancy hypothesis.
\end{table}

\begin{table}[ht]
\centering
\caption{Tri-task transverse gradient ablation: detailed results (WD=1.0, seed~42, 60k step budget).
  Same format as \Cref{tab:app_ortho_dual}.
  The tri-task setting shows the same fragility cliff at $\alpha = 0.10$ but no recovery at any higher deletion level.}
\label{tab:app_ortho_tri}
\small
\begin{tabular}{@{}r r r r r r l@{}}
\toprule
$\alpha$ (\%) & Add Grok & Mul Grok & Sq Grok & Mean & $\Delta t$ (steps) & Delay \\
\midrule
0 (baseline) & 25,600 & 18,200 & 23,400 & 22,400 & --- & --- \\
1 & 35,200 & 21,200 & 28,800 & 28,400 & $+6{,}000$ & $+27\%$ \\
2 & 42,400 & 30,200 & 36,000 & 36,200 & $+13{,}800$ & $+62\%$ \\
3 & 35,000 & 29,000 & 33,600 & 32,500 & $+10{,}100$ & $+45\%$ \\
5 & 35,400 & 26,800 & 33,800 & 32,000 & $+9{,}600$ & $+43\%$ \\
7 & 56,000 & 44,600 & 49,600 & 50,100 & $+27{,}700$ & $+124\%$ \\
10 & FAIL & FAIL & FAIL & --- & --- & $\infty$ \\
15 & FAIL & FAIL & FAIL & --- & --- & $\infty$ \\
25 & FAIL & FAIL & FAIL & --- & --- & $\infty$ \\
\bottomrule
\end{tabular}

\vspace{0.8em}

\small
\textbf{Key observations:}
Unlike dual-task, the tri-task delay is non-monotonic in the 2--5\% range: the mean delay peaks at $\alpha = 0.02$ ($+62\%$), dips at $\alpha = 0.03$--$0.05$ ($+43$--$45\%$), then surges at $\alpha = 0.07$ ($+124\%$).
This suggests complex interactions between the three tasks' transverse gradient components.
The task ordering (mul $\to$ sq $\to$ add) is preserved at all successful deletion levels, indicating that the relative difficulty hierarchy is robust to transverse perturbation.
At $\alpha = 0.07$, the mul--add gap widens from $7{,}400$ steps (baseline) to $11{,}400$ steps, suggesting that addition is more sensitive to transverse deletion than multiplication.
No recovery is observed at any deletion level $\geq 10\%$, consistent with the absence of redundant center manifolds in the tri-task setting.
\end{table}

\end{document}